\documentclass{article}

 \usepackage[preprint]{neurips_2026}

\usepackage[utf8]{inputenc} % allow utf-8 input
\usepackage[T1]{fontenc}    % use 8-bit T1 fonts
\usepackage{hyperref}       % hyperlinks
\usepackage{url}            % simple URL typesetting
\usepackage{booktabs}       % professional-quality tables
\usepackage{amsfonts}       % blackboard math symbols
\usepackage{nicefrac}       % compact symbols for 1/2, etc.
\usepackage{microtype}      % microtypography
\usepackage{xcolor}         % colors

\usepackage{graphicx}
\usepackage{subcaption}
\usepackage{multirow} % for multi-row cells in 

\usepackage{comment}
\newcommand{\say}[1]{``#1''}
\usepackage[inline]{enumitem}

\usepackage[textsize=tiny]{todonotes}
\usepackage{arydshln}
\usepackage{adjustbox}
\usepackage{placeins}

\newcommand{\br}[1]{\left\lbrace #1 \right\rbrace}
\newcommand{\term}[1]{\left( #1 \right)}
\newcommand{\Dist}[1]{\mathrm{Dist}\term{#1}}
\newcommand{\pr}[1]{\mathrm{Pr}\term{#1}}

\usepackage{amsmath}
\usepackage{amssymb}
\usepackage{mathtools}
\usepackage{amsthm}

\usepackage[most]{tcolorbox}
\usepackage{fontawesome5} % For nice icons like a lightbulb

\definecolor{loaytext}{RGB}{180, 0, 0}
\definecolor{loaybg}{RGB}{255, 230, 230}
\definecolor{moradtext}{RGB}{0, 180, 0}
\definecolor{moradbg}{RGB}{230, 255, 230}
\definecolor{jerrytext}{RGB}{0, 0, 180}
\definecolor{jerrybg}{RGB}{230, 230, 255}

\newtcolorbox{takeawaybox}[1]{%
  colback=blue!5!white,    % Main background color
  colframe=blue!75!black,  % Border color
  fonttitle=\bfseries,
  colbacktitle=blue!85!black,
  enhanced,                % Allows for advanced styling like shadows
  attach boxed title to top left={yshift=-2mm, xshift=5mm},
  boxed title style={sharp corners, rounded corners=southeast, colback=blue!75!black},
  title={\faLightbulb \hspace{0.5em} Key Takeaway: #1},
  sharp corners,
  rounded corners=northwest, % Mixed corners look more "designed"
  drop fuzzy shadow,       % Gives it that "floating" animated depth
  left=10pt,
  right=10pt,
  top=10pt,
  bottom=10pt
}

\newtcolorbox{experimenttakeawaybox}[1]{%
  colback=gray!5!white,
  colframe=gray!60!black,
  fonttitle=\bfseries,
  colbacktitle=gray!80!black,
  coltitle=white,
  enhanced,
  attach boxed title to top left={yshift=-2mm, xshift=5mm},
  boxed title style={
    sharp corners,
    rounded corners=southeast,
    colback=gray!80!black
  },
  title={\faBookmark \hspace{0.5em} Takeaways: #1},
  sharp corners,
  rounded corners=northwest,
  drop fuzzy shadow,
  left=10pt, right=10pt, top=10pt, bottom=10pt
}

\usepackage[capitalize,noabbrev]{cleveref}

\usepackage{hyperref}
\usepackage{wrapfig}

\usepackage[inline]{enumitem}

\usepackage[normalem]{ulem}
\usepackage{tablefootnote}
\usepackage{afterpage}

\usepackage{fontawesome5}

\title{SpurAudio: A Benchmark for Studying Shortcut Learning in Few-Shot Audio Classification}

\author{%
  Giries Abu Ayoub~\thanks{Equal contribution} \\
  Department of Computer Science\\
  University of Haifa\\
  \texttt{jerryabuayob@gmail.com} \\
  \And
  Morad Tukan\footnotemark[1] \\
  Independent Researcher \\
  \texttt{muradtuk@gmail.com} \\
  \AND
  Loay Mualem\thanks{Corresponding Author} \\
  University of Stuttgart, Germany \\
    \textsuperscript{}IMPRS-IS\thanks{International Max Planck Research School for Intelligent Systems.}, Germany \\
  \texttt{loaymua@gmail.com} \\
}

\begin{document}

\maketitle

\begin{abstract}
Few-shot classification (FSC) is widely used for learning from limited labeled data, yet most evaluations implicitly assume that target concepts are independent of contextual cues. In real-world settings, however, examples often appear within rich contexts, allowing models to exploit spurious correlations between foreground content and background signals. While such effects have been studied in few-shot image classification, their role in few-shot audio classification remains largely unexplored, and existing audio benchmarks offer limited control over contextual structure. We introduce \textbf{SpurAudio}, a benchmark that leverages the natural separability of foreground events and background environments in audio to enable controlled, multi-level evaluation of contextual shifts across support and query sets. Using this benchmark, we show that many state-of-the-art few-shot methods suffer severe performance degradation when background correlations are disrupted, despite achieving similar accuracy under standard evaluation protocols. Crucially, this vulnerability persists even in large pretrained audio foundation models, ruling out limited backbone capacity as an explanation. Moreover, methods that appear comparable under conventional benchmarks can exhibit markedly different sensitivity to spurious correlations, revealing systematic algorithmic strengths and vulnerabilities tied to how feature representations interact with classifier heads at inference time. These findings provide new insight into the behavior of few-shot methods in audio and highlight the need for benchmarks that explicitly probe context dependence when evaluating FSC models.
\faGithub \space \href{https://github.com/Jerryaa98/SpurAudio}{\underline{https://github.com/Jerryaa98/SpurAudio}}
\end{abstract}
\section{Introduction}

Few-shot classification (\emph{FSC}) aims to recognize novel classes from only a handful of labeled examples~\cite{vinyals2016matching,snell2017prototypical,finn2017model,Wang2021FewShot}.
While recent advances in representation learning have substantially improved data efficiency, \emph{FSC} remains particularly challenging in real-world \emph{audio} applications.
Sound events are often rare, costly to annotate, and highly variable in their acoustic realization, making large-scale supervised training impractical.
As a result, few-shot audio classification is critical for many high-impact domains, including bioacoustic monitoring~\cite{Ghani2024Bioacoustics, Nolasco2023Learning, You2023Transformer, Moummad2023Pretraining, Liu2024Few, Ijaz2024Reshaping, McEwen2024Active, Jana2025Automated}, industrial fault diagnosis~\cite{Siraj2023Few, Liang2023Federated, Saleem2025Optimized, Zabin2025Few}, and healthcare audio analysis~\cite{Disha2025Cough, Florea2025Exploring}, where failures can carry significant ecological, economic, or safety consequences.

A defining challenge in audio, however, lies in its \emph{additive} nature.
Unlike images, where objects are often spatially separable from their surroundings, audio foreground events are superimposed on background sounds in the time--frequency domain~\cite{wichern2019wham,maciejewski2020whamr}.
In practice, target sounds rarely occur in isolation and are embedded in rich and often predictive acoustic contexts.
This makes few-shot audio models vulnerable to exploiting \emph{spurious correlations} between class labels and background cues: models can achieve strong accuracy for the \emph{wrong reason} by keying on context rather than semantic foreground content.
Such non-causal shortcuts can artificially inflate performance under matched training and testing conditions but lead to abrupt failures when background contexts shift.

Prior work has demonstrated that audio representations are sensitive to background interference, degradations, and polyphony~\cite{Salamon2017Deep, Turpault2021Sound, Abesser2023Robust}.
More recently, robustness-oriented approaches such as \emph{RobustCLAP}~\cite{selvakumar2025audio} and related methods aim to improve invariance to noise, corruption, or background variation at the \emph{representation learning} level.
However, these studies are predominantly conducted in supervised or zero-shot settings and do not examine \emph{episodic few-shot generalization}, where models must rapidly adapt from only a handful of labeled examples.
As a result, it remains unclear how much current few-shot audio methods rely on contextual shortcuts and how fragile their performance is under controlled background shifts.

\begin{figure*}[!htb]
    \centering
    \includegraphics[width=0.9\textwidth]{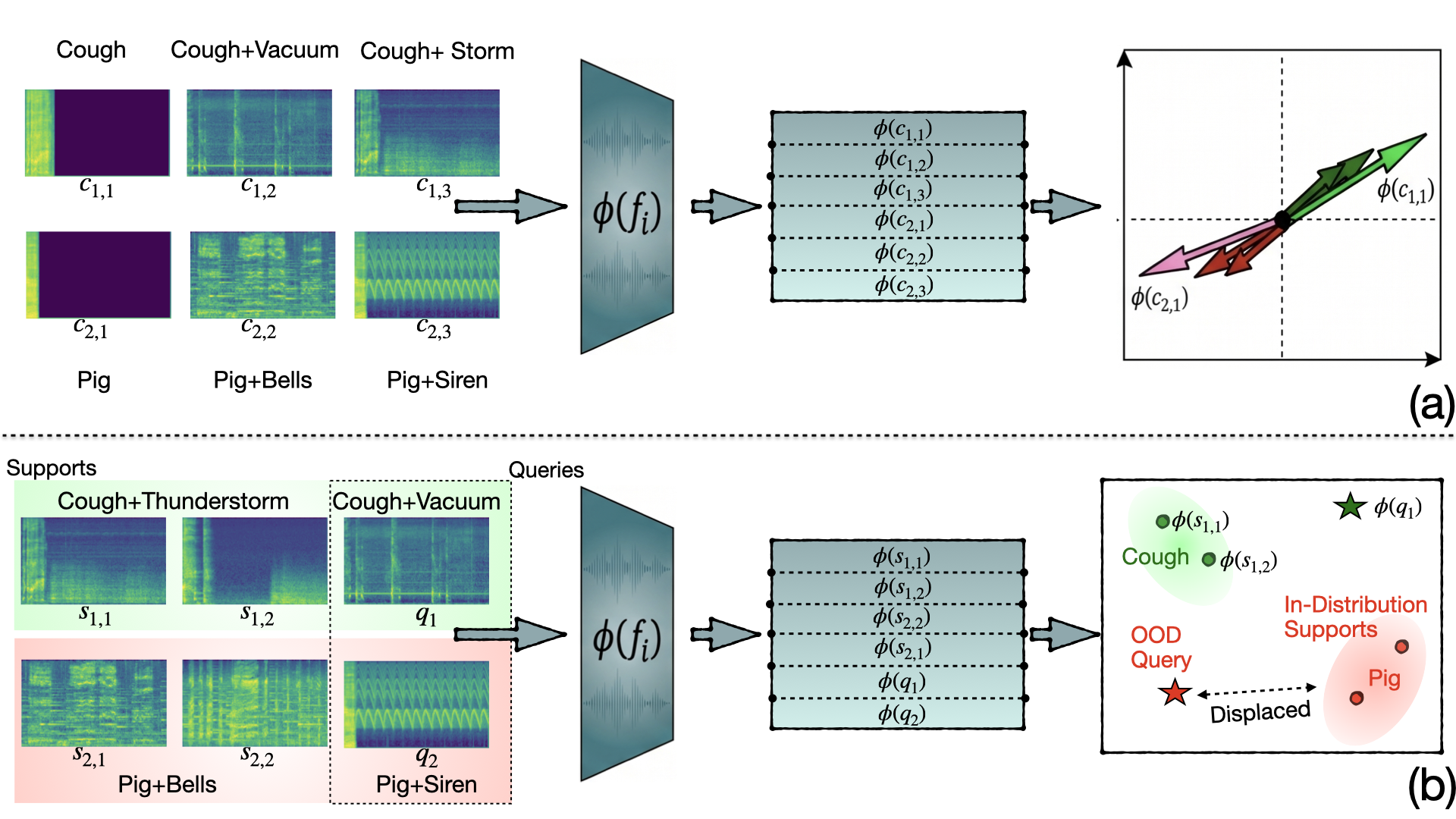}
    \caption{
    {Visualization of SpurAudio's episodic structure.} (a) A 1-shot 2-way episode: two foreground classes (Coughing and Pig) are mixed with different backgrounds (e.g., Vacuum Cleaner, Thunderstorm, Church Bells, Siren) and projected by $\phi$ into a feature space. Samples from the same foreground class point in the same direction, while background noise changes the length of mixed sample vectors (dark hues) relative to clean ones (light hues). (b) An OOD episode where support (s) and query (q) contain disjoint backgrounds (e.g., Church Bells vs. Siren). Despite sharing the same foreground class, queries are displaced from their correct support clusters in the metric space, leading to misclassification.
    }
    \label{fig:iid_ood_illustration}
\end{figure*}

In a typical \emph{FSC} episode, models generalize from a small labeled \emph{support set} to an unlabeled \emph{query set}~\cite{vinyals2016matching,snell2017prototypical}.
Although modern methods achieve strong performance under controlled evaluation protocols, they often degrade substantially under realistic {out-of-distribution (\emph{OOD})} conditions.
Following prior work in vision~\cite{Zhang2024MetaCoCo,Sagawa2020Waterbirds}, OOD failures in FSC can arise from two sources:
(i) {cross-domain \emph{FSC}}, where source and target domains differ (e.g., speech $\rightarrow$ music), and
(ii) {spurious-correlation \emph{FSC} (\emph{SC-FSC})}, where semantic classes remain unchanged but contextual cues vary.

This work focuses on \emph{SC-FSC}, a failure mode that is particularly pernicious in audio.
Because foreground and background signals are inseparably mixed, consistent co-occurrence patterns during training can encourage models to rely on contextual shortcuts rather than semantic foreground content.
This phenomenon is closely related to \emph{shortcut learning}~\cite{geirhos2020shortcut} and the ``Clever Hans'' effect~\cite{Liu2024CleverHans}.
When such correlations are broken at test time, for example, when a machine fault occurs in a different factory environment, performance can degrade catastrophically, undermining real-world deployment.

\subsection{Positioning and Gap in Existing Work}
Existing audio robustness studies primarily investigate domain shifts across datasets, recording conditions, or acoustic scenes~\cite{Heggan2022MetaAudio}.
Even large-scale datasets such as \emph{FSD50K}~\cite{fonseca2021fsd50k} are typically used for supervised or zero-shot training without  controlling or disentangling foreground--background correlations.
While robustness-focused methods (e.g., \emph{RobustCLAP}~\cite{selvakumar2025audio}) aim to improve invariance to noise or corruption at the representation level, they do not evaluate \emph{episodic few-shot generalization} under controlled background manipulations.
Consequently, current benchmarks predominantly evaluate matched conditions or broad domain shifts~\cite{Heggan2022MetaAudio}, often conflating semantic foreground content with background context.
They cannot isolate the effect of non-causal background correlations on episodic generalization and may  overestimate robustness by rewarding reliance on contextual shortcuts.
As a result, the effect of spurious contextual cues on few-shot audio learning remains largely uncharacterized, leaving a critical blind spot in understanding few-shot audio generalization.

\subsection{Our Work}
To address this gap, we introduce \textbf{SpurAudio}, a benchmark systematically designed to isolate and evaluate spurious correlations in few-shot audio classification; see Figure~\ref{fig:iid_ood_illustration} for an illustration of SpurAudio.
Our data is a obtained from mixing foreground events from five real-world datasets with semantically unrelated background textures. This controlled mixing induces strong correlations in the support set (e.g., Class~A with Background~X) while varying background conditions in the query set (e.g., Class~A with Background~Y), enabling clean disentanglement between causal foreground learning and shortcut reliance.
Crucially, SpurAudio serves as a diagnostic \emph{dataset} for analyzing failure modes across different families of FSC methods.

Beyond providing a controlled benchmark, SpurAudio enables in-depth analysis of few-shot audio methods under contextual shifts.
Using this benchmark, we show that many state-of-the-art \emph{FSC} approaches suffer severe performance degradation when background correlations are disrupted, despite achieving similar accuracy under standard evaluation protocols.
Moreover, methods that appear comparable under conventional benchmarks can exhibit markedly different sensitivity to spurious correlations because few-shot performance is shaped by how well feature representations align with classifier heads.
Importantly, this vulnerability is not confined to small backbones: it persists across large pretrained audio foundation models, indicating that spurious background reliance is a fundamental property of few-shot audio inference rather than a limitation of representational capacity. These observations reveal systematic strengths and vulnerabilities of current \emph{FSC} algorithms and highlight the importance of benchmarks that explicitly probe context dependence when evaluating few-shot audio models.

% We evaluate metric-based, meta-learning, and fine-tuning \emph{FSC} approaches on SpurAudio using widely adopted backbones chosen to align with prior \emph{FSC} literature and real-world deployment constraints. 

More importantly, our work highlights subtle patterns that not only open new research directions within few-shot learning, but also point to broader implications for a wide range of audio tasks. Hence, our contributions are three-fold:
\begin{itemize}
    \item We introduce \textbf{SpurAudio}, a controlled {benchmark} enabling manipulation of foreground--background correlations across multiple audio domains.
    \item We characterize {spurious-correlation \emph{OOD} failures in few-shot audio classification}, showing state-of-the-art methods collapse when background contexts shift.
   \item We provide an extensive benchmark of metric-based, meta-learning, contrastive, transductive and fine-tuning approaches, spanning both standard backbones and large pretrained audio foundation models, highlighting systematic context reliance and motivating future work on context-robust few-shot audio learning.
\end{itemize}

\section{SpurAudio Dataset}

This section presents \textbf{SpurAudio}, a dataset specifically designed to study the impact of spurious correlations in few-shot audio classification. SpurAudio is assembled by aggregating audio samples from five publicly available datasets spanning diverse acoustic domains:
\begin{enumerate*}[label=(\roman*)]
\item ESC-50~\cite{piczak2015dataset}: A benchmark dataset of 50 environmental sound classes, including animal vocalizations, natural phenomena, and human activities.
\item UrbanSound8K~\cite{Salamon:UrbanSound:ACMMM:14}: A collection of 8,732 urban audio clips across 10 categories, such as sirens, dog barks, and drilling.
\item VocalSound~\cite{gong_vocalsound}: A dataset of human-produced vocal imitations and sound effects.
\item WILD DESED~\cite{xiao2024wilddesed}: Weakly labeled recordings captured in a variety of outdoor acoustic environments.
\item USM~\cite{abesser2022classifying}: A large-scale dataset of sound events embedded within complex acoustic scenes.
\end{enumerate*}

\subsection{Sound Event Generation}
\label{sec:annotators}
To synthesize realistic sound events that occur in the “wild,” we define two complementary concepts that partition the audio collection:
\begin{enumerate*}[label=(\Roman*)]
\item Foreground (FG): the target event class to be recognized within a few-shot learning episode; and
\item {Background (BG)}: an audio clip sampled from a semantically unrelated class, introduced as a confounding context.
\end{enumerate*}
Foreground and background classes are paired to be {semantically independent}; for example, a “dog barking” foreground combined with “park noise” in the background while still reflecting combinations that plausibly co-occur in real-world acoustic environments. To ensure diversity and avoid over-representation, we further limit the repeated use of the same background class across multiple foreground classes.

\textbf{Data Generation Flow.}
Given a pair of sound clips, a foreground $x_{\mathrm{fg}}(t)$ and a background $x_{\mathrm{bg}}(t)$, our objective is to ensure that the resulting mixture represents a scenario that humans would perceive as a plausible real-world co-occurrence. To this end, two annotators conducted a three-stage curation process:
\begin{enumerate*}[label=(\roman*)]
\item partitioning the complete collection of sound clips into foreground and background sets while maintaining maximal connectivity between them;
\item associating each foreground class with four distinct background classes; and
\item manually curating the resulting combined sound events.
\end{enumerate*}

\textbf{Mixing Process.}
To generate mixtures that resemble naturally occurring sound events, we adopt the mixing procedure proposed in~\cite{wichern2019wham, maciejewski2020whamr}. Given $x_{\mathrm{fg}}(t)$ and $x_{\mathrm{bg}}(t)$, both signals are first resampled to $16$~kHz and trimmed or padded to a fixed duration of $T=5$ seconds, yielding $\hat{x}_{\mathrm{fg}}(t)$ and $\hat{x}_{\mathrm{bg}}(t)$. We then compute the integrated loudness of each signal using the EBU R128 (LUFS) standard and scale the background to a fixed perceptual margin of $8$~dB below the foreground prior to mixing. The final mixture is peak-normalized to prevent clipping. Formally, the mixed signal is defined as
\begin{equation}
\label{eq:mixing}
x_{\mathrm{mix}}(t) = \hat{x}_{\mathrm{fg}}(t) + \alpha \, 10^{\nicefrac{(L_{fg} - L_{bg} - \gamma)}{20}} \, \hat{x}_{\mathrm{bg}}(t),
\end{equation}
where $\gamma = 8$. Details of the foreground–background class pairings are provided in Appendix~\ref{sec:appendix_FG_BG_mappiing}.

\textbf{Choice of $\gamma$.}
Empirical analyses of everyday acoustic environments indicate that naturally occurring foreground–background signal-to-noise ratios span a broad positive range. Large-scale in-situ measurements report that most SNRs lie between approximately $2$~dB and $14$~dB, with mean values around $7$--$8$~dB in noisy daily settings~\cite{wu2018characteristics}. These statistics characterize the acoustic structure of real environments rather than listener-dependent perception. Based on these observations, we set the perceptual loudness margin to $\gamma = 8$~dB, yielding mixtures in which background interference remains audible without overwhelming the foreground signal.

\textbf{Manual Quality Control.}
The automated mixing process may introduce unintended semantic overlap, for example, a “traffic noise” background clip containing a siren, which could confound analysis of spurious correlations. To mitigate this, annotators evaluated each mixed sound event according to the following criteria:
\begin{enumerate*}[label=(\roman*)]
\item degree of acoustic similarity between the background and the foreground;
\item whether the background overwhelms the foreground;
\item whether the background is inaudible; and
\item presence of additional unintended sound events beyond foreground and background labels. 
\end{enumerate*}
Each criterion is scored on a scale from $1$ to $5$, with higher scores indicating clearer perceptual separation. Mixed events with an average score below $4$ are discarded. In total, $50{,}116$ mixtures were generated, from which SpurAudio comprises a curated subset of $16{,}378$ sound events; See Appendix~\ref{sec:appendix_FG_BG_mappiing} for the complete mapping between foreground classes and background contexts.

\section{Families of Few-Shot Learning Methodologies}
\label{sec:fsc}

Few-shot classification (\emph{FSC}) is casted as an episodic learning problem. Each episode $\mathcal{T}$ corresponds to an $N$-way $K$-shot classification task, composed of a \emph{support set} $\mathcal{S} \equiv \{(x_i, y_i)\}_{i=1}^{N \cdot K}$ and a \emph{query set} $\mathcal{Q} \equiv \{(x_j, y_j)\}_{j=1}^{M}$. The labels $y_i, y_j \in \mathcal{C}$, where $\mathcal{C}$ is the set of $N$ classes sampled for that episode. The objective is to infer the labels of the query examples in $\mathcal{Q}$ using only the labeled support examples in $\mathcal{S}$. In the standard setup, during training, \emph{FSC} approaches learn a model over a collection of base classes, denoted $\mathcal{Y}_{\text{train}}$, and are then evaluated on a disjoint set of novel classes $\mathcal{Y}_{\text{test}}$.

In our scenario, each input signal $x$ can contain both a \emph{foreground} event (which determines the class label) and a \emph{background} acoustic environment. This compositional nature inherently gives rise to spurious correlations between the class labels and the background components of the signal. To this end, We examine five principal categories of \emph{FSC} methods: \begin{enumerate*}[label=(\roman*)]
    \item metric-based,
    \item meta-learning-based,
    \item fine-tuning-based,
    \item transductive, and
    \item contrastive-based.
\end{enumerate*}

\textbf{Metric-Based Methods.}
Metric-based \emph{FSC} learns an embedding function 
$\theta : \mathcal{X} \rightarrow \mathbb{R}^d$ that brings samples from the same class close together while separating different classes. Support examples for each class are embedded and summarized by a prototype, typically the mean of their embeddings.  
Let $\br{\theta(x^j_i)}_{i=1}^K$ be the embeddings of $K$ support samples for class $j \in [N]$, and define the prototype as $p^j = \frac{1}{K} \sum_{i=1}^K \theta(x^j_i)$.  
For a query $q$, the class posterior is computed via a softmax over negative distances: $
\pr{q \text{ belongs to class } j \mid \mathcal{P}} = 
\frac{e^{-\Dist{\theta(q), p^j}}}{\sum_{l \in [N]} e^{-\Dist{\theta(q), p^l}}}.$

\textbf{Meta-Learning Based Methods.}
Meta-learning trains models over a distribution of tasks to enable rapid adaptation from limited data.  
Tasks $\mathcal{T} \sim \mathcal{S} \times \mathcal{Q}$ comprise support $\mathcal{S}_i$ and query $\mathcal{Q}_i$ sets. Starting from meta-parameters $\theta$, task-specific parameters are obtained via an adaptation operator $\mathcal{A}$, i.e., $
\theta_i = \mathcal{A}(\theta, \mathcal{S}_i),$ and the meta-objective minimizes expected query loss: $
\min_{\theta} \mathbb{E}_{\mathcal{T}_i} \left[ \mathcal{L}_{\mathcal{T}_i}(\theta_i; \mathcal{Q}_i) \right].$

A canonical example is MAML~\cite{finn2017model}, which learns an initialization $\theta$ such that a few gradient steps on the support set yield good task performance:
$
\theta_i^{(k+1)} = \theta_i^{(k)} - \alpha \nabla_{\theta_i^{(k)}} \mathcal{L}(\theta_i^{(k)}; \mathcal{S}_i), 
\quad \theta_i^{(0)} = \theta,$  
with meta-objective
$
\min\limits_{\theta} \sum_i \mathcal{L}(\theta_i^{(M)}; \mathcal{Q}_i),
$  
where $M$ is the number of inner-loop updates.

\textbf{Fine-Tuning-Based Methods.}
Fine-tuning-based FSC first trains a backbone on base classes to learn transferable representations, then adapts to novel classes by replacing and fine-tuning the classifier optionally updating parts of the backbone using the support set. While this enables greater adaptation than fixed-embedding methods, it relies on very limited data, making it sensitive to optimization choices, prone to overfitting, and susceptible to reinforcing spurious correlations learned during pretraining, often resulting in high performance variability across tasks.

\textbf{Contrastive-Based Methods.}
Contrastive learning aims to learn representations in which semantically related audio samples are mapped close together, while unrelated samples are pushed apart. In few-shot audio classification, such objectives are commonly used to improve representation quality by leveraging data augmentations or weak supervision, and have been shown to enhance overall robustness. A typical formulation employs the InfoNCE loss, which encourages similarity between positive pairs while contrasting them against negatives.

\textbf{Transductive-Based Methods.}
Transductive methods leverage the statistical distribution of the unlabeled query set during inference by processing query samples collectively rather than in isolation, they effectively mitigate support set bias and better align representations across novel tasks.

%, i.e., $
% \mathcal{L}_{\text{InfoNCE}} = - \log \frac{\exp(\mathrm{sim}(z_i, z_i^+)/\tau)}{\sum_j \exp(\mathrm{sim}(z_i, z_j)/\tau)}.$

% We include contrastive-based methods in our benchmark to evaluate whether the invariances learned through contrastive objectives effectively reduce sensitivity to spurious foreground--background correlations, or whether such correlations persist despite contrastive training under contextual distribution shifts.

\section{Experiments}
\label{sec:experiments}

In this section, we study the effect of spurious correlations in our dataset, \emph{SpurAudio}, under few-shot learning settings. Specifically, we evaluate $1$ and $5$ shots classification tasks, with $10$ query samples per class, across different encoders and \emph{FSC} algorithms. For each configuration, performance is averaged over three random seeds, and we report both the mean classification accuracy and standard deviation; we refer the reader to Section~\ref{sec:experimental_setup} in the appendix for the full experimental setup.

\textbf{Roadmap.}
First, we introduce \emph{FSC} evaluation tasks that control foreground--background relationships in audio episodes, enabling systematic manipulation of background context across episodes (Section~\ref{sec:highlighting_effect_suprious_backgrounds}). Second, we quantify the \emph{IID} vs.\ \emph{OOD} performance gap across multiple \emph{FSC} model families and backbone architectures (Tables~\ref{tab:conv64},~\ref{tab:resnet12}, and~\ref{tab:resnet18}), and stress-test the proposed tasks by progressively increasing \emph{OOD} difficulty, revealing a consistent amplification of the \emph{IID}--\emph{OOD} gap (Section~\ref{seC:effect_of_suprious_correlation}). Third, we conduct a diagnostic analysis by visualizing embedding spaces of support and query samples and studying the effect of the mixing coefficient $\alpha$ (Eq.~\eqref{eq:mixing}) on the magnitude of the gap (Sections~\ref{sec:embedding_analysis} and~\ref{sec:appendix_embeddings_of_FSL}); as an additional validation, we demonstrate that the generated audio mixtures resemble real-world recordings by showing close proximity between CLAP embeddings of SpurAudio and those of FSD50K~\cite{fonseca2021fsd50k} (Sections~\ref{sec:distribution_of_spuraudio} and~\ref{sec:appendix_analysis_distribution}). Fourth, we isolate foreground signals and re-evaluate \emph{FSC} methods to verify that background content is the primary source of the shortcut learning (See Figure~\ref{fig:clean_queries_comparison}), then use head--backbone replacement studies to disentangle representation learning from decision rules, revealing how different \emph{FSC} algorithms handle spurious correlations (Section~\ref{sec:appendix_Cross-Family_Analysis_of_Few-Shot_Learning_Deeper_Algorithms}). Finally, to rule out the possibility that the observed \emph{IID}--\emph{OOD} gap is merely an artifact of limited CNN capacity, we extend our analysis to large pretrained audio models used as frozen encoders in combination with a wide range of non-backbone-dependent few-shot heads (Section~\ref{sec:large_model_experiments}). 

\textbf{Training \& Evaluation.}
We adapt the LibFewShot~\cite{li2023libfewshot} framework to work with audio inputs, while adopting its training parameters and implementing the \emph{FSC} algorithms described above. All methods are evaluated using episodic classification accuracy under both \emph{IID} and \emph{OOD} task sampling. Results are averaged over multiple random seeds and evaluation episodes. To quantify sensitivity to spurious correlations and distribution shifts, we report the accuracy gap $\Delta = \mathrm{Acc}_{\mathrm{IID}} - \mathrm{Acc}_{\mathrm{OOD}}$, where larger values indicate greater performance degradation under \emph{OOD} conditions.

\subsection{Highlighting the Effect of Spurious Background Shifts}
\label{sec:highlighting_effect_suprious_backgrounds}
The \emph{SpurAudio} dataset is split into three main subsets: \begin{enumerate*}[label=(\roman*)]
    \item training, 
    \item validation, and 
    \item test.
\end{enumerate*}
Training and validation primarily consist of \emph{IID} (in-distribution) tasks. The test set is further divided into two subsets: one composed of \emph{IID} tasks, and the other of \emph{OOD} (out-of-distribution) tasks. Both subsets share the same set of foreground sound events but differ in their background pairings, enabling us to assess the effect of spurious correlations introduced by background context. Specifically, we focus on two types of tasks: 
\begin{enumerate*}[label=(\Roman*)]
    \item {\emph{IID} tasks} -- support and query examples share the same background pairings for each of the $N$ classes, and
    \item {\emph{OOD} tasks} -- the background pairings are altered to break spurious foreground--background correlations, challenging the model to rely on the foreground content rather than contextual shortcuts.
\end{enumerate*}

\textbf{\emph{OOD} task generation.} 
For each \emph{OOD} task, each class’s foreground sound event is paired with background sounds that may also appear with other classes’ foregrounds. This creates overlap in background usage across classes, explicitly breaking the spurious association while keeping $N$, $K$, and the number of query samples unchanged. In Section~\ref{sec:experiments}, we demonstrate the resulting performance gap between \emph{IID} and \emph{OOD} tasks.

\subsection{In-Depth Analysis}

% \moradcomment{Add details about training from LibFewShot}

\subsubsection{Effect of Spurious Correlation Strength}
\label{seC:effect_of_suprious_correlation}

\begin{wrapfigure}[19]{r}{0.5\textwidth}
    \vspace{-\intextsep-0.2cm}% kills top space
    \centering
    \includegraphics[width=0.5\textwidth]{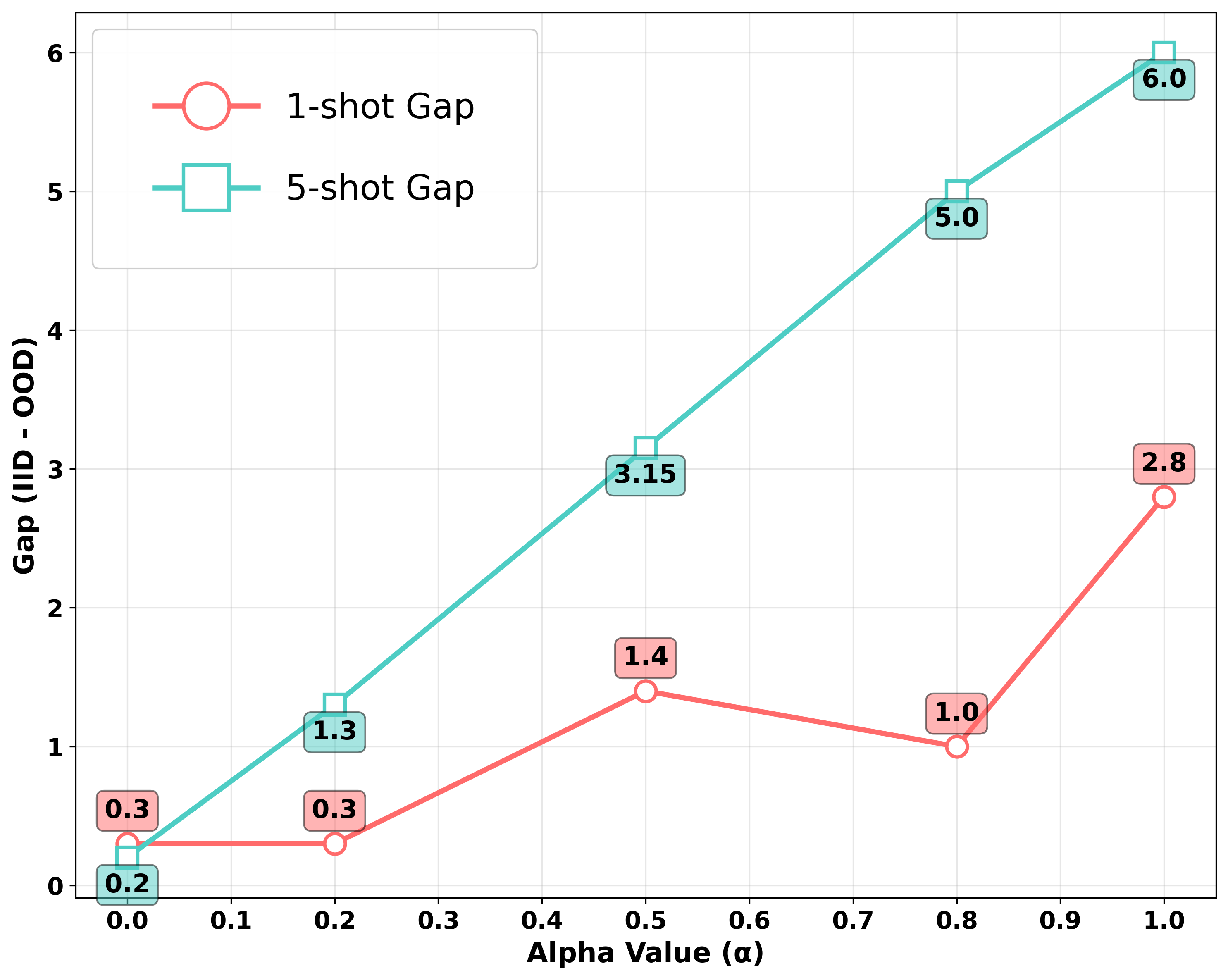}
    
    \caption{{IID--OOD gap versus spurious correlation strength.} We plot $\Delta(\alpha)$ for 1/5 shot. The gap grows with $\alpha$, showing strong foreground -- background correlations increase \emph{OOD} degradation.}
    \label{fig:alpha_gap}
    \vspace{-\baselineskip}% kills bottom space
\end{wrapfigure}
In the standard \emph{OOD} evaluation, support and query backgrounds are sampled such that some background patterns seen in the support set do not appear in the query set of other classes. 
To more directly probe the effect of spurious correlations, we construct a \emph{stronger correlation} setting in which every background present in the support set of each class in the test split also appears in the query sets of all other test classes; see Figure~\ref{fig:spurious_effect_on_gap_extreme} at Section~\ref{sec:appendix_supprious_correlation_strength} of the appendix. This maximizes background overlap across test tasks, removing background-specific cues for classification. 
% \begin{wrapfigure}[19]{r}{0.5\textwidth}
%     \vspace{-\intextsep}% kills top space
%     \centering
%     \includegraphics[width=0.5\textwidth]{figures/alpha_graphs/alpha_graph_proto_Hybrid_backbone.png}
%     \caption{{IID--OOD performance gap as a function of spurious-correlation strength.} We plot $\Delta(\alpha)$ for 1-shot and 5-shot settings. The gap increases with $\alpha$, indicating that stronger foreground--background correlations lead to larger \emph{OOD} degradation.}
%     \label{fig:alpha_gap}
%     \vspace{-\baselineskip}% kills bottom space
% \end{wrapfigure}
Across all methods, stronger correlations induce larger accuracy drops than the standard setting, indicating that few-shot methods rely more on background cues as their predictive power grows. Thus, degradation scales with spurious-correlation strength rather than distribution shift alone.

\begin{table}[!htbp]
    \centering
    \caption{Conv64 results on 1-shot and 5-shot tasks. $\ast$ refers to an additional attention module on top of the Conv64 model.}
    \label{tab:conv64}
    \adjustbox{max width=\textwidth}{
    \vspace{0.5em}
    \begin{tabular}{clcccccc}
        \toprule
        & & \multicolumn{3}{c}{1-shot} & \multicolumn{3}{c}{5-shot} \\
        \cmidrule(lr){3-5} \cmidrule(lr){6-8}
        Family & Method & Avg. \emph{IID} & Avg. \emph{OOD} & $\Delta$ & Avg. \emph{IID} & Avg. \emph{OOD} & $\Delta$ \\
        \midrule
        \multirow{4}{*}{Finetuning-based} & Baseline~\cite{baseline} (2019)       & 60.10 $\pm$ 2.01 & 53.43 $\pm$ 2.11 & 6.67   & 75.79 $\pm$ 1.01 & 63.14 $\pm$ 1.43 & 12.65 \\
        & Baseline++~\cite{baseline}  (2019)     & 55.30 $\pm$ 1.47 & 52.33 $\pm$ 2.61 & 2.97   & 65.20 $\pm$ 2.46 & 59.23 $\pm$ 2.43 & 5.97 \\
        & Meta-Baseline~\cite{meta-baseline} (2021)  & 61.05 $\pm$ 1.22 & 58.38 $\pm$ 0.37 & 2.67   & 74.01 $\pm$ 0.51 & 67.49 $\pm$ 2.21 & 6.52  \\
        & DiffKendal~\cite{zheng2023diffkendall} (2023)     & 58.55 $\pm$ 2.44 & 54.63 $\pm$ 0.80 & 3.93   & 74.17 $\pm$ 1.43 & 64.42 $\pm$ 1.10 & 9.75 \\
        \hdashline
        \multirow{6}{*}{Meta learning-based} 
        & MAML~\cite{maml} (2017)           & 52.14 $\pm$ 2.42 & 53.73 $\pm$ 1.46 & $\mathbf{-1.59}$ & 69.11 $\pm$ 0.87 & 64.87 $\pm$ 1.28 & 4.24 \\
        & LEO~\cite{rusu2018meta} (2019)            & 43.18 $\pm$ 1.99 & 41.75 $\pm$ 1.89 & 1.43   & 57.28 $\pm$ 1.54 & 52.93 $\pm$ 1.14 & 4.35 \\
        & R2D2~\cite{r2d2} (2019)           & 53.09 $\pm$ 6.09 & 51.87 $\pm$ 2.28 & 1.21   & 72.64 $\pm$ 3.01 & 67.19 $\pm$ 3.24 & 5.45 \\
        & ANIL~\cite{anil} (2020)           & 51.09 $\pm$ 3.90 & 49.23 $\pm$ 3.55 & 1.86   & 65.15 $\pm$ 0.33 & 59.75 $\pm$ 1.26 & 5.40 \\
        & BOIL~\cite{oh2020boil} (2020)           & 53.79 $\pm$ 1.08 & 51.99 $\pm$ 1.05 & 1.80   & 65.59 $\pm$ 0.95 & 60.36 $\pm$ 0.97 & 5.23 \\
        & METAL~\cite{baik2021metal} (2021)          & 57.27 $\pm$ 1.94 & 52.71 $\pm$ 1.56 & 4.56   & 71.29 $\pm$ 0.84 & 67.08 $\pm$ 2.70 & 4.21 \\
        \hdashline
        \multirow{7}{*}{Metric-based}
        & Proto~\cite{snell2017prototypical} (2017)          & 52.13 $\pm$ 2.13 & 49.83 $\pm$ 0.91 & 2.30   & 71.94 $\pm$ 3.17 & 61.17 $\pm$ 1.82 & 10.77 \\
        & RelationNet~\cite{relationnet} (2018)    & 59.22 $\pm$ 1.35 & 56.17 $\pm$ 1.74 & 3.05   & 74.76 $\pm$ 1.41 & 70.15 $\pm$ 4.22 & 4.61 \\
        & DN4~\cite{dn4} (2019)            & 63.08 $\pm$ 1.10 & 55.48 $\pm$ 3.84 & 7.60   & 80.58 $\pm$ 2.02 & \textbf{72.64 $\pm$ 2.90} & 7.94 \\
        
        & ADM~\cite{adm} (2020)            & 63.23 $\pm$ 3.06 & 56.31 $\pm$ 2.56 & 6.92   & \textbf{81.80 $\pm$ 0.90} & 69.36 $\pm$ 2.91 & 12.65 \\
        & ADM KL~\cite{adm} (2020)         & 60.09 $\pm$ 0.60 & 56.65 $\pm$ 2.60 & 3.44   & 80.99 $\pm$ 2.50 & 71.99 $\pm$ 2.71 & 9.00 \\
        & ATL-Net~\cite{atlnet} (2020)        & 60.19 $\pm$ 1.52 & 54.54 $\pm$ 2.39 & 5.65   & 80.53 $\pm$ 1.05 & 68.91 $\pm$ 0.56 & 11.62 \\
        & MCL~\cite{mcl} (2022)            & \textbf{64.60 $\pm$ 1.87} & \textbf{60.15 $\pm$ 1.36} & 4.45   & 76.72 $\pm$ 2.64 & 69.84 $\pm$ 1.52 & 6.88 \\
        \hdashline
        \multirow{3}{*}{Contrastive \footnotemark}
        & Proto + SimCLR + Conv64~\cite{simclr} (2020)      &  $45.55 \pm 1.43$ & $42.62 \pm 1.67$ & $2.93$ & $59.62 \pm 1.31$ & $53.68 \pm 1.85$ & $5.94$ \\
        & Proto + Contr~\cite{proto-contrastive} (2025)  &                    $58.55 \pm 0.82$ & $55.14 \pm 0.91$ & $3.41$ & $76.26 \pm 0.74$ & $67.59 \pm 0.88$ & $8.67$ \\
        & Proto + Contr ($\ast$)~\cite{proto-contrastive} (2025)               & $55.98 \pm 1.05$ & $54.23 \pm 1.12$ & $1.75$ & $74.03 \pm 0.96$ & $65.21 \pm 1.18$ & $8.82$ \\ 
        \hdashline
        \multirow{5}{*}{Transductive}
        & LaplacianShot~\cite{laplacianshot} (2020) & 41.33 $\pm$ 0.63 & 41.71 $\pm$ 0.69 & $-0.38$ & 57.97 $\pm$ 0.56 & 54.38 $\pm$ 0.55 & $3.59$ \\
        & BDCSPN~\cite{bdcspn} (2022) & 39.18 $\pm$ 0.54 & 38.91 $\pm$ 0.60 & 0.27 & 48.41 $\pm$ 0.48  & 47.43 $\pm$ 0.47 & \textbf{0.98} \\
        & PADDLE~\cite{paddle} (2022) & 43.26 $\pm$ 0.62 & 43.35 $\pm$ 0.64 & $-0.09$ & 56.61 $\pm$ 0.53 & 54.39 $\pm$ 0.53 & $2.22$ \\
        & Proto-LP~\cite{protolp} (2023)  & 43.61 $\pm$ 0.86 & 42.3 $\pm$ 0.8 & $1.31$ & 56.57 $\pm$ 0.67 & 54.56 $\pm$ 0.65 & $2.01$ \\
        & BPA~\cite{shalam2024balanced} (2024) & 53.54 $\pm$ 0.87 & 50.88 $\pm$ 1.34 & 2.66 & 72.30 $\pm$ 0.66 & 62.60 $\pm$ 0.65 & 9.70 \\
        & ECPE~\cite{ecpe} (2026)  & 45.27 $\pm$ 0.80 & 44.59 $\pm$ 0.72 & $0.68$ & 58.15 $\pm$ 0.56 & 56.67 $\pm$ 0.55 & $1.48$ \\

        \bottomrule
    \end{tabular}}
\end{table}

\afterpage{\footnotetext{See Section~\ref{sec:contrastive_experiments} in the Appendix for additional information concerning the contrastive approaches.}}

\subsubsection{Embedding Geometry and Spurious-Correlation Effects}
\label{sec:embedding_analysis}

To understand how spurious backgrounds affect few-shot audio classification, we analyze embedding geometry under \emph{IID} and \emph{OOD} settings. Figures~\ref{fig:embeddings_tsne_10}--\ref{fig:embeddings_tsne_1} show t-SNE projections of support and query embeddings (see Section~\ref{sec:appendix_embeddings_of_FSL}), and Figure~\ref{fig:alpha_gap} plots $\Delta$
against the spurious-correlation strength $\alpha$
(Eq.~\eqref{eq:mixing}).

\textbf{Embedding structure under \emph{IID} vs.\ \emph{OOD}.}  
Across methods, \emph{IID} episodes show tighter class clusters and better alignment between support and query samples. Under \emph{OOD} background shifts, query embeddings drift away from their corresponding support clusters, increasing overlap between classes and making nearest-prototype decisions less reliable. 
This geometric mismatch provides an interpretable explanation for the observed accuracy drop under spurious-correlation shifts.

\textbf{Effect of spurious-correlation strength.}  
Figure~\ref{fig:alpha_gap} shows that the performance gap $\Delta(\alpha) = \text{Acc}_{\text{IID}} - \text{Acc}_{\text{OOD}}$ grows with $\alpha$ in both $1$-shot and $5$-shot regimes. Interestingly, the gap can be larger in the higher-shot setting, suggesting that additional support examples may reinforce background-dependent representations rather than improving robustness when correlations are strong as depicted In Figure~\ref{fig:more_shots_effect}; See section~\ref{sec:appendix_more_shots_effect} at the appendix.

\subsubsection{On The Distribution of SpurAudio}
\label{sec:distribution_of_spuraudio}
% \loaycomment{title of subsubsection should be changead}
We show that the distribution underlying SpurAudio forms only a sub-distribution of the broader pool of sound events encountered in the \say{wild}, and that these events are situated within diverse acoustic contexts. Our dataset incorporates standard audio mixing at realistic signal-to-noise ratios, without adding synthetic artifacts or adversarial perturbations. In Section~\ref{sec:appendix_distribution_spuraudio}, we empirically show that the resulting sound events are close to those found in FSD50K dataset~\cite{fonseca2021fsd50k}. In particular, we find that embeddings of sound events from SpurAudio samples preserve the structure induced by embeddings of real, \say{in-the-wild} sound events that are semantically similar (in terms of class labels), while also remaining in close proximity to those real events in the embedding space; see Figure~\ref{fig:distribution_analysis}. %appendix.

\subsubsection{Why Spurious Correlation Harms Few-Shot Audio Classification}
\label{sec:why_spurious_harms}

Spurious background correlations harm few-shot audio classification by biasing embedding-based similarity, even when foreground semantics remain unchanged. Through controlled IID--OOD evaluations, head--backbone replacement, and background perturbations, we identify a recurring failure mechanism across architectures and algorithm families (Figures~\ref{fig:IID_backbone_algorithm_visualization_conv64f}--\ref{fig:OOD_backbone_algorithm_visualization_resnet12}, Tables~\ref{tab:conv64}--\ref{tab:resnet18}, Appendix~\ref{sec:appendix_extended_analysis}).

% Spurious background correlations harm few-shot audio classification because modern embedding models encode background variation in a way that systematically biases similarity-based inference, even when foreground semantics remain unchanged. Through controlled IID--OOD evaluations, head--backbone replacement, and explicit background perturbations, we identify a single, recurring failure mechanism across architectures and algorithm families (Figures~\ref{fig:IID_backbone_algorithm_visualization_conv64f}--\ref{fig:OOD_backbone_algorithm_visualization_resnet12}, Tables~\ref{tab:conv64}--\ref{tab:resnet18}, Appendix~\ref{sec:appendix_extended_analysis}).

\begin{table}[!htbp]
    \centering
    \caption{ResNet12 results on 1-shot and 5-shot tasks. $\ast$ refers to an additional attention module on top of the ResNet12 model.}
    \label{tab:resnet12}
    \adjustbox{max width=\textwidth}{
    \vspace{0.5em}
    \begin{tabular}{clcccccc}
        \toprule
        & & \multicolumn{3}{c}{1-shot} & \multicolumn{3}{c}{5-shot} \\
        \cmidrule(lr){3-5} \cmidrule(lr){6-8}
        Family & Method & Avg. \emph{IID} & Avg. \emph{OOD} & $\Delta$ & Avg. \emph{IID} & Avg. \emph{OOD} & $\Delta$ \\
        \midrule
        \multirow{3}{*}{Finetuning-based} & Baseline~\cite{baseline} (2019) & 59.45 $\pm$ 0.47 & 56.58 $\pm$ 1.47 & 2.87 & 79.42 $\pm$ 0.76 & 68.58 $\pm$ 1.16 & 10.84 \\
        & Baseline++~\cite{baseline}  (2019) & 57.31 $\pm$ 2.68 & 55.57 $\pm$ 1.38 & 1.74 & 76.18 $\pm$ 1.12 & 64.97 $\pm$ 2.63 & 11.21 \\
        & Meta-Baseline~\cite{meta-baseline} (2021) & 59.98 $\pm$ 4.70 & 57.07 $\pm$ 2.69 & 2.91 & 72.87 $\pm$ 1.38 & 67.34 $\pm$ 0.90 & 5.53 \\ \hdashline
        \multirow{3}{*}{Meta learning-based}  
        & LEO~\cite{rusu2018meta} (2019) & 42.97 $\pm$ 1.36 & 41.66 $\pm$ 1.80 & 1.31 & 53.61 $\pm$ 1.36 & 49.96 $\pm$ 1.12 & 3.65 \\
        & R2D2~\cite{r2d2} (2019) & 59.36 $\pm$ 2.52 & 55.63 $\pm$ 1.57 & 3.74 & 79.59 $\pm$ 1.03 & 66.21 $\pm$ 3.20 & 13.38 \\ 
        & ANIL~\cite{anil} (2020) & 56.84 $\pm$ 2.33 & 52.44 $\pm$ 0.48 & 4.40 & 76.29 $\pm$ 1.18 & 66.90 $\pm$ 1.10 & 9.40 \\ 
        \hdashline
        \multirow{6}{*}{Metric-based}
        & Proto~\cite{snell2017prototypical} (2017) & 59.49 $\pm$ 2.14 & 55.52 $\pm$ 3.88 & 3.97 & 79.34 $\pm$ 1.15 & 68.89 $\pm$ 0.79 & 10.45 \\
        & DN4~\cite{dn4} (2019) & 66.55 $\pm$ 0.95 & \textbf{64.43 $\pm$ 2.93} & 2.12 & \textbf{84.14 $\pm$ 1.05} & \textbf{78.42 $\pm$ 2.90} & 5.72 \\
        & ADM~\cite{adm} (2020) & 62.10 $\pm$ 6.80 & 54.25 $\pm$ 8.45 & 7.85 & 83.07 $\pm$ 0.72 & 75.46 $\pm$ 1.81 & 7.61 \\
        & ATL-Net~\cite{atlnet} (2020) & \textbf{66.93 $\pm$ 1.69} & 64.25 $\pm$ 0.88 & 2.68 & 82.54 $\pm$ 1.05 & 72.50 $\pm$ 1.16 & 10.03 \\
        & FRN~\cite{frn} (2021) & 64.67 $\pm$ 1.76 & 61.82 $\pm$ 1.83 & 2.86 & 71.68 $\pm$ 2.47 & 63.37 $\pm$ 9.14 & 8.30 \\
        & DeepDBC~\cite{deepbdc} (2022) & 64.71 $\pm$ 1.47 & 61.33 $\pm$ 1.78 & 3.38 & 80.68 $\pm$ 2.22 & 73.35 $\pm$ 1.08 & 7.33 \\
        & HELA-VFA~\cite{lee2024hela} (2024)$\ast$ & 48.35 $\pm$ 0.68 & 46.49 $\pm$ 0.62 & 1.86 & 64.26 $\pm$ 0.59 & 55.57 $\pm$ 0.58 & 8.70 \\
        \hdashline
        \multirow{5}{*}{Transductive}
        & LaplacianShot~\cite{laplacianshot} (2020) & 46.74 $\pm$ 0.67 & 46.13 $\pm$ 0.62 & 0.61 & 59.23 $\pm$ 0.59 & 55.06 $\pm$ 0.56 & 4.17 \\
        & BDCSPN~\cite{bdcspn} (2022) & 46.92 $\pm$ 0.69 & 45.39 $\pm$ 0.65 & 1.53 & 56.53 $\pm$ 0.58 & 53.29 $\pm$ 0.53 & 3.24 \\
        & PADDLE~\cite{paddle} (2022) & 49.33 $\pm$ 0.67 & 48.05 $\pm$ 0.59 & 1.28 & 59.66 $\pm$ 0.57 & 56.49 $\pm$ 0.54 & 3.17 \\
        & Proto-LP~\cite{protolp} (2023)  & 49.50 $\pm $ 0.83 & 49.30 $\pm $ 0.65 & $2.00$ & 60.84 $\pm $ 0.69 & 57.69 $\pm $ 0.66 & \textbf{3.15}\\
        & BPA~\cite{shalam2024balanced} (2024) & 59.88 $\pm$ 0.72 & 52.82 $\pm$ 0.65 & 7.06 & 78.70 $\pm$ 0.62 & 73.30 $\pm$ 0.60 & 5.40 \\
        & ECPE~\cite{ecpe} (2026) & 51.39 $\pm$ 0.76 & 50.91 $\pm$ 0.71 & \textbf{0.48} & 62.56 $\pm$ 0.58 & 58.83 $\pm$ 0.60 & 3.73 \\
        \bottomrule
    \end{tabular}}
\end{table}

\textbf{Background effects on representation geometry.}
Background variation primarily affects \emph{feature magnitude} rather than semantic direction. Figure~\ref{fig:geometry_grid} shows that background perturbations consistently contract embedding norms while preserving class-aligned structure. In addition, this contraction is consistent across backbones and training paradigms, indicating that magnitude sensitivity is a fundamental property of current audio embeddings rather than an artifact of a specific model.
Moreover, this magnitude sensitivity is not localized to the final embedding layer: stage-wise hooking experiments on Conv64F and ResNet12 (Appendix~\ref{sec:where_is_spurious_correlation_encoded}, Tables~\ref{tab:conv_stage_results}--\ref{tab:resnet12_results}) reveal that the \emph{IID}--\emph{OOD} gap is present from the first convolutional or residual block, indicating that spurious background correlations are encoded throughout the feature hierarchy rather than emerging only at the final embedding.

\begin{figure}[!htbp]
    \centering
    \includegraphics[width=\textwidth]{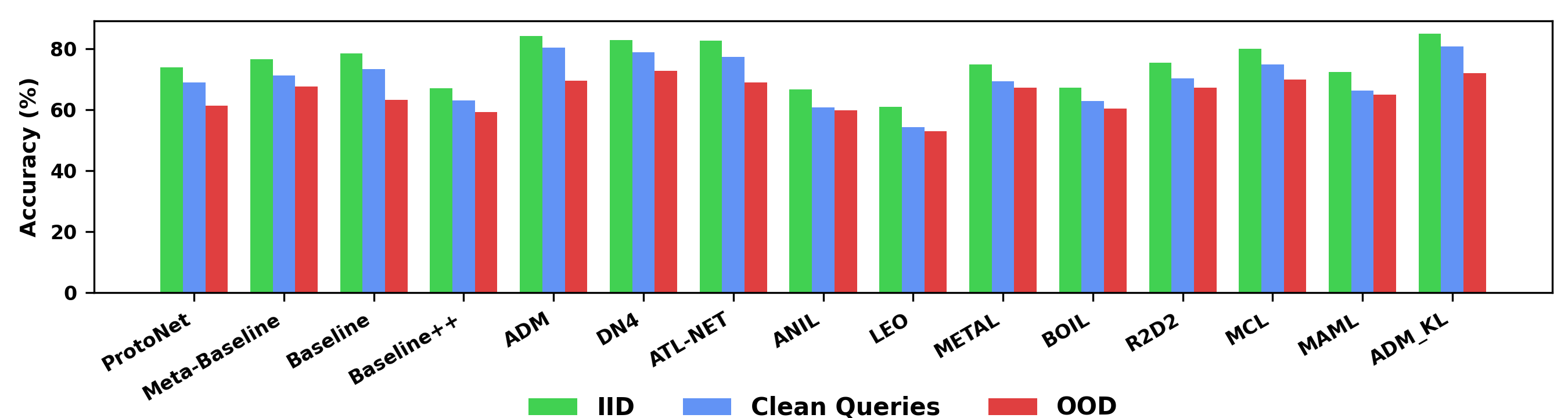}
%  \caption{{Impact of Spurious Background Correlations on Query Evaluation.} 
% We analyze few-shot accuracy across four query composition settings to isolate the effect of background noise. 
% (1) {Green (IID):} Standard evaluation with matched background distributions between supports and queries. 
% (2) {Blue (Clean):} IID support paired with clean, background-free queries. 
% (3) {Red (OOD):} Evaluation with mismatched (spurious) backgrounds. 
% Results indicate that removing background noise (Blue) improves accuracy over the OOD setting but remains below the IID baseline, confirming that models exploit background correlations as a shortcut. All results are averaged across 3 seeds.}
%     \label{fig:clean_queries_comparison}
\caption{{Impact of Spurious Background Correlations on Query Evaluation.}
We compare few-shot accuracy under three query settings:
{Green (IID):} matched support-query backgrounds;
{Blue (Clean):} IID support with background-free queries;
and {Red (OOD):} mismatched spurious backgrounds.
Clean queries improve over OOD but remain below IID, indicating that models rely on background shortcuts.
Results are averaged over 3 seeds.}
\label{fig:clean_queries_comparison}
\end{figure}

\textbf{Failure of global, unnormalized similarity.}
Many few-shot methods implicitly treat representation strength as a proxy for semantic similarity. Approaches based on dot-product similarity, Euclidean prototypes, or global pooling directly entangle background-dependent activation magnitude with class identity. While this bias is largely invisible under IID conditions, background shifts cause systematic overestimation of dissimilarity, leading to large $\Delta$ (e.g., Baseline, ProtoNet, ATL-NET in Figure~\ref{fig:clean_queries_comparison}).

\textbf{Limits of feature normalization.}
Cosine-based classifiers (Baseline++, Meta-Baseline) mitigate magnitude sensitivity by normalizing features prior to comparison. This normalization explains their improved stability on shallow backbones where background variation is primarily radial. 

\textbf{ResNet backbones and background leakage.} The background information for deeper ResNet e.g., ResNet12, becomes directionally entangled with foreground features, leaving substantial $\Delta$ even after normalization (Table~\ref{tab:resnet18}). This shows that normalization alone is insufficient and motivates inference mechanisms that either localize matching (e.g., DN4) or adapt representations at test time.

\textbf{Robust inference mechanisms.}
Up to this point, we have found that methods that more strongly suppress global similarity exhibit consistently greater robustness. Local-descriptor approaches (DN4) bypass global aggregation entirely, while adaptive meta-learning methods (e.g., MAML, BOIL) use the support set to reshape the representations and downweight background-dependent features. By contrast, frozen-backbone approaches (e.g., ANIL, R2D2) preserve these background shortcuts, resulting in larger and more persistent OOD performance gaps (Appendix~\ref{sec:appendix_head_backbone_replacement_FT}, \ref{sec:appendix_iid_ood_background_perturbation_analysis_meta}).

\subsection{Large Audio Models}
\label{sec:large_model_experiments}
A natural objection is that spurious background reliance may reflect the limited capacity of the CNN backbones used so far. We address this by stress-testing our findings on five state-of-the-art large audio encoders spanning contrastive (CLAP~\cite{laionclap2023}), supervised transformer (AST~\cite{gong2021ast}), masked autoencoding (AudioMAE-AS20K~\cite{huang2022masked}), and instruction-tuned multimodal (Qwen2-Audio-7B~\cite{Qwen2-Audio}), each used as a frozen encoder paired with eleven non-backbone-dependent heads ranging from classical baselines~\cite{snell2017prototypical, baseline, dn4} to modern transductive and label-propagation methods~\cite{protolp, shalam2024balanced, lee2024hela, laplacianshot, bdcspn, paddle, ecpe}; For additional results using $1$ shot \emph{FSC} concerning large audio models and transformer based models, we refer the reader to Section~\ref{sec:appendix_more_results_audio_models}.

\begin{experimenttakeawaybox}{}
From a 4-layer CNN to a 7B-parameter audio LLM (Tables~\ref{tab:conv64},~\ref{tab:resnet12},~\ref{tab:results_large_models_5_shots_inverted},~\ref{tab:resnet18}), three findings emerge:
\begin{enumerate}[label=(\roman*)]
\item \textbf{Every \emph{FSC} method suffers an \emph{IID}--\emph{OOD} gap.} The gap appears in every method family, at every backbone size, and at every shot count, and grows as $\alpha$ increases (Figure~\ref{fig:alpha_gap}). It is therefore not caused by any single architecture.
\item \textbf{Larger pre-trained encoders are not robust.} While such encoders can reach $\approx 96\%$ \emph{IID} accuracy on corpora that \uline{cover} SpurAudio's classes, they still suffer gaps under standard heads. This means {background shifts perturb embedding magnitudes while leaving angular structure largely preserved}, and standard heads read magnitude changes as class differences.
\item \textbf{The inference head matters more than the encoder.} Since background shifts perturb magnitude but not direction, heads that ignore magnitude (cosine similarity, query-set graphs, neighborhood propagation) shrink the \emph{IID}--\emph{OOD} gap by an order of magnitude, while heads built on absolute distance inherit the shift. Results confirm this: transductive heads (Proto-LP~\cite{protolp}, BD-CSPN~\cite{bdcspn}, ECPE~\cite{ecpe}) shrink $\Delta$ by an order of magnitude (Proto-LP: $1.5$--$2.6\%$ on CLAP, AST, AudioMAE, Qwen2-Audio), while magnitude-sensitive heads (Hela-VFA, BPA) hit double-digit gaps even with the strongest encoders. While the same transductive heads result in small $\Delta$ on Conv64 and ResNet12, their absolute accuracy on these backbones lags the best metric-based methods. This is because transductive inference relies on the cluster assumption~\cite{chapelle2009semi} and on a well-clustered query manifold~\cite{li2020fewer,ziko2021transductive}, a property that emerges in large pretrained encoders. The large IID gap between Conv64 and CLAP under a fixed ProtoNet head (Tables~\ref{tab:conv64} and~\ref{tab:results_large_models_5_shots_inverted}) emphasizes that large pretrained encoders produce sharper class clusters that transductive heads can exploit.
\end{enumerate}
\end{experimenttakeawaybox}

\begin{table}[!htbp]
\centering
\caption{5-shot IID and OOD accuracy (\%) and gaps
across few-shot methods on large audio models.}
\resizebox{\textwidth}{!}{%
\begin{tabular}{l|ccc|ccc|ccc|ccc}
\hline
\multirow{2}{*}{\textbf{Method}}
  & \multicolumn{3}{c|}{\textbf{CLAP}}
  & \multicolumn{3}{c|}{\textbf{AudioMAE-AS20K}}
  & \multicolumn{3}{c|}{\textbf{AST}}
  & \multicolumn{3}{c}{\textbf{Qwen2-Audio-7B}} \\
 & IID & OOD & $\Delta$ & IID & OOD & $\Delta$ & IID & OOD & $\Delta$ & IID & OOD & $\Delta$ \\
\hline
\textbf{Proto}~\cite{snell2017prototypical} (2017)
& $95.30 \pm 0.26$ & $86.98 \pm 0.54$ & 8.32
& $93.05 \pm 0.32$ & $80.02 \pm 0.55$ & 13.03
& $95.88 \pm 0.25$ & $87.42 \pm 0.56$ & 8.46
& $94.72 \pm 0.25$ & $84.99 \pm 0.60$ & 9.73 \\
\textbf{Baseline}~\cite{baseline} (2019)
& $95.50 \pm 0.26$ & $87.03 \pm 0.55$ & 8.47
& $93.15 \pm 0.32$ & $81.00 \pm 0.57$ & 12.15
& $95.84 \pm 0.25$ & $88.20 \pm 0.55$ & 7.64
& $93.83 \pm 0.31$ & $84.48 \pm 0.61$ & 9.35 \\
\textbf{Baseline++}~\cite{baseline} (2019)
& $95.32 \pm 0.26$ & $86.61 \pm 0.56$ & 8.71
& $94.36 \pm 0.29$ & $82.12 \pm 0.53$ & 12.24
& $96.43 \pm 0.22$ & $87.29 \pm 0.56$ & 9.14
& $94.91 \pm 0.27$ & $84.52 \pm 0.60$ & 10.39 \\
\textbf{DN4}~\cite{dn4} (2019)
& $94.96 \pm 0.28$ & $85.91 \pm 0.57$ & 9.05
& $91.33 \pm 0.36$ & $78.68 \pm 0.56$ & 12.65
& $95.23 \pm 0.26$ & $87.09 \pm 0.57$ & 8.14
& $92.67 \pm 0.33$ & $82.98 \pm 0.62$ & 9.69 \\
\textbf{LaplacianShot}~\cite{laplacianshot} (2020)
& $95.47 \pm 0.28$ & $90.45 \pm 0.58$ & 5.02
& $95.27 \pm 0.29$ & $86.60 \pm 0.54$ & 8.67
& $95.39 \pm 0.28$ & $86.81 \pm 0.63$ & 8.58
& $94.19 \pm 0.27$ & $86.57 \pm 0.58$ & 7.62 \\
\textbf{BD-CSPN}~\cite{bdcspn} (2020)
& $95.76 \pm 0.29$ & $93.14 \pm 0.50$ & 2.62
& $96.30 \pm 0.27$ & $92.29 \pm 0.54$ & 4.01
& $96.55 \pm 0.25$ & $92.00 \pm 0.52$ & 4.55
& $95.81 \pm 0.25$ & $88.85 \pm 0.61$ & 6.96 \\
\textbf{PADDLE}~\cite{paddle} (2022)
& $96.18 \pm 0.26$ & $90.95 \pm 0.53$ & 5.23
& $96.43 \pm 0.24$ & $90.17 \pm 0.52$ & 6.26
& $96.51 \pm 0.22$ & $89.00 \pm 0.54$ & 7.51
& $95.36 \pm 0.24$ & $85.69 \pm 0.59$ & 9.67 \\
\textbf{Proto-LP}~\cite{protolp} (2023)
& $\mathbf{97.43 \pm 0.25}$ & $\mathbf{95.68 \pm 0.44}$ & $\mathbf{1.75}$
& $\mathbf{97.54 \pm 0.25}$ & $\mathbf{95.14 \pm 0.44}$ & $\mathbf{2.40}$
& $\mathbf{97.64 \pm 0.24}$ & $\mathbf{95.08 \pm 0.52}$ & $\mathbf{2.56}$
& $\mathbf{98.22 \pm 0.17}$ & $\mathbf{96.73 \pm 0.29}$ & $\mathbf{1.49}$ \\
\textbf{BPA}~\cite{shalam2024balanced} (2024)
& $96.38 \pm 0.23$ & $86.89 \pm 0.57$ & 9.49
& $96.14 \pm 0.26$ & $86.05 \pm 0.55$ & 10.09
& $96.22 \pm 0.22$ & $84.45 \pm 0.54$ & 11.77
& $95.80 \pm 0.22$ & $83.50 \pm 0.58$ & 12.30 \\
\textbf{Hela-VFA}~\cite{lee2024hela} (2024)
& $94.83 \pm 0.24$ & $81.18 \pm 0.50$ & 13.65
& $93.61 \pm 0.29$ & $75.03 \pm 0.54$ & 18.58
& $93.69 \pm 0.27$ & $76.50 \pm 0.55$ & 17.19
& $94.68 \pm 0.24$ & $80.07 \pm 0.51$ & 14.61 \\
\textbf{ECPE}~\cite{ecpe} (2026)
& $96.65 \pm 0.24$ & $92.96 \pm 0.50$ & 3.69
& $96.65 \pm 0.24$ & $92.58 \pm 0.48$ & 4.07
& $96.30 \pm 0.24$ & $90.63 \pm 0.56$ & 5.67
& $96.04 \pm 0.23$ & $88.16 \pm 0.57$ & 7.88 \\
\hline
\end{tabular}%
}
\label{tab:results_large_models_5_shots_inverted}
\end{table}

\section{Conclusions and Future Work}
\label{sec:conclusions}

We introduced \textbf{SpurAudio}, a benchmark for studying spurious foreground--background correlations in few-shot audio classification, and showed that the resulting \emph{IID}--\emph{OOD} gap is consistent across model families, backbone scales, and large pretrained encoders. Our geometric analysis traces this failure to inference-time interactions with background cues: background variation primarily perturbs embedding \emph{magnitudes} while leaving angular structure intact, explaining why magnitude-sensitive heads degrade sharply while transductive, relational ones remain robust. These findings open a new avenue for few-shot research centered on embedding geometry, including magnitude-aware objectives, adaptive similarity metrics, and multimodal extensions of SpurAudio.

\begin{ack}
% Use unnumbered first level headings for the acknowledgments. All acknowledgments
% go at the end of the paper before the list of references. Moreover, you are required to declare

This research was funded by the Ministry of Science, Research and the Arts Baden-Wuerttemberg in the Artificial Intelligence Software Academy (AISA). L.\ Mualem also acknowledge the support of the Stuttgart Center for Simulation Science (SimTech) and thank the International Max Planck Research School for Intelligent Systems (IMPRS-IS) for support. L.\ Mualem was supported by a postdoctoral scholarship from the Planning and Budgeting Committee (PBC) of the Council for Higher Education in Israel. L.\ Mualem gratefully acknowledge the computing time provided on the high-performance computer HoreKa by the National High-Performance Computing Center at KIT (NHR@KIT). This center is jointly supported by the Federal Ministry of Education and Research and the Ministry of Science, Research and the Arts of Baden-Württemberg, as part of the National High-Performance Computing (NHR) joint funding program (https://www.nhr-verein.de/en/our-partners). HoreKa is partly funded by the German Research Foundation (DFG). 

\end{ack}

\bibliographystyle{plain}
\bibliography{main}

\newpage
\appendix
\onecolumn
% \section*{Supplementary Material}

% \vspace*{2\baselineskip}   % extra space before the block

\noindent
\rule{\textwidth}{1pt}   % thin line above (adjust 0.4pt–1pt for thickness)

\vspace{1ex}

\begin{center}
{\Large \bfseries Supplementary Material Contents}
\end{center}

\vspace{1ex}

\noindent
\rule{\textwidth}{1pt}   % thin line below

\vspace{2\baselineskip}

% === Your clickable contents list starts here ===
% Edit the entries to match your actual \section / \subsection titles & labels

{\Large \bfseries Appendix Contents}

\begin{enumerate}[label=\Alph*,leftmargin=*,align=left]
  \item \textbf{Experimental Setup} \dotfill \hyperlink{sec:experimental_setup}{\pageref{sec:experimental_setup}}
  \item \textbf{Foreground--Background Mapping} \dotfill \hyperlink{sec:appendix_FG_BG_mappiing}{\pageref{sec:appendix_FG_BG_mappiing}}

  \item \textbf{Embeddings of different \emph{FSL} Methods} \dotfill \hyperlink{sec:appendix_embeddings_of_FSL}{\pageref{sec:appendix_embeddings_of_FSL}}

  \item \textbf{Effect of Spurious Correlation Strength} \dotfill \hyperlink{sec:appendix_supprious_correlation_strength}{\pageref{sec:appendix_supprious_correlation_strength}}

  \item \textbf{Visualization of the Results in Table~\ref{tab:conv64}} \dotfill \hyperlink{sec:appendix_visualization_of_table1}{\pageref{sec:appendix_visualization_of_table1}}

  \item \textbf{ \emph{FSL} with ResNet18 as Embedding Backbone} \dotfill \hyperlink{sec:appendix_renset18_iid_ood_table_results}{\pageref{sec:appendix_renset18_iid_ood_table_results}} 

  \item \textbf{Analysis of Shortcut Reliance via Representation Geometry and Background Perturbations} 
  \dotfill \hyperlink{sec:appendix_extended_analysis}{\pageref{sec:appendix_extended_analysis}}
    \begin{enumerate}[label=\Alph{enumi}.\arabic*,leftmargin=*,align=left]
      \item Fine-Tuning Based \emph{FSC} \dotfill \hyperlink{sec:appendix_Fine-Tuning_Based}{\pageref{sec:appendix_Fine-Tuning_Based}}
      \begin{enumerate} [label=\Alph{enumi}.1.\arabic*,leftmargin=*,align=left]
          \item Head--Backbone Replacement Analysis \dotfill \hyperlink{sec:appendix_head_backbone_replacement_FT}{\pageref{sec:appendix_head_backbone_replacement_FT}}
          \item IID--OOD Background Perturbation Analysis \dotfill \hyperlink{sec:appendix_iid_ood_background_perturbation_analysis_FT}{\pageref{sec:appendix_iid_ood_background_perturbation_analysis_FT}}
          \item Linking Representation Geometry to Shortcut Exploitation \dotfill \hyperlink{sec:appendix_linking_representation_geometry_FT}{\pageref{sec:appendix_linking_representation_geometry_FT}}
      \end{enumerate}
      \item Meta-Learning-Based \emph{FSC} \dotfill \hyperlink{sec:appendix_Meta-Learning_Based}{\pageref{sec:appendix_Meta-Learning_Based}}
      \begin{enumerate}[label=\Alph{enumi}.2.\arabic*,leftmargin=*,align=left]
          \item Head--Backbone Replacement Analysis (IID Baseline) \dotfill \hyperlink{sec:appendix_head_backbone_replacement_analysis_meta}{\pageref{sec:appendix_head_backbone_replacement_analysis_meta}}
          \item IID--OOD Background Perturbation Analysis \dotfill \hyperlink{sec:appendix_iid_ood_background_perturbation_analysis_meta}{\pageref{sec:appendix_iid_ood_background_perturbation_analysis_meta}}
          \item Linking Representation Geometry to Shortcut Exploitation \dotfill \hyperlink{sec:appendix_linking_representation_geometry_meta}{\pageref{sec:appendix_linking_representation_geometry_meta}}
          \item OOD Heatmap Comparison: Shifts in Robustness \dotfill \hyperlink{sec:appendix_ood_heat_map_comparison}{\pageref{sec:appendix_ood_heat_map_comparison}}
      \end{enumerate}
      \item Metric-Based \emph{FSC} \dotfill \hyperlink{sec:appendix_Metric_Based}{\pageref{sec:appendix_Metric_Based}}
      \begin{enumerate}[label=\Alph{enumi}.3.\arabic*,leftmargin=*,align=left]
          \item Head--Backbone Replacement Analysis (IID Baseline) \dotfill \hyperlink{sec:appendix_head_backbone_replacement_analysis_metric}{\pageref{sec:appendix_head_backbone_replacement_analysis_metric}}
          \item IID--OOD Background Perturbation Analysis \dotfill \hyperlink{sec:appendix_iid_ood_background_perturbation_analysis_metric}{\pageref{sec:appendix_iid_ood_background_perturbation_analysis_metric}}
          \item Linking Representation Geometry to Shortcut Exploitation \dotfill \hyperlink{sec:appendix_linking_representation_geometry_metric}{\pageref{sec:appendix_linking_representation_geometry_metric}}
          \item OOD Heatmap Comparison: Shifts in Robustness \dotfill \hyperlink{sec:ood_heatmap_comparison_metric}{\pageref{sec:ood_heatmap_comparison_metric}}
      \end{enumerate}
      \item Cross-Family Analysis of Few-Shot Learning Algorithms (IID Setting) \dotfill \hyperlink{sec:appendix_Cross-Family_Analysis_of_Few-Shot_Learning_Algorithms_IID}{\pageref{sec:appendix_Cross-Family_Analysis_of_Few-Shot_Learning_Algorithms_IID}}
      \item Cross-Family Analysis of Few-Shot Learning Algorithms (OOD Setting) \dotfill \hyperlink{sec:appendix_Cross-Family_Analysis_of_Few-Shot_Learning_Algorithms_OOD}{\pageref{sec:appendix_Cross-Family_Analysis_of_Few-Shot_Learning_Algorithms_OOD}}
      \item Cross-Family Analysis of Few-Shot Learning Algorithms in Deeper Architectures \dotfill \hyperlink{sec:appendix_Cross-Family_Analysis_of_Few-Shot_Learning_Deeper_Algorithms}{\pageref{sec:appendix_Cross-Family_Analysis_of_Few-Shot_Learning_Deeper_Algorithms}}
      % \item Concerning ResNet18 embedding model \dotfill \hyperlink{sec:appendix_embedding_model_algorithm_confusion_matrix_resnet18}{\pageref{sec:appendix_embedding_model_algorithm_confusion_matrix_resnet18}}
      % etc.
    \end{enumerate}

  \item \textbf{Geometric Disentanglement of Background Correlations} \dotfill \hyperlink{sec:appendix_geometry_analysis}{\pageref{sec:appendix_geometry_analysis}}
    \begin{enumerate}[label=\Alph{enumi}.\arabic*,leftmargin=*,align=left]
          \item Methodology: Radial-Angular Decomposition \dotfill \hyperlink{sec:appendix_radial_angular_decoposition}{\pageref{sec:appendix_radial_angular_decoposition}}
    \item Empirical Observation: The "Magnitude Contraction" Phenomenon \dotfill \hyperlink{sec:appendix_magnitude_contraction}{\pageref{sec:appendix_magnitude_contraction}}
    \item Implication for Few-Shot Metric Learning \dotfill \hyperlink{sec:appendix_implication_for_fsc_metric}{\pageref{sec:appendix_implication_for_fsc_metric}}
    \end{enumerate}
  \item \textbf{Where in the Network Are Spurious Correlations Encoded?} \dotfill \hyperlink{sec:where_is_spurious_correlation_encoded}{\pageref{sec:where_is_spurious_correlation_encoded}}
  
  \item \textbf{On the Distribution of SpurAudio} \dotfill \hyperlink{sec:appendix_distribution_spuraudio}{\pageref{sec:appendix_distribution_spuraudio}}
    \begin{enumerate}
    [label=\Alph{enumi}.\arabic*,leftmargin=*,align=left]
        \item Methodology and Metrics \dotfill \hyperlink{sec:appendix_methodology_and_metrics}{\pageref{sec:appendix_methodology_and_metrics}}
        \item Analysis of Results \dotfill \hyperlink{sec:appendix_analysis_distribution}{\pageref{sec:appendix_analysis_distribution}}
        \begin{enumerate}
    [label=H.2.\arabic*,leftmargin=*,align=left]
    \item Semantic Alignment \dotfill \hyperlink{sec:appendix_semantic_alignment}{\pageref{sec:appendix_semantic_alignment}}
    \item Distributional and Structural Integrity \dotfill \hyperlink{sec:appendix_distritional_structural_integrity}{\pageref{sec:appendix_distritional_structural_integrity}}
    \end{enumerate}
    \end{enumerate}

 \item \textbf{More Results on Large-Audio Models} \dotfill \hyperlink{sec:appendix_more_results_audio_models}{\pageref{sec:appendix_more_results_audio_models}}
    
  \item \textbf{Classification Accuracy as a Function of Number Of Shots} \dotfill
    \hyperlink{sec:appendix_more_shots_effect}{\pageref{sec:appendix_more_shots_effect}}
  % \item \textbf{Additional Experiments} \dotfill \hyperlink{sec:large_model_experiments}{\pageref{sec:large_model_experiments}}

  \item \textbf{Additional Contrastive Learning Experiments} \dotfill
    \hyperlink{sec:contrastive_experiments}{\pageref{sec:contrastive_experiments}}

    \item \textbf{Limitations} \dotfill \hyperlink{sec:limitations}{\pageref{sec:limitations}}

  % Add more top-level items as needed
\end{enumerate}

\vspace{3\baselineskip}

\clearpage

\section{Experimental Setup}
\label{sec:experimental_setup}
To study spurious correlations in few-shot settings, we consider four main factors that inform the design of our benchmark: \begin{enumerate*}[label=(\roman*)]
    \item Episodic sampling,
    \item Audio preprocessing,
    \item Backbone architectures, and 
    \item \emph{FSL} methodologies.
\end{enumerate*}

\textbf{Episode sampling.}
We follow the standard $N$-way $K$-shot protocol~\cite{vinyals2016matching} with $N{=}5$, $K{\in}\{1,5\}$, and $10$ query samples per class, sampling between $600$ and $2000$ training tasks and $1000$ test tasks depending on the algorithm setting.

\textbf{Audio preprocessing.}
We adopt the preprocessing strategy of~\cite{Heggan2022MetaAudio}. All sound events are resampled to $16$~kHz and trimmed or repeated to a fixed duration of $T=5$~s. Each waveform is converted into a mel spectrogram using a $1024$-point short-time Fourier transform with a hop size of $512$ samples and $128$ mel frequency bands. Power spectrograms are computed and transformed to the logarithmic (decibel) scale to compress the dynamic range.

\textbf{Embedding architectures.}
Few-shot learning methods require a backbone that maps the mel spectrogram into feature vectors or descriptor sets. We employ architectures used in \emph{MetaCoCo}~\cite{Zhang2024MetaCoCo} and LibFewShot~\cite{li2023libfewshot}, specifically Conv-64F, ResNet-12, and ResNet-18, covering both shallow and deep designs.

\textbf{Few-shot learning methodologies.}
We evaluate representative approaches across multiple methodological families:
\begin{itemize}
    \item \textbf{Fine-tuning-based:} Baseline and Baseline++~\cite{baseline}, and Meta-Baseline~\cite{meta-baseline}, combining standard supervised pretraining with episodic adaptation.
    
    \item \textbf{Gradient-based meta-learning:} MAML~\cite{finn2017model} and variants including ANIL~\cite{anil}, BOIL~\cite{oh2020boil}, R2D2~\cite{r2d2}, METAL~\cite{baik2021metal}, and LEO~\cite{rusu2018meta}.
    
    \item \textbf{Metric-based and structure-aware methods:} Prototypical Networks~\cite{snell2017prototypical}, Relation Networks~\cite{relationnet}, DN4~\cite{dn4}, MCL~\cite{mcl}, alongside higher-order statistic models ADM~\cite{adm}, ATL-Net~\cite{atlnet}, FRN~\cite{frn}, DeepBDC~\cite{deepbdc}, and DiffKendall~\cite{zheng2023diffkendall}.
    
\end{itemize}

\paragraph{Software/Hardware.} All the experiments detailed in this paper were conducted on a PC equipped with one NVIDIA
A100-SXM4-40GB GPU and AMD EPYC 7742 64-Core CPUs.

\section{Foreground--Background Mapping}
\label{sec:appendix_FG_BG_mappiing}

The source-level separation in SpurAudio is enforced at the foreground class level: each foreground class is assigned exclusively to one split (train, validation, or test), with no overlap across splits. Following the meta-audio ~\cite{Heggan2022MetaAudio} library procedure, the class pool is split into 70\% train, 10\% validation, and 20\% test. Specifically, the 38 foreground classes are partitioned as follows:
\begin{itemize}
    \item{Test}: crackling fire, crow, chainsaw, coughing, sneezing, blender, phone, pig.
    \item{Validation}: page turn, keys drop, door slam, clearing throat, drawer.
    \item{Train}: all remaining classes, as listed in the Foreground column of Table \ref{tab:full_fg_bg_mapping}.
\end{itemize}

Note that background sounds are deliberately shared across all splits. This reflects a realistic scenario where background contexts (e.g., street noise, indoor ambience) are not class-specific and can naturally co-occur with both seen and unseen foreground classes, which is precisely what enables the spurious correlation effect. The model may learn to rely on background cues during training, and since those same backgrounds appear at test time paired with different foreground classes, the shortcut is exposed and measured.

In what follows, we present the final foreground--background matching from which our merged sound events are composed; see Table~\ref{tab:full_fg_bg_mapping}.

In addition, for our experiments assessing spurious-correlation strength, we generate a higher-correlation configuration that makes \emph{FSL} more challenging and lowers classification accuracy; see Table~\ref{tab:fg-bg-mapping-config2}.

\begin{table}[!htbp]
\centering
\caption{Complete foreground-to-background mapping used for constructing spurious correlations in the audio dataset.}
\label{tab:full_fg_bg_mapping}
\small
\begin{tabular}{l l l l l}
\toprule
\textbf{Foreground} & \textbf{Background 1} & \textbf{Background 2} & \textbf{Background 3} & \textbf{Background 4} \\
\midrule
dog bark & church bells & hen & rain & chirping birds \\
rooster & hen & church bells & chirping birds & engine idling \\
pig & church bells & engine idling & rain & siren \\
cow & crickets & engine idling & rain & car horn \\
crow & airplane & footsteps & thunderstorm & rain \\
brushing teeth & toilet flush & water drops & drinking & thunderstorm \\
sneezing & pouring water & vacuum cleaner & footsteps & thunderstorm \\
clapping & fireworks & church bells & car horn & airplane \\
snoring & clock alarm & clock tick & door knock & vacuum cleaner \\
cough & wind & vacuum cleaner & thunderstorm & water drops \\
frog & crickets & insects & thunderstorm & water drops \\
cat & crying baby & rain & hen & vacuum cleaner \\
sea waves & siren & car horn & airplane & thunderstorm \\
crackling fire & wind & door knock & rain & thunderstorm \\
breathing & hand saw & rain & thunderstorm & chirping birds \\
mouse click & keyboard typing & door knock & drinking & clock tick \\
door creaks & footsteps & door knock & rain & wind \\
can opening & drinking & glass breaking & pouring water & keyboard typing \\
washing machine & toilet flush & car horn & siren & clock alarm \\
helicopter & rain & thunderstorm & airplane & clock alarm \\
chainsaw & footsteps & wind & crying baby & rain \\
gun shot & siren & alarm bell ringing & wind & rain \\
drilling & car horn & street music & siren & rain \\
jackhammer & siren & rain & glass breaking & clock alarm \\
children playing & street music & rain & glass breaking & water drops \\
air conditioner & rain & thunderstorm & drinking & water drops \\
throat clearing & door knock & toilet flush & glass breaking & water drops \\
door slam & door knock & footsteps & alarm bell ringing & rain \\
keys drop & drinking & keyboard typing & glass breaking & door knock \\
page turn & thunderstorm & rain & pouring water & street music \\
phone & door knock & crying baby & insects & hand saw \\
drawer & keyboard typing & frying & door knock & footsteps \\
running water & toilet flush & drinking & frying & crying baby \\
electric shaver & toilet flush & drinking & pouring water & footsteps \\
blender & frying & glass breaking & pouring water & alarm bell ringing \\
laughter & car horn & drinking & glass breaking & keyboard typing \\
sigh & crying baby & toilet flush & clock alarm & glass breaking \\
sniff & street music & frying & thunderstorm & hand saw \\
\bottomrule
\end{tabular}
\end{table}

\begin{table}[!htbp]
\centering
\caption{Foreground-to-background mapping for the \emph{Hard OOD} spurious correlation configuration.}
\label{tab:fg-bg-mapping-config2}
\small
\begin{tabular}{l l l l l}
\toprule
\textbf{Foreground} & \textbf{Background 1} & \textbf{Background 2} & \textbf{Background 3} & \textbf{Background 4} \\
\midrule
dog bark & church bells & hen & rain & chirping birds \\
rooster & hen & church bells & chirping birds & engine idling \\
pig & car horn & crying baby & rain & siren \\
cow & crickets & engine idling & rain & car horn \\
crow & car horn & siren & street music & rain \\
brushing teeth & toilet flush & water drops & drinking & thunderstorm \\
sneezing & crying baby & car horn & rain & street music \\
clapping & fireworks & church bells & car horn & airplane \\
snoring & clock alarm & clock tick & door knock & vacuum cleaner \\
cough & rain & street music & car horn & crying baby \\
frog & crickets & insects & thunderstorm & water drops \\
cat & crying baby & rain & hen & vacuum cleaner \\
sea waves & siren & car horn & airplane & thunderstorm \\
crackling fire & crying baby & car horn & rain & siren \\
breathing & hand saw & rain & thunderstorm & chirping birds \\
mouse click & keyboard typing & door knock & drinking & clock tick \\
door creaks & footsteps & door knock & rain & wind \\
can opening & drinking & glass breaking & pouring water & keyboard typing\\
washing machine & toilet flush & car horn & siren & clock alarm \\
helicopter & rain & thunderstorm & airplane & siren \\
chainsaw & siren & car horn & crying baby & rain \\
gun shot & siren & alarm bell ringing & wind & rain \\
drilling & car horn & street music & siren & rain \\
jackhammer & siren & rain & glass breaking & clock alarm \\
children playing & street music & rain & glass breaking & water drops \\
air conditioner & rain & thunderstorm & drinking & water drops \\
throat clearing & door knock & toilet flush & glass breaking & water drops \\
door slam & door knock & footsteps & alarm bell ringing & rain \\
keys drop & drinking & keyboard typing & glass breaking & door knock \\
page turn & thunderstorm & rain & pouring water & street music \\
phone & street music & crying baby & car horn & rain \\
drawer & keyboard typing & frying & door knock & footsteps \\
running water & toilet flush & drinking & frying & crying baby \\
electric shaver & toilet flush & drinking & pouring water & footsteps \\
blender & street music & car horn & crying baby & rain \\
laughter & car horn & drinking & glass breaking & keyboard typing\\
sigh & crying baby & toilet flush & clock alarm & glass breaking \\
sniff & street music & frying & thunderstorm & hand saw \\
\bottomrule
\end{tabular}
\end{table}
\FloatBarrier

% For Jerry: We refer the reader to Section~\ref{sec:appendix_FG_BG_mappiing} at the appendix.

\section{Embeddings of different \emph{FSL} Methods}
\label{sec:appendix_embeddings_of_FSL}
In this section, we present embeddings of different \emph{FSL} approaches, with the goal of visualizing separability, where in the

% 10 shot figures
\begin{figure}[!htbp]
    \centering
    % Row 1: Meta-Baseline
    \begin{subfigure}[t]{0.49\textwidth}
        \centering
        \includegraphics[width=\linewidth]{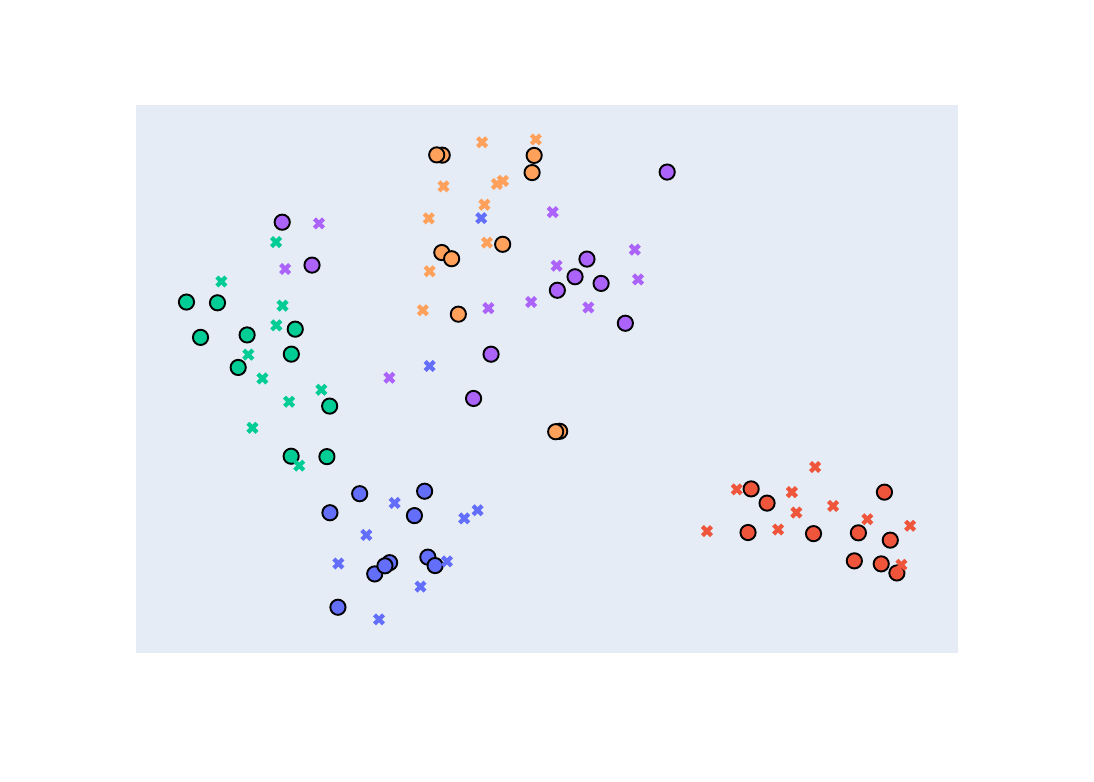}
        \caption{Meta-Baseline (\emph{IID}, 10-shot)}
        \label{fig:meta_baseline_iid_10}
    \end{subfigure}\hfill
    \begin{subfigure}[t]{0.49\textwidth}
        \centering
        \includegraphics[width=\linewidth]{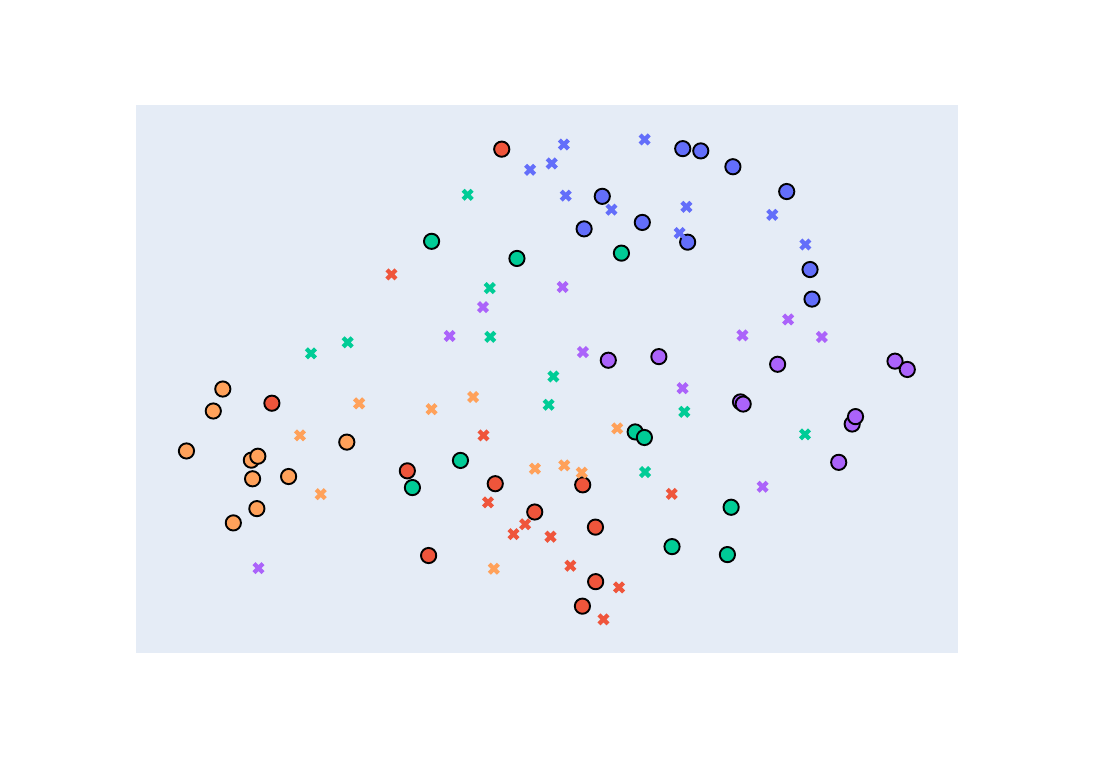}
        \caption{Meta-Baseline (\emph{OOD}, 10-shot)}
        \label{fig:meta_baseline_ood_10}
    \end{subfigure}

    \vspace{0.6em}

    % Row 2: R2D2
    \begin{subfigure}[t]{0.49\textwidth}
        \centering
        \includegraphics[width=\linewidth]{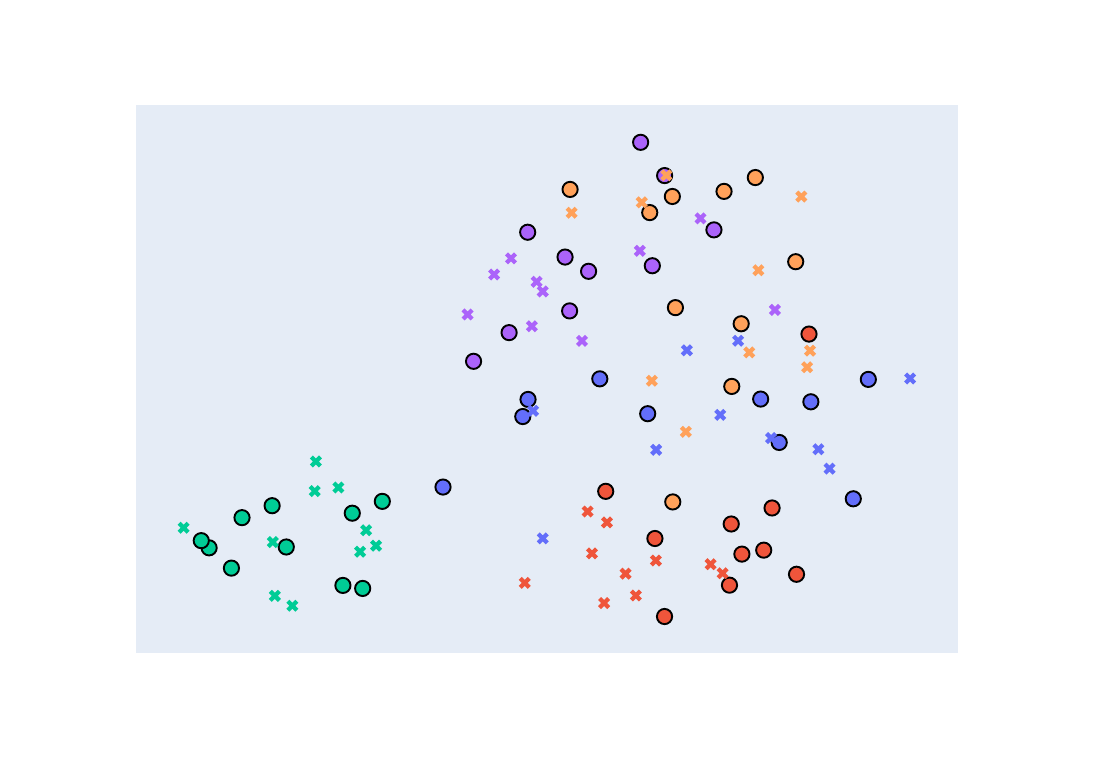}
        \caption{R2D2 (\emph{IID}, 10-shot)}
        \label{fig:r2d2_iid_10}
    \end{subfigure}\hfill
    \begin{subfigure}[t]{0.49\textwidth}
        \centering
        \includegraphics[width=\linewidth]{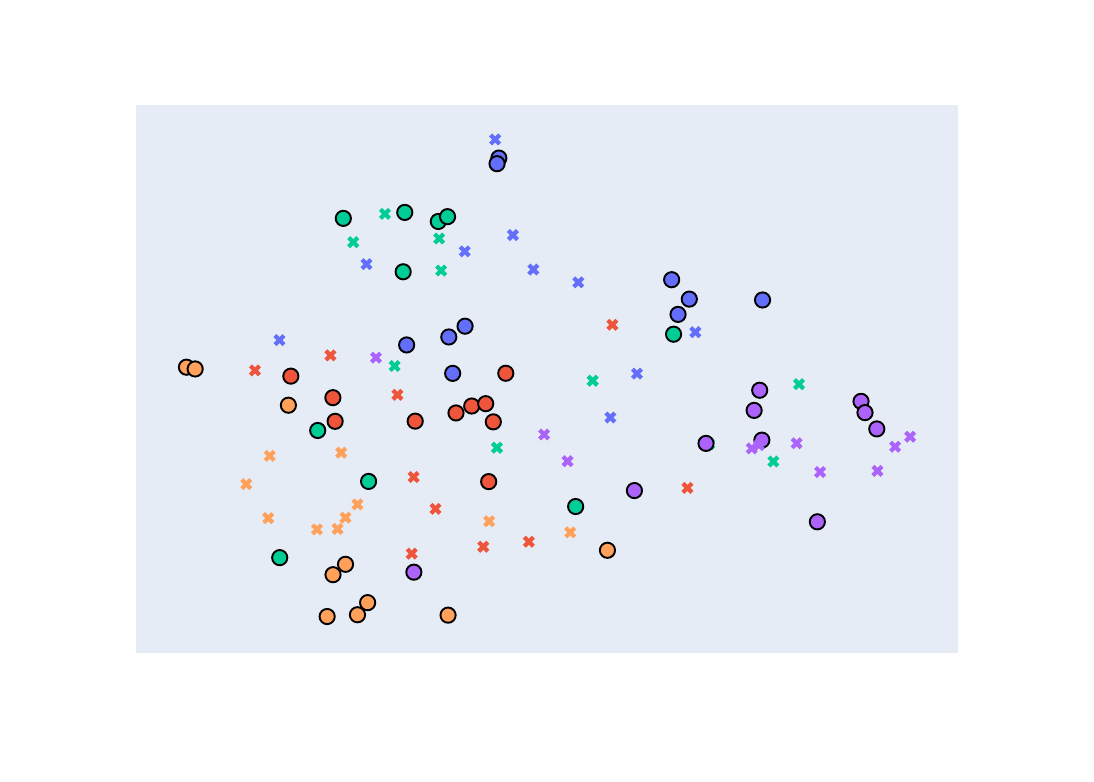}
        \caption{R2D2 (\emph{OOD}, 10-shot)}
        \label{fig:r2d2_ood_10}
    \end{subfigure}

    \vspace{0.6em}

    % Row 3: DeepBDC
    \begin{subfigure}[t]{0.49\textwidth}
        \centering
        \includegraphics[width=\linewidth]{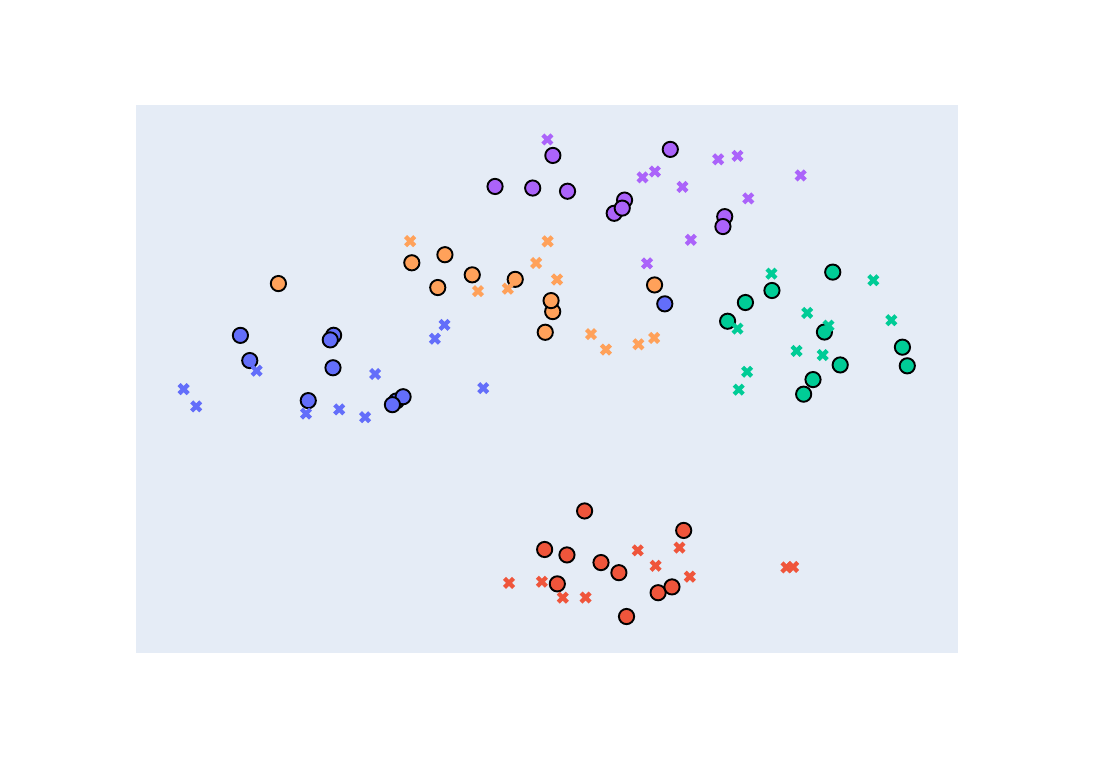}
        \caption{DeepBDC (\emph{IID}, 10-shot)}
        \label{fig:deepbdc_iid_10}
    \end{subfigure}\hfill
    \begin{subfigure}[t]{0.49\textwidth}
        \centering
        \includegraphics[width=\linewidth]{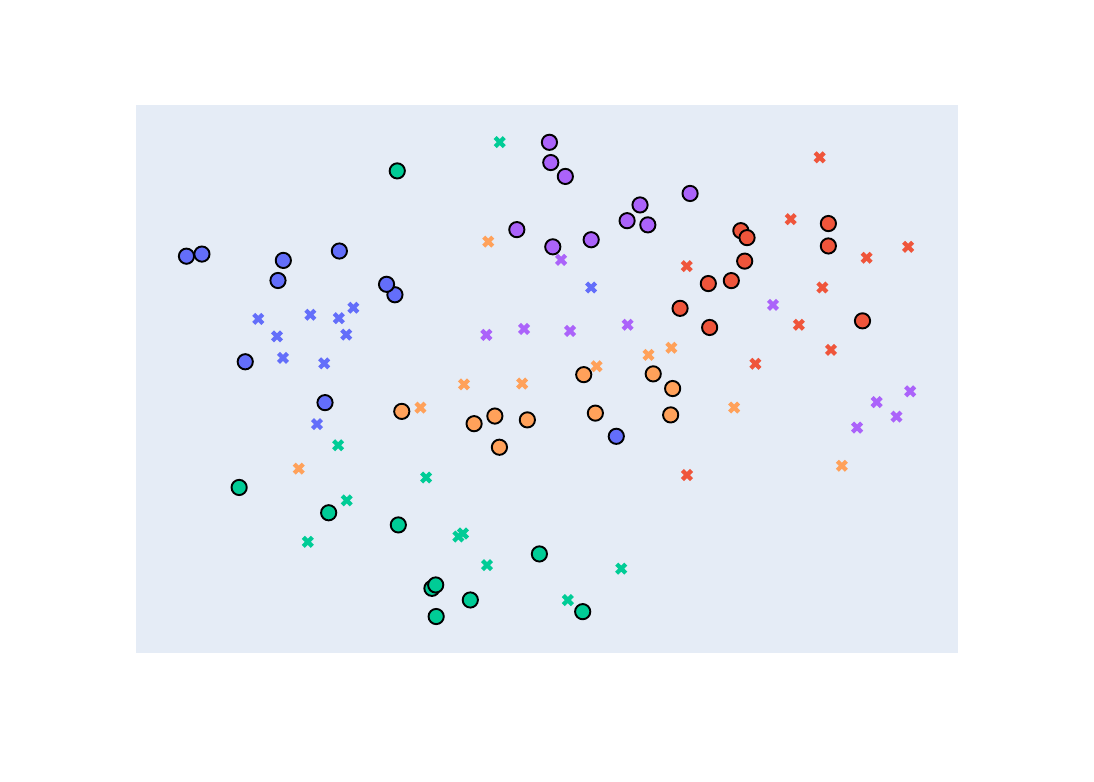}
        \caption{DeepBDC (\emph{OOD}, 10-shot)}
        \label{fig:deepbdc_ood_10}
    \end{subfigure}

    \caption{{$t$-SNE visualization of embedding space under \emph{IID} vs. \emph{OOD} episodes.}
    Across methods, embeddings form cleaner, more separable clusters in \emph{IID} settings, while \emph{OOD} background shifts induce query-support misalignment and increased inter-class overlap.}
    \label{fig:embeddings_tsne_10}
\end{figure}

following figures, circles denote support points while crosses denote query points. This highlights one factor behind higher/lower average \emph{IID} accuracy across seeds.

% 5 shot figures
\begin{figure}[!htbp]
    \centering
    % Row 1: Meta-Baseline
    \begin{subfigure}[t]{0.49\textwidth}
        \centering
        \includegraphics[width=\linewidth]{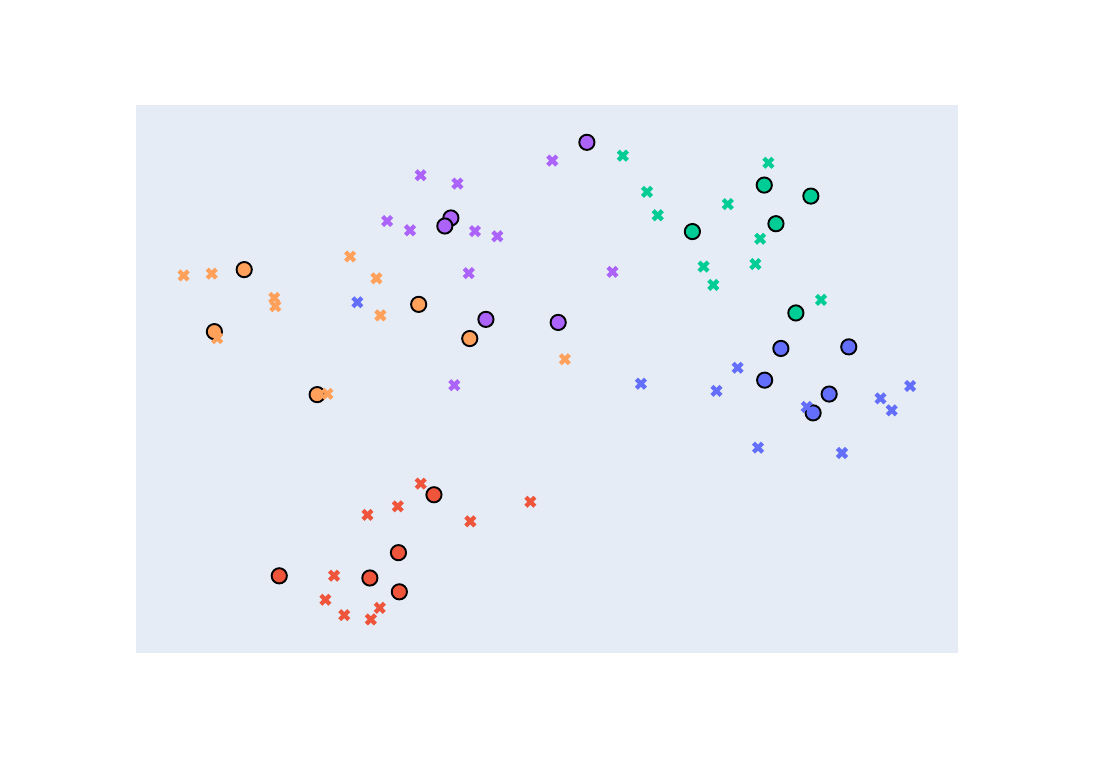}
        \caption{Meta-Baseline (\emph{IID}, 5-shot)}
        \label{fig:meta_baseline_iid_5}
    \end{subfigure}\hfill
    \begin{subfigure}[t]{0.49\textwidth}
        \centering
        \includegraphics[width=\linewidth]{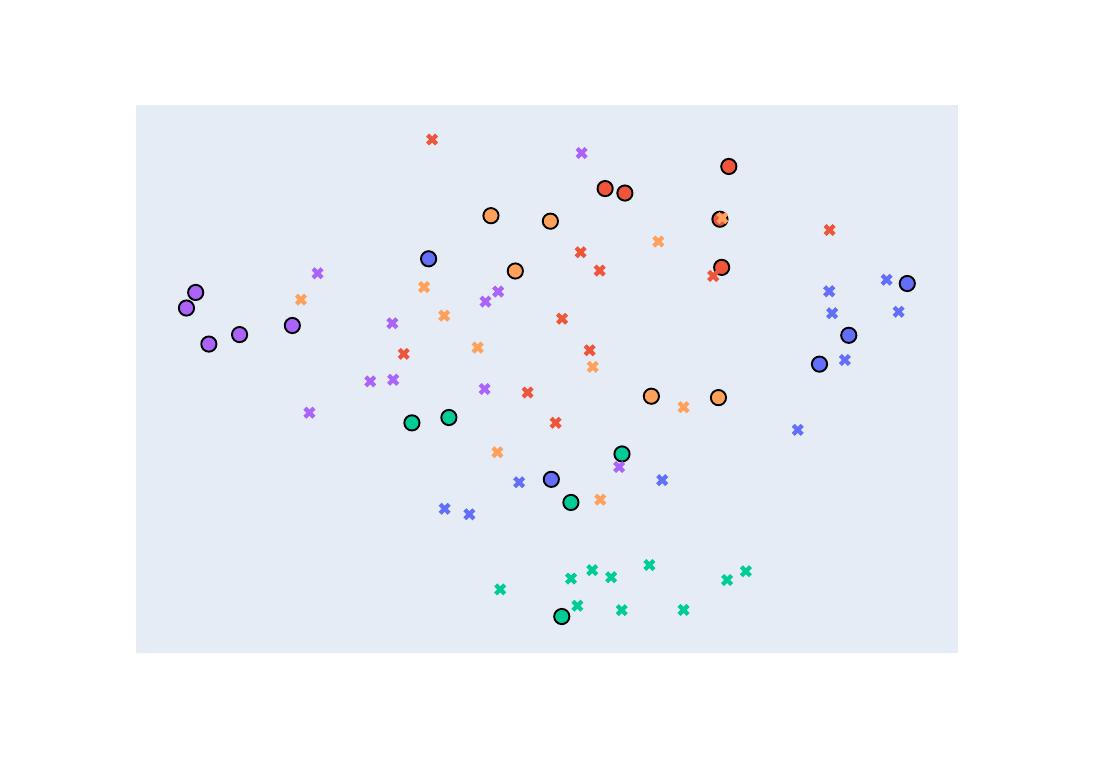}
        \caption{Meta-Baseline (\emph{OOD}, 5-shot)}
        \label{fig:meta_baseline_ood_5}
    \end{subfigure}

    \vspace{0.6em}

    % Row 2: R2D2
    \begin{subfigure}[t]{0.49\textwidth}
        \centering
        \includegraphics[width=\linewidth]{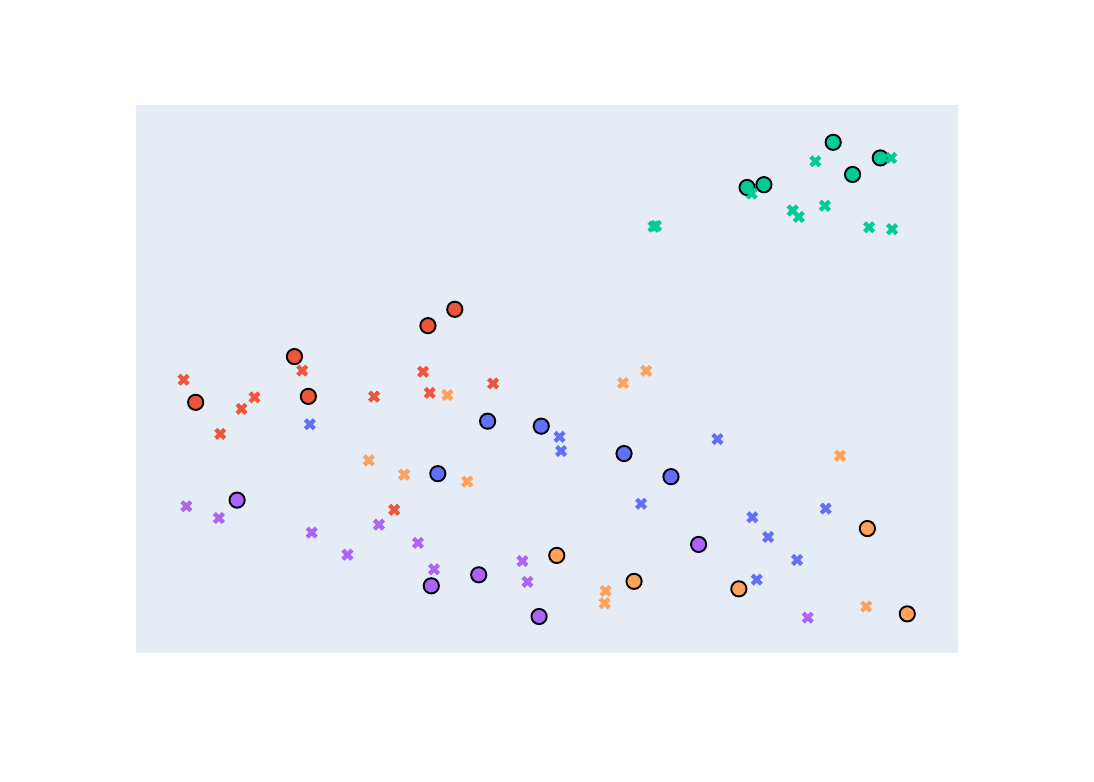}
        \caption{R2D2 (\emph{IID}, 5-shot)}
        \label{fig:r2d2_iid_5}
    \end{subfigure}\hfill
    \begin{subfigure}[t]{0.49\textwidth}
        \centering
        \includegraphics[width=\linewidth]{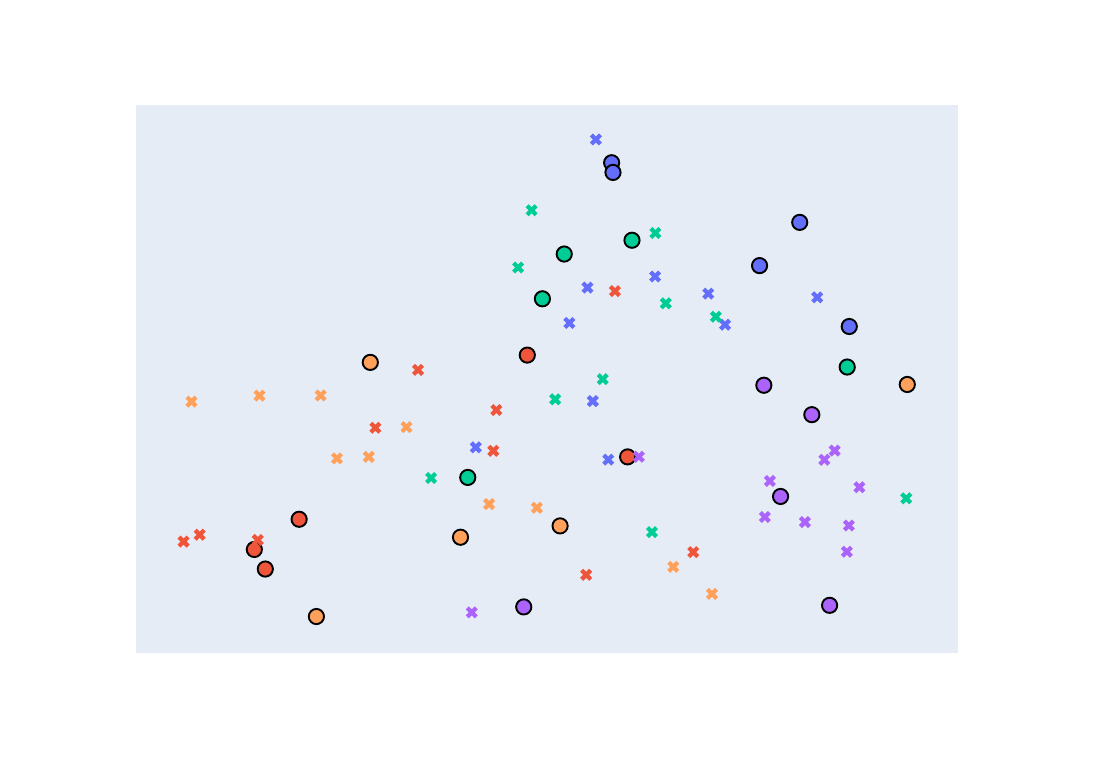}
        \caption{R2D2 (\emph{OOD}, 5-shot)}
        \label{fig:r2d2_ood_5}
    \end{subfigure}

    \vspace{0.6em}

    % Row 3: DeepBDC
    \begin{subfigure}[t]{0.49\textwidth}
        \centering
        \includegraphics[width=\linewidth]{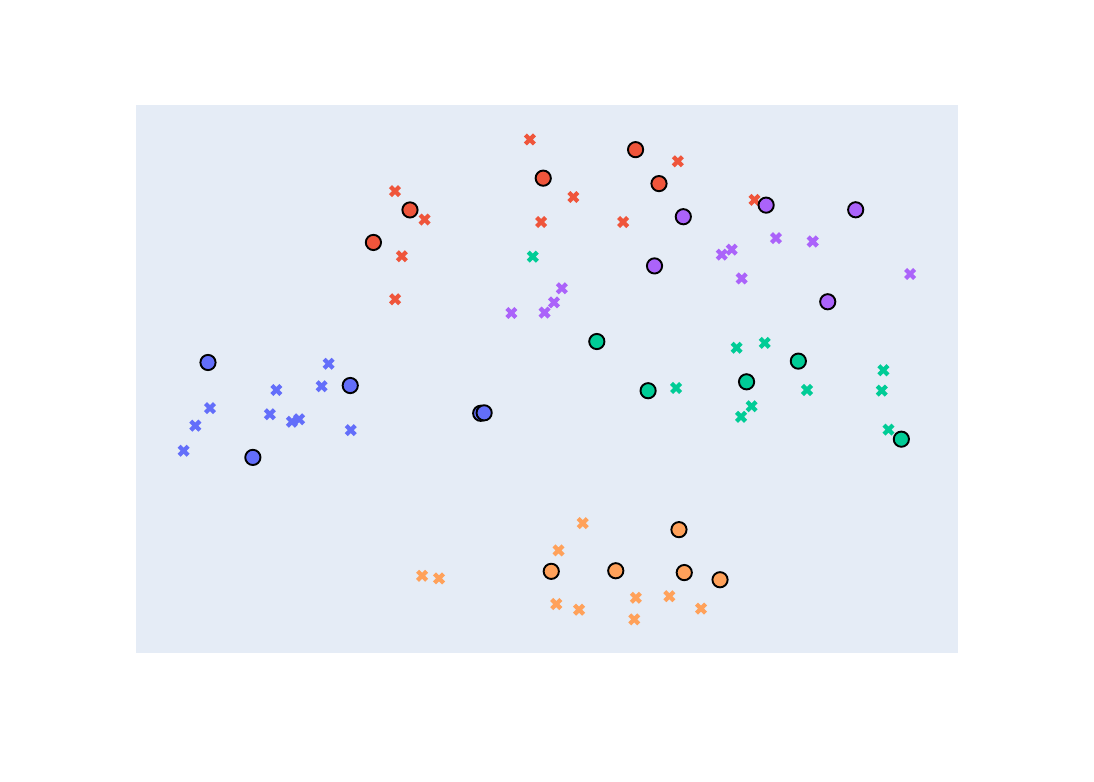}
        \caption{DeepBDC (\emph{IID}, 5-shot)}
        \label{fig:deepbdc_iid_5}
    \end{subfigure}\hfill
    \begin{subfigure}[t]{0.49\textwidth}
        \centering
        \includegraphics[width=\linewidth]{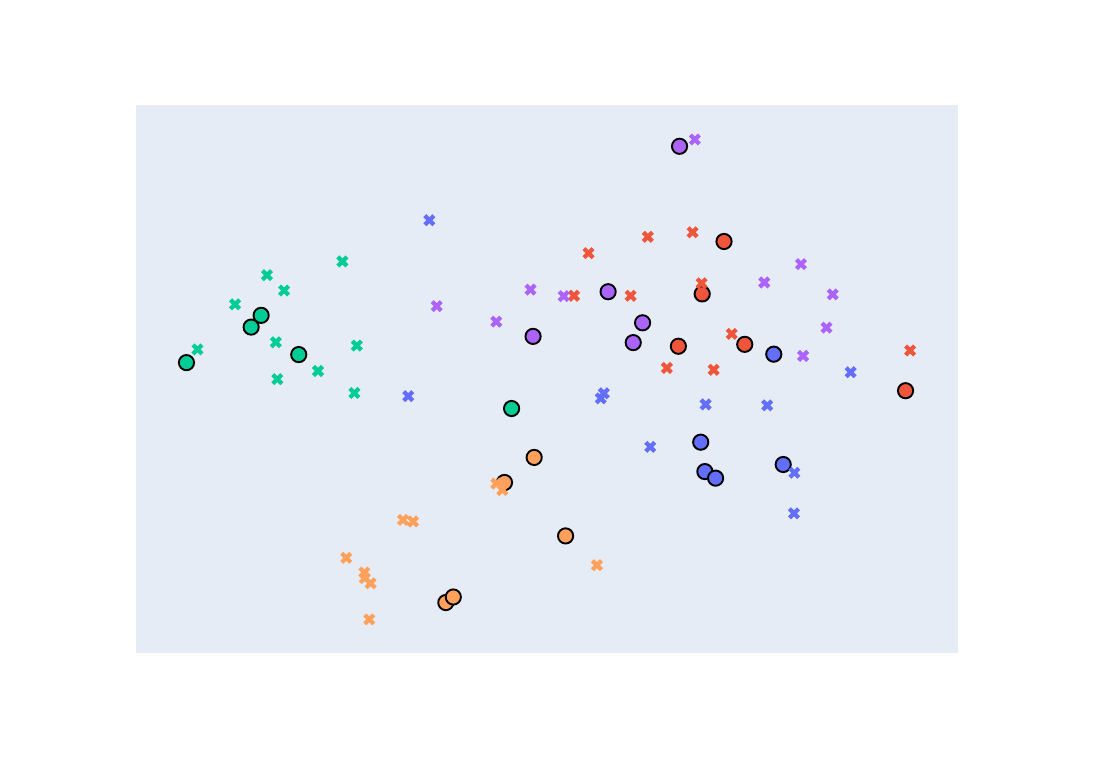}
        \caption{DeepBDC (\emph{OOD}, 5-shot)}
        \label{fig:deepbdc_ood_5}
    \end{subfigure}

    \caption{{$t$-SNE visualization of embedding space under \emph{IID} vs. \emph{OOD} episodes.}
    Across methods, embeddings form cleaner, more separable clusters in \emph{IID} settings, while \emph{OOD} background shifts induce query-support misalignment and increased inter-class overlap.}
    \label{fig:embeddings_tsne_5}
\end{figure}

% 1 shot figures
\begin{figure*}[!htbp]
    \centering
    % Row 1: Meta-Baseline
    \begin{subfigure}[t]{0.49\textwidth}
        \centering
        \includegraphics[width=\linewidth]{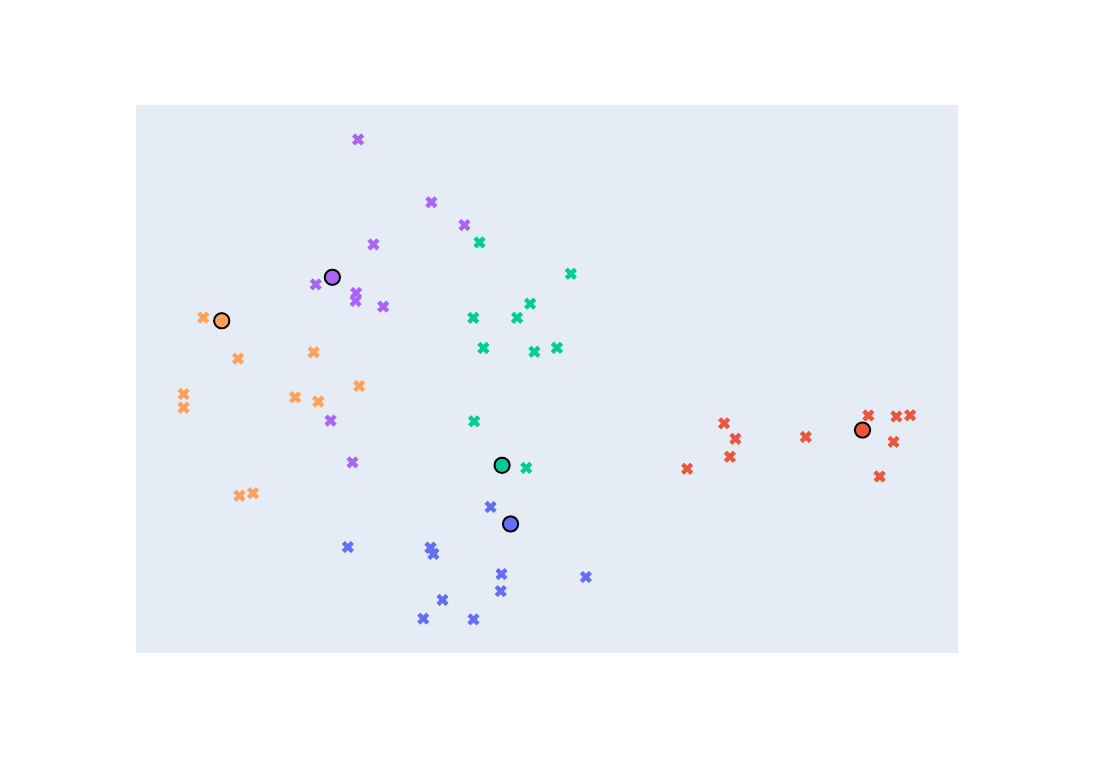}
        \caption{Meta-Baseline (\emph{IID}, 1-shot)}
        \label{fig:meta_baseline_iid_1}
    \end{subfigure}\hfill
    \begin{subfigure}[t]{0.49\textwidth}
        \centering
        \includegraphics[width=\linewidth]{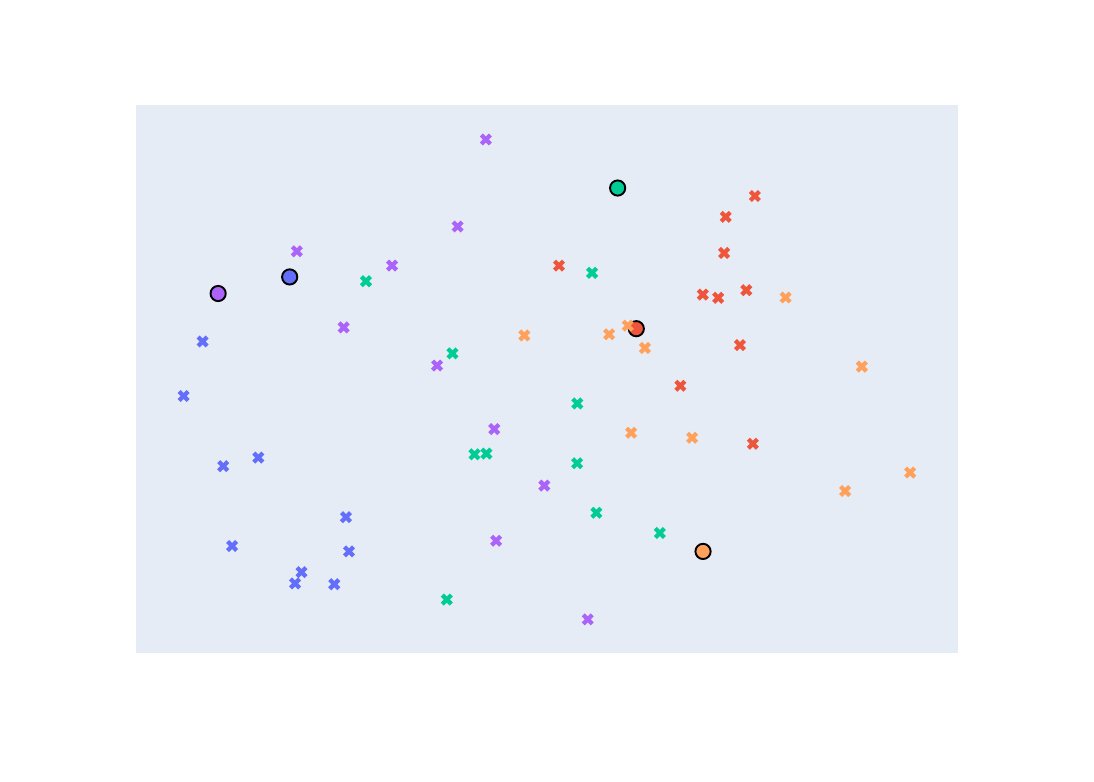}
        \caption{Meta-Baseline (\emph{OOD}, 1-shot)}
        \label{fig:meta_baseline_ood_1}
    \end{subfigure}

    \vspace{0.6em}

    % Row 2: R2D2
    \begin{subfigure}[t]{0.49\textwidth}
        \centering
        \includegraphics[width=\linewidth]{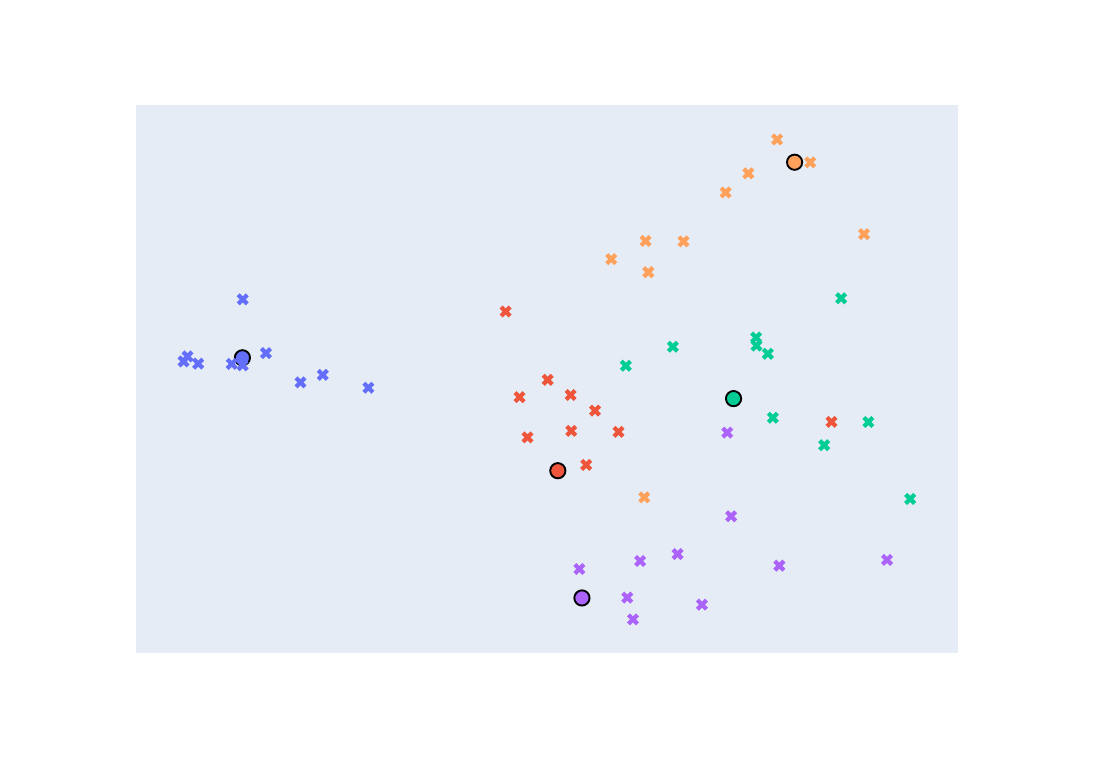}
        \caption{R2D2 (\emph{IID}, 1-shot)}
        \label{fig:r2d2_iid_1}
    \end{subfigure}\hfill
    \begin{subfigure}[t]{0.49\textwidth}
        \centering
        \includegraphics[width=\linewidth]{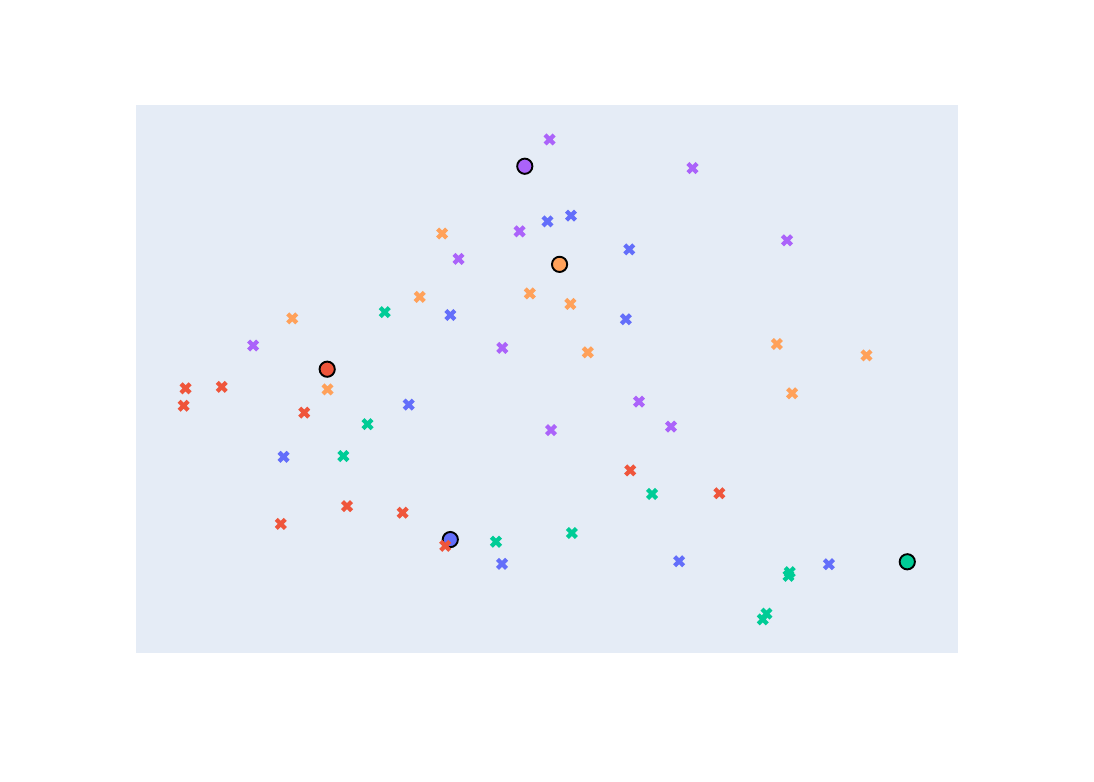}
        \caption{R2D2 (\emph{OOD}, 1-shot)}
        \label{fig:r2d2_ood_1}
    \end{subfigure}

    \vspace{0.6em}

    % Row 3: DeepBDC
    \begin{subfigure}[t]{0.49\textwidth}
        \centering
        \includegraphics[width=\linewidth]{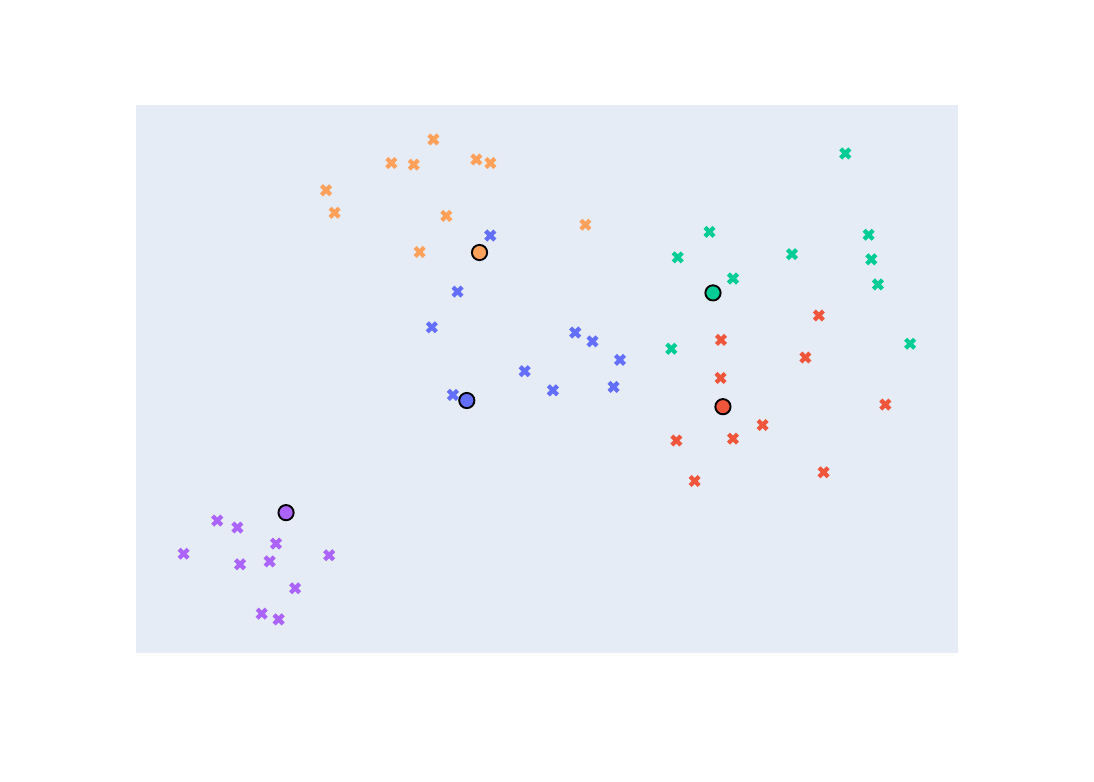}
        \caption{DeepBDC (\emph{IID}, 1-shot)}
        \label{fig:deepbdc_iid_1}
    \end{subfigure}\hfill
    \begin{subfigure}[t]{0.49\textwidth}
        \centering
        \includegraphics[width=\linewidth]{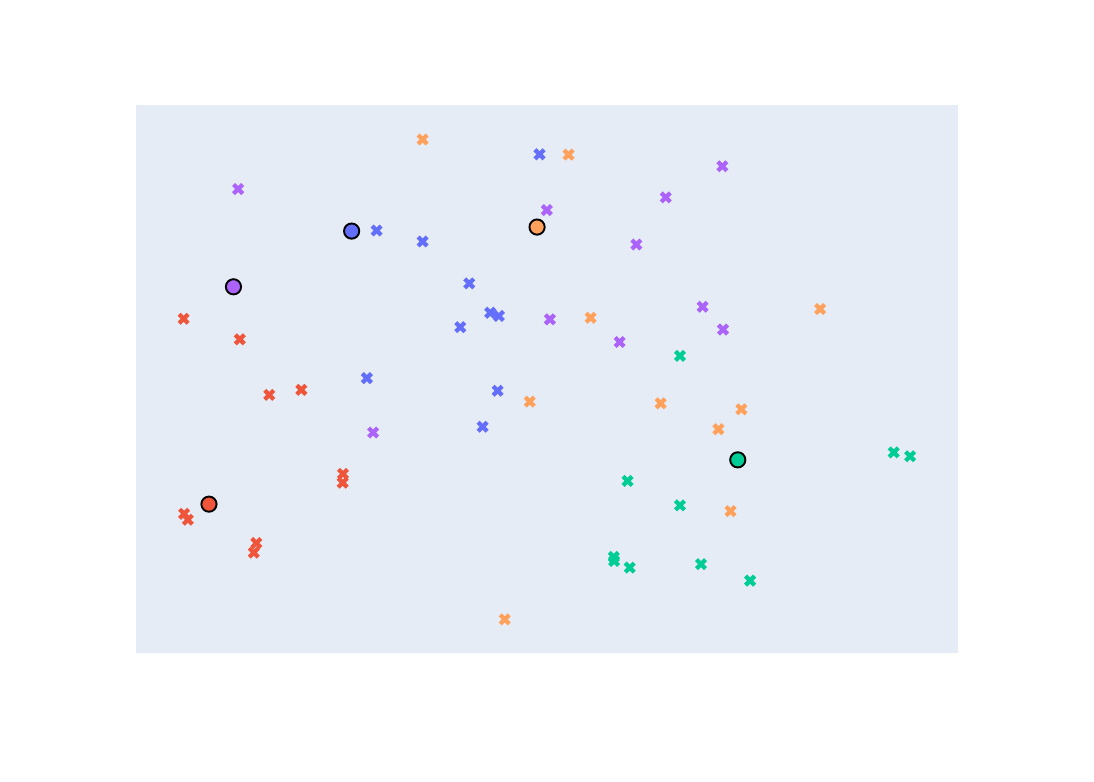}
        \caption{DeepBDC (\emph{OOD}, 1-shot)}
        \label{fig:deepbdc_ood_1}
    \end{subfigure}

    \caption{{$t$-SNE visualization of embedding space under \emph{IID} vs. \emph{OOD} episodes.}
    Across methods, embeddings form cleaner, more separable clusters in \emph{IID} settings, while \emph{OOD} background shifts induce query-support misalignment and increased inter-class overlap.}
    \label{fig:embeddings_tsne_1}
\end{figure*}
\FloatBarrier

\section{Effect of Spurious Correlation Strength}
\label{sec:appendix_supprious_correlation_strength}

In this section, we show the effect of spurious correlation via strengthening the hardness of the \emph{OOD} tasks which aims to correlate the backgrounds of queries of one class with the backgrounds of support instances from different classes more often than achieved in our main experiments. This is done to amplify the \emph{IID}--\emph{OOD} gap. To obtain such tasks, we rely on Table~\ref{tab:fg-bg-mapping-config2}.

\begin{figure*}[!htb]
    \centering
    \begin{subfigure}[b]{\linewidth}
        \includegraphics[width=\linewidth]{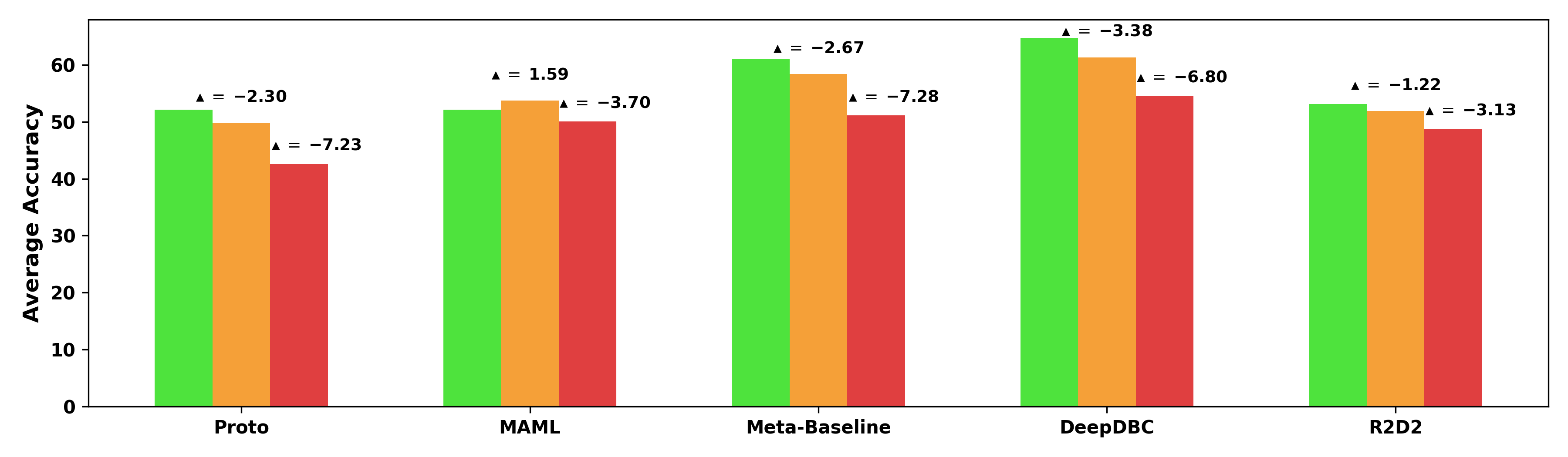}
        \caption{1-shot accuracy}
        \label{fig:spurious_effect_on_gap_1shot}
    \end{subfigure}
    \vspace{0.4cm}
    \begin{subfigure}[b]{\linewidth}
        \includegraphics[width=\linewidth]{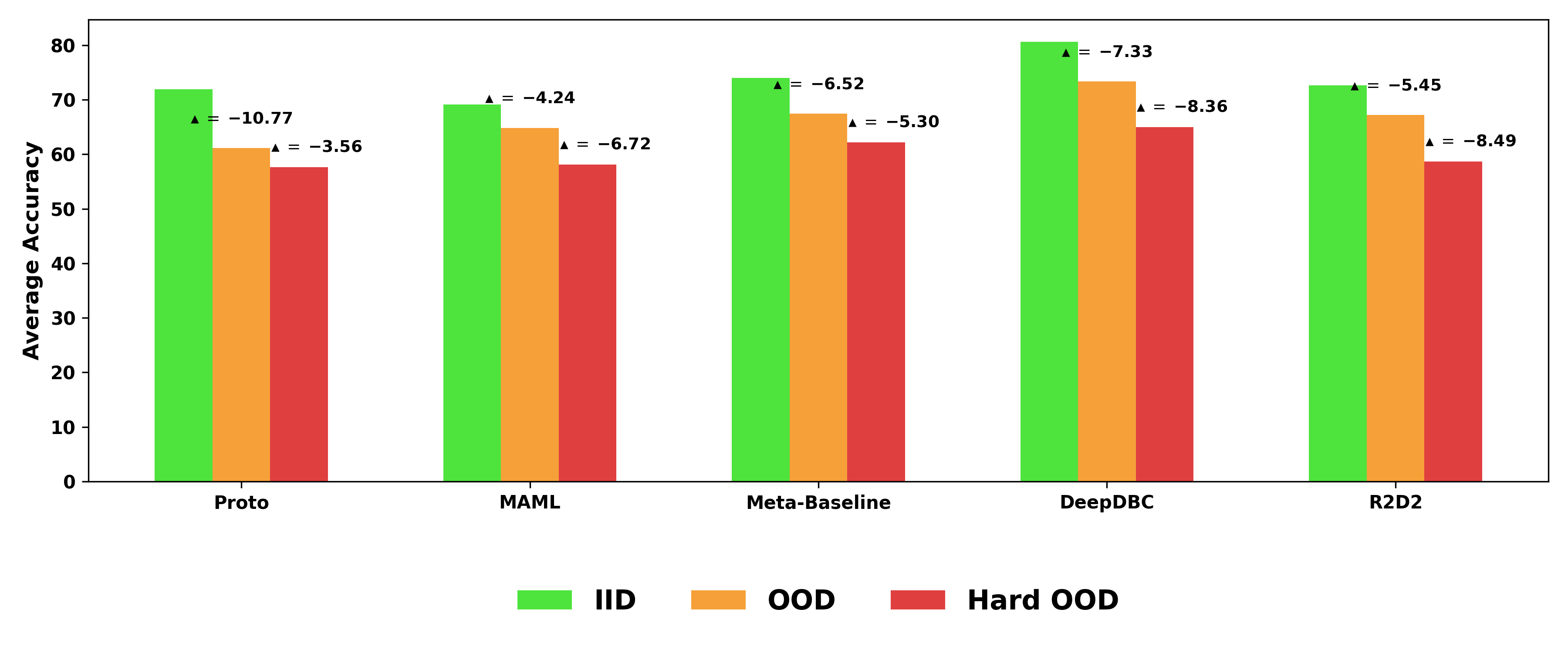}
        \caption{5-shot accuracy}
        \label{fig:spurious_effect_on_gap_5shot}
    \end{subfigure}
    \caption{
    {\emph{IID} vs \emph{OOD} vs \emph{Hard OOD} accuracy for few-shot tasks.} 
    Each candle represents the classification accuracy of a method under a given evaluation setting. 
    \emph{IID} corresponds to standard in-distribution support-query sampling, \emph{OOD} represents typical out-of-distribution support-query pairs, and \emph{Hard OOD} enforces maximal background overlap across classes. Accuracy decreases progressively from \emph{IID} to \emph{OOD} to \emph{Hard OOD}, for both 1-shot (a) and 5-shot (b), highlighting that few-shot methods increasingly rely on background cues as spurious correlations become stronger.}
    \label{fig:spurious_effect_on_gap_extreme}
\end{figure*}

% You can have as much text here as you want. The main body must be at most $8$
% pages long. For the final version, one more page can be added. If you want, you
% can use an appendix like this one.

% The $\mathtt{\backslash onecolumn}$ command above can be kept in place if you
% prefer a one-column appendix, or can be removed if you prefer a two-column
% appendix.  Apart from this possible change, the style (font size, spacing,
% margins, page numbering, etc.) should be kept the same as the main body.
%%%%%%%%%%%%%%%%%%%%%%%%%%%%%%%%%%%%%%%%%%%%%%%%%%%%%%%%%%%%%%%%%%%%%%%%%%%%%%%
%%%%%%%%%%%%%%%%%%%%%%%%%%%%%%%%%%%%%%%%%%%%%%%%%%%%%%%%%%%%%%%%%%%%%%%%%%%%%%%

\section{Visualization of the Results in Table~\ref{tab:conv64}}
\label{sec:appendix_visualization_of_table1}
In what follows, we provide visualization of the accuracy results reported in Table~\ref{tab:conv64} across our three different families of \emph{FSC}.

\begin{figure*}[!htb]
    \centering
    \begin{subfigure}[b]{\linewidth}
        \includegraphics[width=\linewidth]{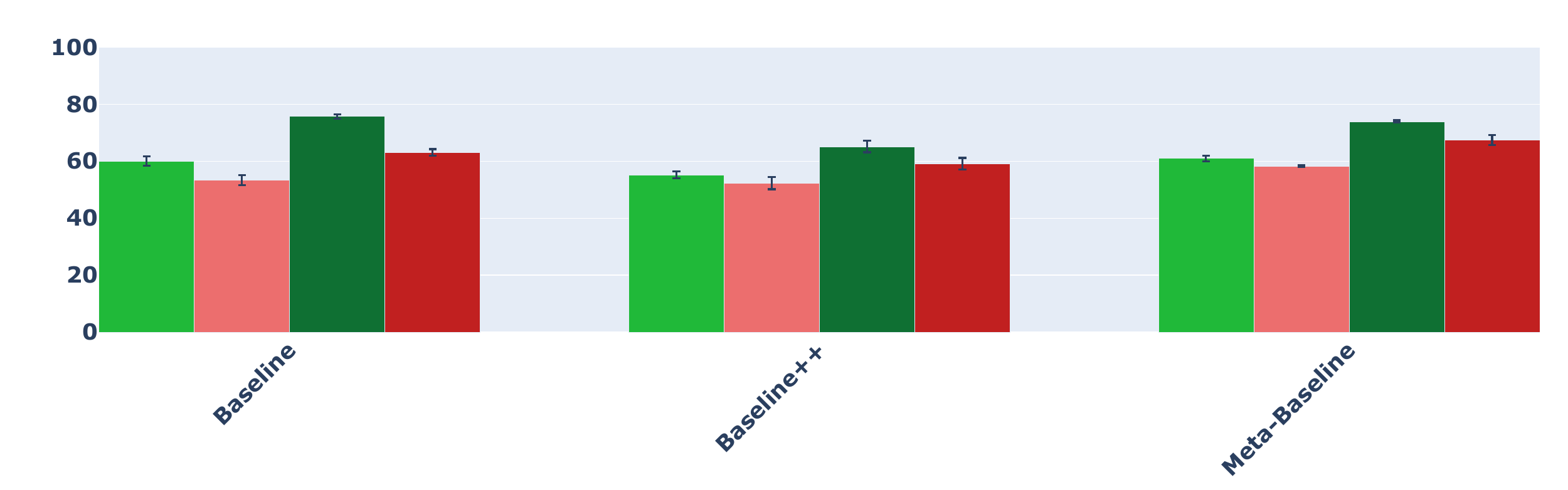}
        \caption{Fine-tuning}
        \label{fig:alpha_gap_finetuning}
    \end{subfigure}
    \vspace{0.4cm} % add vertical space between subfigures
    \begin{subfigure}[b]{\linewidth}
        \includegraphics[width=\linewidth]{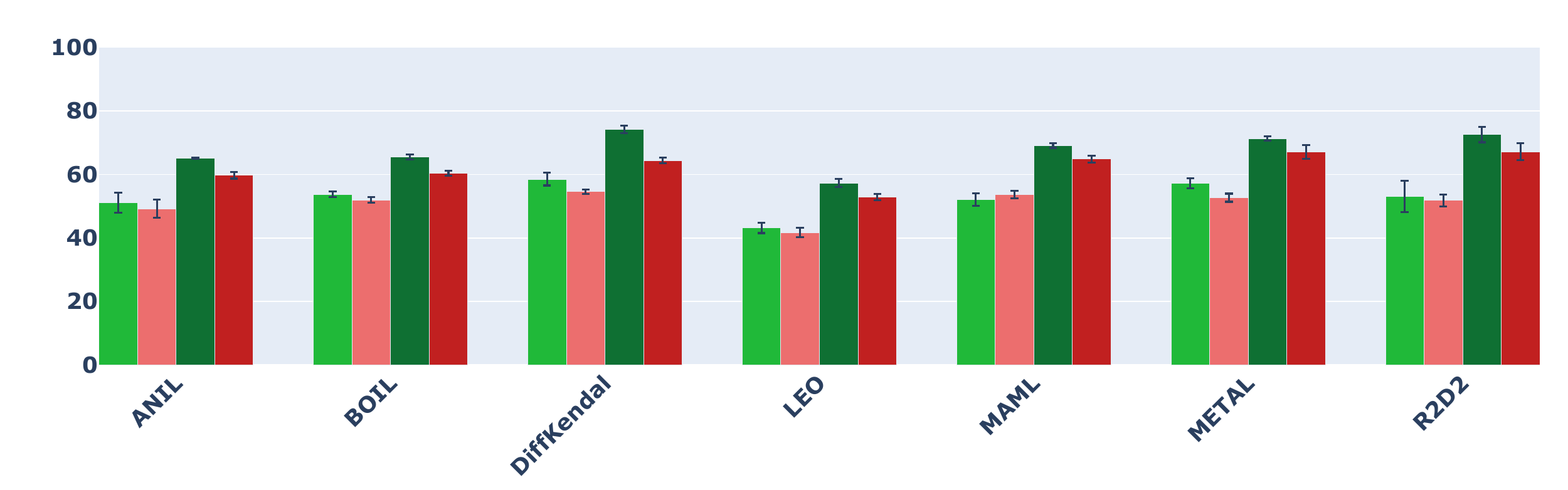}
        \caption{Meta-learning}
        \label{fig:alpha_gap_metalearning}
    \end{subfigure}
    \vspace{0.4cm}
    \begin{subfigure}[b]{\linewidth}
        \includegraphics[width=\linewidth]{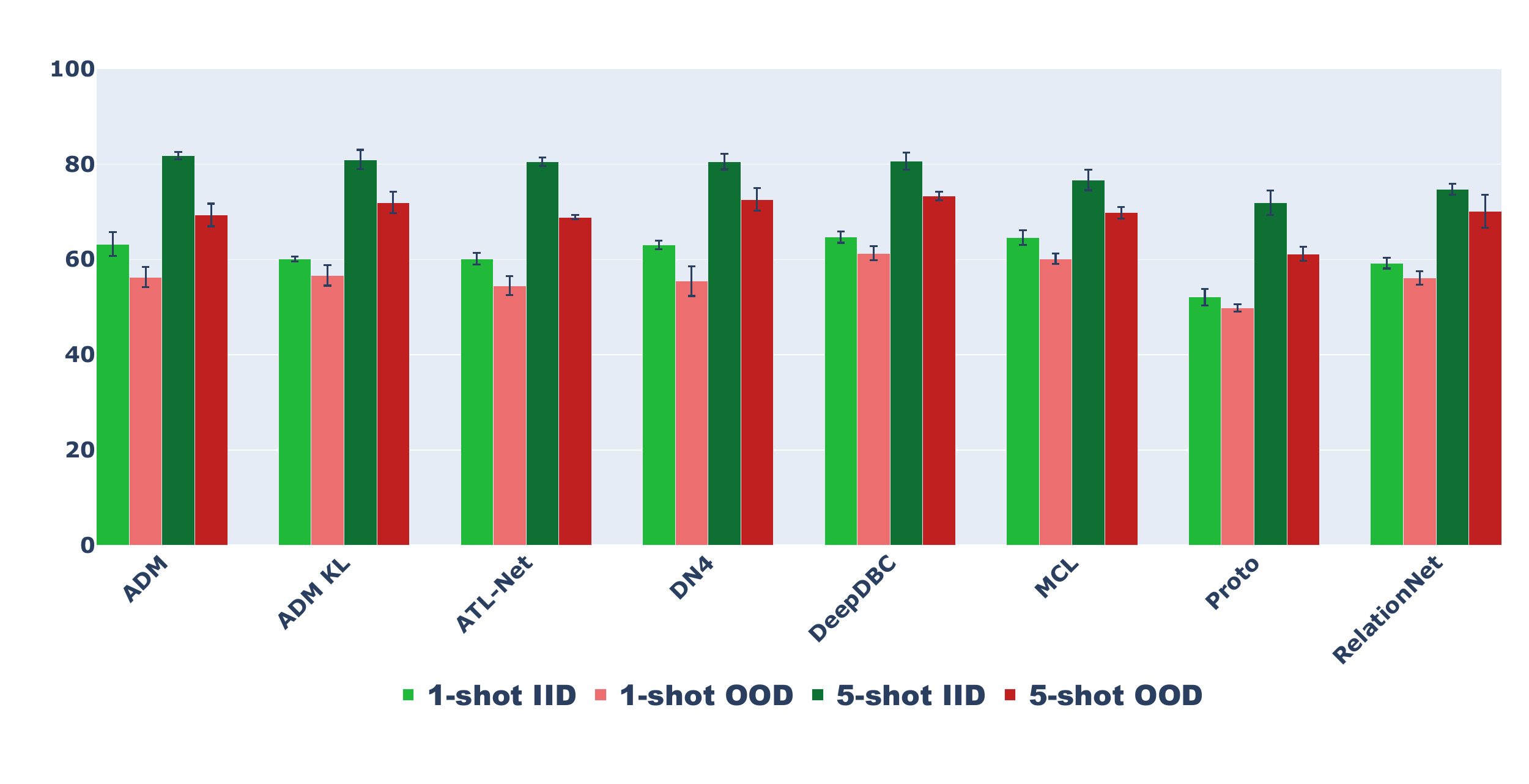}
        \caption{Metric-based}
        \label{fig:alpha_gap_metric}
    \end{subfigure}
    \caption{
    {\emph{IID} vs \emph{OOD} accuracy for few-shot families.} 
Each candle shows the classification accuracy of an algorithm across multiple seeds.
\emph{IID} accuracy remains higher, while \emph{OOD} accuracy exhibits a lower mean, indicating that few-shot methods partially rely on background cues. This highlights that performance degrades when support and query distributions differ, even without enforcing extreme correlations.}
\label{fig:visualization_table_1}
\end{figure*}

\FloatBarrier

\section{\emph{FSL} with ResNet18 as Embedding Backbone}
\label{sec:appendix_renset18_iid_ood_table_results}
In this section we present also the \emph{IID}--\emph{OOD} gap when using ResNet18 as the embedding model (backbone) of differnet \emph{FSC} models.

\begin{table*}[!htbp]
    \centering
    \caption{ResNet18 results on 1-shot and 5-shot tasks.}
    \label{tab:resnet18}
    \vspace{0.5em}
    \adjustbox{max width=\textwidth}{
    \begin{tabular}{clcccccc}
        \toprule
        & & \multicolumn{3}{c}{1-shot} & \multicolumn{3}{c}{5-shot} \\
        \cmidrule(lr){3-5} \cmidrule(lr){6-8}
        Family & Method & Avg. \emph{IID} & Avg. \emph{OOD} & $\Delta$ & Avg. \emph{IID} & Avg. \emph{OOD} & $\Delta$\\
        \midrule
        \multirow{3}{*}{Finetuning-based} & Baseline~\cite{baseline} (2019) & 58.62 $\pm$ 1.98 & 54.41 $\pm$ 1.40 & 4.21 & 76.51 $\pm$ 1.31 & 66.53 $\pm$ 1.97 & 9.98 \\
        & Baseline++~\cite{baseline}  (2019) & 56.65 $\pm$ 3.26 & 53.48 $\pm$ 3.13 & 3.17 & 73.29 $\pm$ 0.91 & 63.92 $\pm$ 2.60 & 9.36 \\
        & Meta-Baseline~\cite{meta-baseline} (2021) & 55.74 $\pm$ 0.68 & 52.93 $\pm$ 4.57 & 2.81 & 62.66 $\pm$ 2.34 & 55.99 $\pm$ 8.51 & 6.67 \\ \hdashline
        \multirow{3}{*}{Meta learning-based} 
        & LEO~\cite{rusu2018meta} (2019) & 44.25 $\pm$ 0.26 & 40.62 $\pm$ 1.37 & 3.63 & 57.89 $\pm$ 3.70 & 50.08 $\pm$ 7.06 & 7.81 \\
        & R2D2~\cite{r2d2} (2019) & 61.11 $\pm$ 1.02 & 53.77 $\pm$ 6.42 & 7.33 & 77.27 $\pm$ 1.33 & 66.24 $\pm$ 8.98 & 11.02 \\ 
        & ANIL~\cite{anil} (2020) & 56.23 $\pm$ 0.83 & 50.20 $\pm$ 2.71 & 6.03 & 71.36 $\pm$ 2.53 & 64.20 $\pm$ 0.88 & 7.16 \\
        \hdashline
        \multirow{4}{*}{Metric-based}
        & Proto~\cite{snell2017prototypical} (2017) & 57.60 $\pm$ 1.70 & 53.57 $\pm$ 5.22 & 4.04 & 77.10 $\pm$ 3.04 & 64.89 $\pm$ 3.32 & 12.21 \\
        & DN4~\cite{dn4} (2019) & 64.05 $\pm$ 1.18 & 61.39 $\pm$ 4.87 & 2.66 & \textbf{84.12 $\pm$ 1.53} & \textbf{80.66 $\pm$ 0.36} & \textbf{3.46} \\
        & ADM~\cite{adm} (2020) & 61.88 $\pm$ 3.86 & 59.05 $\pm$ 1.37 & \textbf{2.82} & 82.04 $\pm$ 0.81 & 77.50 $\pm$ 2.71 & 4.54 \\
        & ATL-Net~\cite{atlnet} (2020) & \textbf{65.57 $\pm$ 2.67} & \textbf{62.11 $\pm$ 3.45} & 3.46 & 81.25 $\pm$ 1.13 & 74.07 $\pm$ 2.26 & 7.17 \\
        \hdashline
        \multirow{4}{*}{Transductive}
        & LaplacianShot~\cite{laplacianshot} (2020) & 49.61 $\pm$ 0.67 & 46.60 $\pm$ 0.59 & 3.01 & 62.39 $\pm$ 0.55 & 56.50 $\pm$ 0.47 & 5.89 \\
        & BDCSPN~\cite{bdcspn} (2022) & 47.60 $\pm$ 0.65 & 45.47 $\pm$ 0.56 & 2.13 & 57.09 $\pm$ 0.52 & 53.17 $\pm$ 0.45 & 3.92 \\
        & PADDLE~\cite{paddle} (2022) & 50.75 $\pm$ 0.64 & 48.85 $\pm$ 0.51 & 1.90 & 61.49 $\pm$ 0.51 & 57.16 $\pm$ 0.48 & 4.33 \\
        & Proto-LP~\cite{protolp} (2023)  & 51.07 $\pm$ 0.90 & 50.27 $\pm$ 0.81 & $0.8$ & 64.89 $\pm$ 0.61 & 59.89 $\pm$ 0.59 & $5.00$ \\
        & ECPE~\cite{ecpe} (2026) & 53.33 $\pm$ 0.75 & 52.04 $\pm$ 0.68 & 1.29 & 65.18 $\pm$ 0.51 & 60.92 $\pm$ 0.50 & 4.26 \\
        \bottomrule
    \end{tabular}}
\end{table*}
\FloatBarrier

\section{Analysis of Shortcut Reliance via Representation Geometry and Background Perturbations}
\label{sec:appendix_extended_analysis}
To better understand whether shortcut reliance is primarily encoded in the backbone representations or induced by the inference head, we perform a systematic head–backbone replacement study. In particular, we examine whether embeddings learned under different training objectives and classifier heads transfer consistently when paired with alternative inference heads at test time. We first analyze inter-family backbone–head swaps, where backbones and heads originate from the same methodological family, and then extend this analysis to cross-family swaps, in which components trained under different learning paradigms are combined. This progression allows us to isolate compatibility effects within families and to assess the robustness of learned representations under more disruptive transfer settings.
\subsection{Fine-Tuning Based \emph{FSC}}
\label{sec:appendix_Fine-Tuning_Based}
\subsubsection{Head--Backbone Replacement Analysis}
\label{sec:appendix_head_backbone_replacement_FT}

We first analyze the interaction between embedding backbones and inference heads by systematically replacing the classifier head used at test time.
In this experiment, rows correspond to the inference head (algorithm), while columns correspond to the backbone used to extract embeddings.
This controlled intervention allows us to probe how different training objectives encode discriminative and spurious information in the learned representations; see Figures~\ref{fig:IID_backbone_algorithm_visualization_conv64f} and~\ref{fig:OOD_backbone_algorithm_visualization_conv64f}.

% \paragraph{Meta-Baseline head $\rightarrow$ Meta-Baseline backbone.}
% This configuration corresponds to the standard Meta-Baseline evaluation protocol.
% Both the backbone and the inference head rely on a standard cross-entropy objective without explicit feature normalization.
% As a result, class discrimination can exploit both angular separation and feature magnitude, including magnitude-based cues induced by dataset-specific background statistics.
% This setting serves as a reference point.

\paragraph{Meta-Baseline head $\rightarrow$ Baseline backbone.}
When the Meta-Baseline head is applied to a Baseline-trained backbone, inference still relies on unnormalized dot-product similarities, while the backbone was optimized in a non-episodic manner.
This mild mismatch slightly reduces reliance on magnitude-based shortcuts learned during Baseline training, resulting in a small performance improvement.

\paragraph{Meta-Baseline head $\rightarrow$ Baseline++ backbone.}
Applying the Meta-Baseline head to a Baseline++ backbone introduces a strong geometric mismatch.
The Baseline++ backbone is trained to produce normalized, directionally discriminative embeddings, whereas the Meta-Baseline head assumes unnormalized features.
Consequently, angular information is not optimally exploited, leading to degraded performance.

\paragraph{Baseline head $\rightarrow$ Meta-Baseline backbone.}
Using a standard Baseline head with a Meta-Baseline backbone causes inference to overemphasize feature magnitude. In Meta-Baseline embeddings, feature norms often correlate with background and dataset-specific artifacts. This increases shortcut reliance and results in a slight performance drop.

% \paragraph{Baseline head $\rightarrow$ Baseline backbone.}
% This setting reflects the canonical Baseline training and evaluation pipeline.
% Both the backbone and the head are optimized under the same cross-entropy objective, allowing the model to exploit both angular and magnitude-based cues, yielding stable reference performance.

\paragraph{Baseline head $\rightarrow$ Baseline++ backbone.}
Combining a Baseline head with a Baseline++ backbone leads to severe performance degradation.
Baseline++ training enforces feature normalization and encodes class separability primarily in angular space, while the Baseline head relies on raw dot products.
This incompatibility prevents effective utilization of the learned representations.

\paragraph{Baseline++ head $\rightarrow$ Meta-Baseline backbone.}
Applying a cosine-based Baseline++ head to a Meta-Baseline backbone yields the largest performance improvement.
The Meta-Baseline backbone produces episodically discriminative embeddings whose feature magnitudes often encode spurious background cues. Feature normalization suppresses these magnitude-based shortcuts and forces decisions to rely on angular similarity, significantly improving few-shot generalization.

\paragraph{Baseline++ head $\rightarrow$ Baseline backbone.}
When applied to a Baseline-trained backbone, the Baseline++ head removes magnitude-based confidence cues that are frequently correlated with background conditions.
Although the backbone was not explicitly trained for angular discrimination, the cosine classifier acts as a regularizer that reduces shortcut exploitation, resulting in a substantial performance gain.

% \paragraph{Baseline++ head $\rightarrow$ Baseline++ backbone.}
% This configuration represents the standard Baseline++ evaluation.
% Both training and inference enforce feature and weight normalization, ensuring that classification depends exclusively on angular similarity and yielding stable performance.

\paragraph{Pairwise backbone replacement analysis (\emph{IID} vs.\ \emph{OOD}).}
Thus far, under \emph{IID} evaluation, we observed that replacing the backbone of simpler baselines (e.g., Baseline and Meta-Baseline) with stronger representations (e.g., Baseline++) consistently leads to accuracy gains, whereas the reverse replacement typically results in noticeable degradation.
This asymmetric behavior suggests that stronger backbones encode more discriminative and transferable features that benefit multiple downstream classifiers, while weaker backbones fail to support more advanced inference mechanisms. We repeat the same analysis for \emph{OOD} tasks, where the query samples contain foreground events that are distributionally shifted from the support set, often due to mismatched background or contextual acoustic conditions. Notably, the qualitative trends observed under IID evaluation persist under \emph{OOD} settings: backbones trained with stronger inductive biases continue to generalize better across algorithms, while weaker backbones exacerbate performance drops. In several cases, the performance gap widens under \emph{OOD} conditions, indicating that models relying more heavily on spurious background correlations are less robust to distribution shifts.

Overall, these results suggest that backbone quality plays a central role in both IID and \emph{OOD} generalization.
In particular, representations that are less sensitive to shortcut features and background-specific cues yield more stable performance when transferred across learning algorithms and evaluation regimes.

\subsubsection{\emph{IID}--\emph{OOD} Background Perturbation Analysis}
\label{sec:appendix_iid_ood_background_perturbation_analysis_FT}
To quantify shortcut reliance during few-shot inference, we evaluate all methods under \emph{IID} and \emph{OOD} episodic settings; see Table~\ref{tab:conv64}. In the \emph{OOD} setting, the foreground sound event of the query remains unchanged, while background sound events are systematically mismatched between the support and query sets.
Therefore, any performance degradation from \emph{IID} to \emph{OOD} cannot be attributed to semantic class ambiguity and directly reflects sensitivity to background--class correlations.

Recall the performance gap as
\[
\Delta = \text{Avg. \emph{IID}} - \text{Avg. \emph{OOD}},
\]
which serves as a measure of background shortcut reliance.

We observe that fine-tuning-based methods exhibit substantially different $\Delta$ values in the 5-shot setting. Baseline shows the largest gap, indicating strong reliance on background consistency.
DiffKendall exhibits a moderate gap, suggesting partial mitigation of shortcut effects.
Meta-Baseline further reduces $\Delta$, reflecting improved robustness due to episodic training.
Baseline++ achieves the smallest gap, consistent with its reliance on cosine similarity and feature normalization.

Notably, the magnitude of $\Delta$ increases with the number of shots, indicating that additional support examples amplify shortcut exploitation when background statistics are consistent within a class.

\subsubsection{Linking Representation Geometry to Shortcut Exploitation}
\label{sec:appendix_linking_representation_geometry_FT}

The head--backbone replacement experiment and the \emph{IID}--\emph{OOD} evaluation provide complementary perspectives on shortcut learning.
The replacement analysis reveals where spurious correlations are encoded in the representation, while the \emph{IID}--\emph{OOD} gap quantifies how strongly these correlations are exploited during few-shot inference.

Taken together, the results indicate that background-induced shortcuts are predominantly encoded in feature magnitudes, whereas foreground semantic information is encoded in feature direction.
Inference heads that rely on unnormalized dot-product similarities are therefore more susceptible to exploiting background consistency, resulting in larger \emph{IID}--\emph{OOD} performance gaps.
Conversely, cosine-based classifiers suppress magnitude-based cues and yield improved robustness under background perturbations.

The strong agreement between these two independent analyses provides converging evidence that magnitude-based decision rules are a primary driver of shortcut reliance in few-shot audio classification under dataset heterogeneity.

\subsection{Meta-Learning-Based \emph{FSC}}
\label{sec:appendix_Meta-Learning_Based}
\subsubsection{Head--Backbone Replacement Analysis (\emph{IID} Baseline)}
\label{sec:appendix_head_backbone_replacement_analysis_meta}
\begin{figure}[!htbp]
    \centering
        \includegraphics[width=\linewidth]{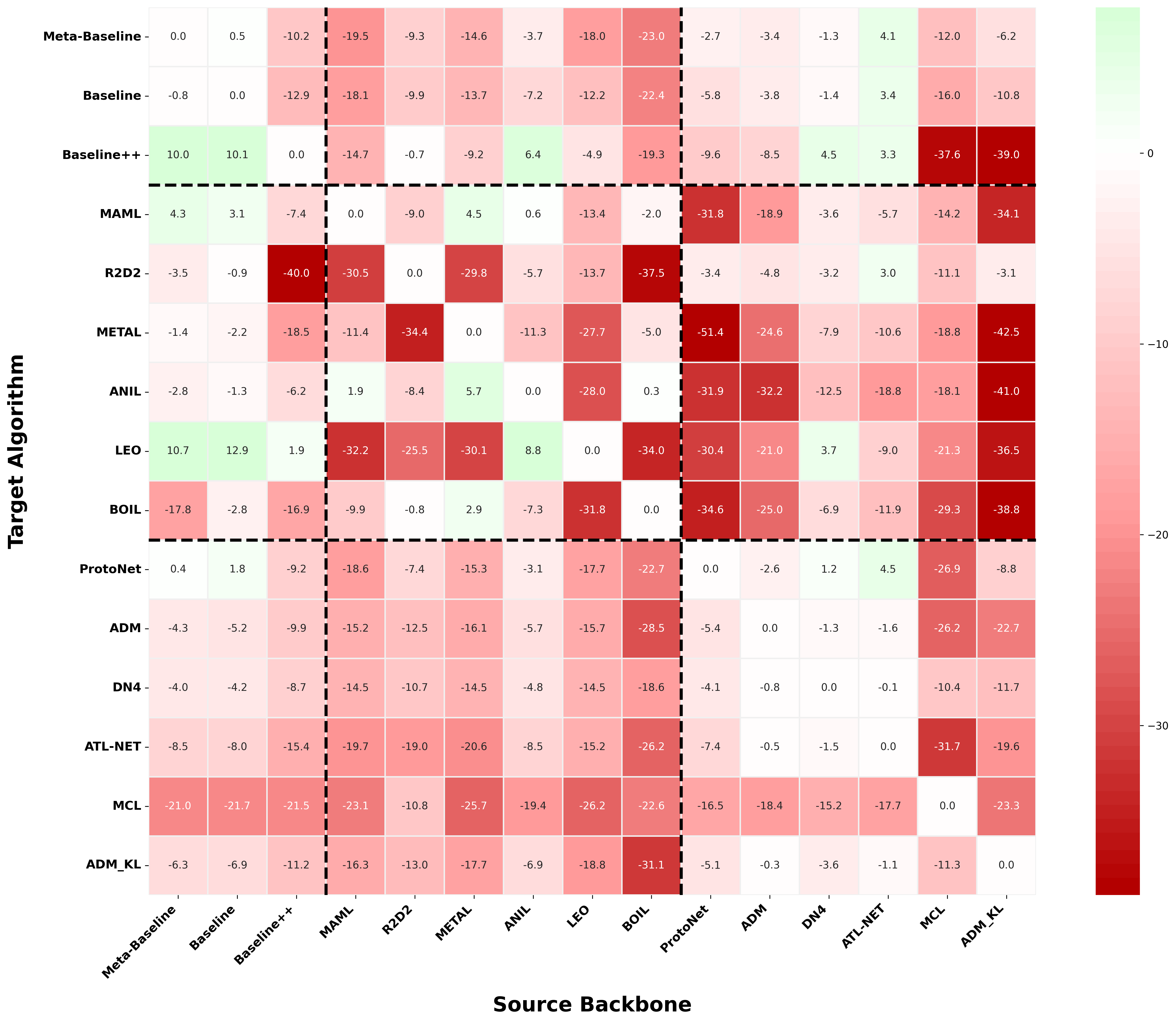}
        \caption{\emph{IID} configuration, multiple Conv64F backbones on multiple algorithms}
        \label{fig:IID_backbone_algorithm_visualization_conv64f}
    \end{figure}

\begin{figure}[!htbp]
        \centering

        \includegraphics[width=\linewidth]{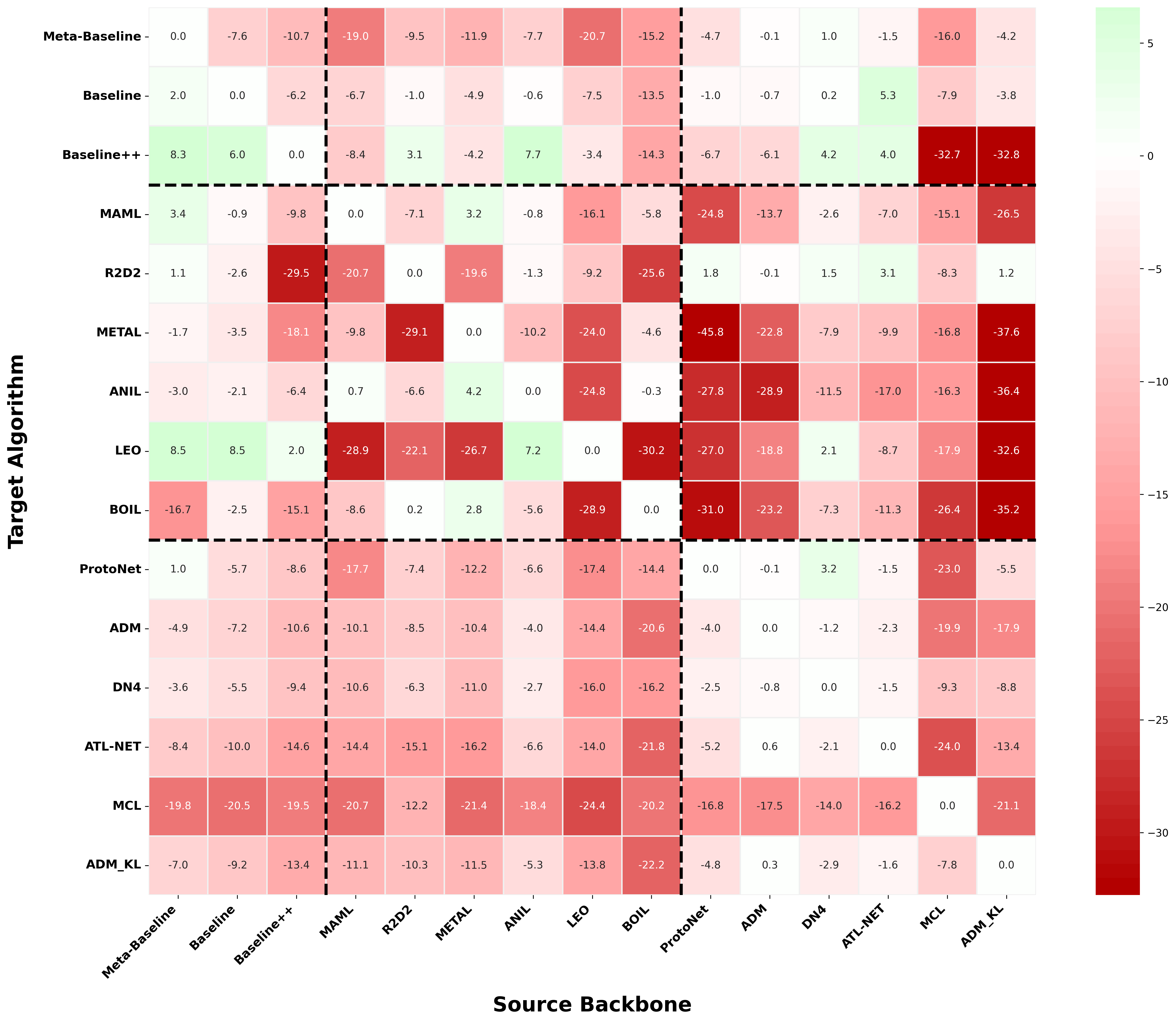}
        \caption{\emph{OOD} configuration, multiple Conv64 backbones on multiple algorithms}
        \label{fig:OOD_backbone_algorithm_visualization_conv64f}   
\end{figure}

This section analyzes the compatibility between the classification heads (rows) and feature backbones (columns) of different meta-learning algorithms; e.g., Figure~\ref{fig:IID_backbone_algorithm_visualization_conv64f}. Our aim here is to link the performance impact ($\Delta$) to the compatibility between the \textit{head's inference mechanism} and the \textit{backbone's learned geometry}.

\paragraph{MAML Head (Gradient-Based Adaptation)}
The MAML head initializes a linear classifier $w$ and performs a small number of gradient descent steps
($w' \leftarrow w - \alpha \nabla_w \mathcal{L}_{CE}$). Rather than requiring features to be linearly separable
\emph{a priori}, MAML is trained to produce representations that become linearly separable after a few gradient updates.

\begin{itemize}
    \item \textbf{vs. METAL Backbone:} METAL optimizes a distance-based objective that encourages compact class
    representations with reduced intra-class variance. Such geometrically regularized features can yield
    better-conditioned gradients for cross-entropy optimization, which may facilitate faster or more stable
    adaptation of the MAML head compared to less structured feature spaces.
   \item \textbf{vs. ANIL \& BOIL:} Performance remains largely neutral with a slight degregation in BOIL. As close variants of MAML, ANIL and BOIL induce feature spaces with similar local geometry, allowing the MAML head to transfer without significant difficulty, albeit without the additional geometric regularization benefits provided by METAL.
    \item \textbf{vs. R2D2:} R2D2 optimizes a ridge regression objective ($\lVert Xw - y \rVert^2 + \lambda \lVert w \rVert^2$), which permits substantial variation in feature scale and anisotropic covariance structure. When paired with such features, MAML's cross-entropy objective can exhibit poor conditioning, resulting in less stable or slower gradient-based adaptation. 
    \item \textbf{vs. LEO:} LEO backbones are trained to support latent-space weight generation via a variational encoder ($z \sim \mathcal{N}(\mu,\sigma)$), rather than direct linear separability in feature space. This induces representations whose geometry is aligned with latent inference, limiting the effectiveness of MAML's direct gradient-based linear adaptation.
\end{itemize}

\paragraph{R2D2 Head (Ridge Regression)}
The R2D2 head computes a closed-form ridge regression solution
$W = (X^T X + \lambda I)^{-1} X^T Y$, which requires the feature covariance matrix $X^T X$ to be reasonably
well-conditioned.
\begin{itemize}
    \item \textbf{vs. BOIL:} \textbf{Severe degradation.} BOIL freezes the head and adapts only the backbone,
    effectively forcing the feature extractor to fit a fixed, randomly initialized classifier. This can
    induce highly distorted or collapsed feature representations, leading to poorly conditioned or
    near-rank-deficient covariance matrices and unstable regression solutions.
    \item \textbf{vs. METAL \& MAML:} METAL and MAML do not explicitly optimize features for linear regression
    consistency. METAL emphasizes distance-based clustering, while MAML emphasizes rapid adaptability under
    cross-entropy loss. The resulting feature representations may exhibit anisotropic variance or weak
    linear correlations with labels, which can degrade the performance of a ridge regression solver.
    \item \textbf{vs. LEO:} LEO backbones are trained to support latent-space weight generation under a Gaussian
    prior rather than direct linear prediction in feature space. As a result, the induced representations
    are not well aligned with the assumptions underlying linear regression, limiting the stability of the
    R2D2 solution.
    \item \textbf{vs. ANIL:} ANIL produces generic, static feature representations optimized for linear
    classification rather than regression. While these features may not be optimally conditioned for
    ridge regression, their stability is typically sufficient to avoid numerical failure.
\end{itemize}

\paragraph{METAL Head (Task-Dependent Metric Scaling)}
The METAL head computes class probabilities using a scaled Euclidean distance,
$P(y \mid x) \propto \exp\!\left(-\alpha\, d(x,c_k)\right)$, where $\alpha$ is a learnable task-dependent
scaling parameter.
\begin{itemize}
    \item \textbf{vs. BOIL:} METAL is relatively robust to BOIL-induced feature irregularities, as the learnable
    scaling parameter $\alpha$ can partially compensate for distorted or unevenly scaled feature dimensions.
    \item \textbf{vs. MAML \& ANIL:} MAML and ANIL backbones are trained to optimize dot-product-based logits under
    cross-entropy loss. The resulting angularly structured representations are not optimally aligned with
    METAL's distance-based inference, leading to a degradation due to geometric mismatch.
    \item \textbf{vs. R2D2 \& LEO:} R2D2 (regression-oriented) and LEO (latent-variable-based) backbones induce
    feature geometries that are not explicitly optimized for Euclidean distance comparisons, resulting in
    reduced effectiveness of METAL's metric-based classification.
\end{itemize}

\paragraph{ANIL Head (Feature Reuse / Linear Classifier)}
ANIL freezes the backbone $f_\theta$ and trains a linear classifier $w$ on top. It serves as a proxy for evaluating the quality of static features.
\begin{itemize}
    \item \textbf{vs. METAL Backbone:} ANIL shows strong transfer. Since ANIL does not adapt the backbone, it benefits most from features that are already well-separated; METAL provides tightly clustered, ready-to-use features.
    \item \textbf{vs. MAML:} MAML's backbone produces features of sufficient quality that even without adaptation, ANIL's linear classifier can achieve competitive performance.
    \item \textbf{vs. BOIL:} BOIL backbones are trained with body-only adaptation, yielding expressive features that can support ANIL's frozen head without additional adaptation.
    \item \textbf{vs. R2D2:} R2D2 features are optimized for regression objectives, emphasizing scale consistency rather than purely linear separability, which may slightly reduce ANIL’s effectiveness.
    \item \textbf{vs. LEO:} LEO features are latent-space optimized; their probabilistic structure is not directly aligned with ANIL’s linear classifier.
\end{itemize}

\paragraph{LEO Head (Latent Encoding Optimization)}
LEO generates parameters for a linear classifier via a latent space with Gaussian prior $p(z) = \mathcal{N}(0, I)$.
\begin{itemize}
    \item \textbf{vs. ANIL:} LEO can benefit from stationary, frozen features such as ANIL's, as they provide a consistent input distribution for latent encoding.
    \item \textbf{vs. MAML:} High-variance adaptive features from MAML may be misaligned with LEO's latent encoding, potentially reducing transfer effectiveness.
    \item \textbf{vs. R2D2:} Regression-optimized features from R2D2 are not specifically aligned with LEO’s Gaussian latent prior, which may limit performance.
    \item \textbf{vs. METAL \& BOIL:} Features strongly constrained by geometric clustering or backbone-specific adaptation may not be ideally aligned with LEO’s probabilistic latent representation.
\end{itemize}

\paragraph{BOIL Head (Body-Only Adaptation)}
BOIL freezes the head and updates backbone parameters via gradient descent.
\begin{itemize}
    \item \textbf{vs. METAL:} Starting adaptation from METAL's clustered features can provide a favorable initialization for gradient updates compared to random initialization.
    \item \textbf{vs. R2D2:} Features with low intra-class variance, such as R2D2’s, may be relatively stable and support body-only adaptation without significant feature distortion.
    \item \textbf{vs. MAML \& ANIL:} These backbones may require more extensive adaptation than is feasible in a few gradient steps under a frozen head, limiting BOIL’s performance.
    \item \textbf{vs. LEO:} When using a LEO backbone, BOIL exhibits severe performance degradation, as LEO relies on task-specific latent-space optimization and decoder-based parameter generation, making body-only gradient updates under a frozen head poorly aligned with its intended adaptation mechanism.

\end{itemize}

\subsubsection{\emph{IID}--\emph{OOD} Background Perturbation Analysis}
\label{sec:appendix_iid_ood_background_perturbation_analysis_meta}
This section examines the 5-shot performance gap ($\Delta$) between IID (matched backgrounds) and \emph{OOD} (mismatched or spurious backgrounds).

\begin{itemize}
    \item \textbf{MAML vs. ANIL:} MAML generally exhibits a smaller drop in performance compared to ANIL. This aligns with the principle that full network adaptation can mitigate reliance on spurious correlations. By updating both backbone and head on the support set, MAML can adjust feature representations that are correlated with background noise, whereas ANIL freezes the backbone, retaining any pre-trained shortcuts.
    
    \item \textbf{R2D2 Sensitivity:} R2D2 often achieves high \emph{IID} accuracy but experiences a notable drop under \emph{OOD} conditions. Since R2D2 computes a closed-form classifier on fixed embeddings, it cannot dynamically reweigh features during few-shot updates, making it similarly sensitive to background shifts as ANIL.
    
    \item \textbf{METAL Robustness:} METAL typically shows high accuracy with minimal degradation under \emph{OOD}. Its task-dependent metric scaling encourages class prototypes to form compact clusters, reducing the influence of background variability and improving robustness to distribution shifts.
\end{itemize}

\subsubsection{Linking Representation Geometry to Shortcut Exploitation}
\label{sec:appendix_linking_representation_geometry_meta}
\begin{itemize}
    \item \textbf{Feature Reuse (ANIL/R2D2) vs. Feature Adaptation (MAML):} Algorithms that rely on feature reuse are more susceptible to spurious background correlations because static backbones encode such correlations permanently. Feature adaptation, as in MAML, allows the backbone to adjust the feature manifold according to the support set, mitigating the effect of spurious correlations when foreground and background are mismatched.
    
    \item \textbf{Gradient-Based ``Unlearning'':} MAML's gradient updates on the backbone can suppress features associated with spurious correlations. BOIL shares this potential in principle, but its entangled head-backbone optimization limits effective adaptation in practice.
\end{itemize}

\subsubsection{\emph{OOD} Heatmap Comparison: Shifts in Robustness}
\label{sec:appendix_ood_heat_map_comparison}
Comparing \emph{IID} and \emph{OOD} heatmaps highlights differences in backbone robustness.

\begin{itemize}
    \item \textbf{R2D2 Backbone with MAML Head:} While this pairing shows reduced performance in \emph{IID} settings, the gap diminishes under \emph{OOD} conditions.
    \begin{itemize}
        \item \textit{Mechanistic Interpretation:} R2D2's regression objective produces regularized, low-variance feature embeddings. In \emph{IID}, this regularization can restrict adaptation under MAML. In \emph{OOD}, the same regularization stabilizes the model by limiting overfitting to background noise.
    \end{itemize}
    
    \item \textbf{METAL Backbone:} METAL consistently demonstrates strong performance under OOD.
    \begin{itemize}
        \item \textit{Mechanistic Interpretation:} METAL enforces geometric compactness of class prototypes. This tight clustering reduces susceptibility to spurious background correlations, providing robustness in both \emph{IID} and \emph{OOD} settings.
    \end{itemize}
    
    \item \textbf{Persistent Architectural Mismatches:} Combinations such as LEO Head with MAML Backbone or R2D2 Head with BOIL Backbone show consistently low performance across \emph{IID} and \emph{OOD} conditions. These failures are attributable to fundamental mathematical incompatibilities between head and backbone, rather than dataset-specific effects.
\end{itemize}

\subsection{Metric-Based \emph{FSC}}
\label{sec:appendix_Metric_Based}

This section analyzes the compatibility between the classification heads (rows) and feature backbones (columns) of Metric-based algorithms in the In-Distribution (\emph{IID}) setting.

\subsubsection{Head--Backbone Replacement Analysis (\emph{IID} Baseline)}
\label{sec:appendix_head_backbone_replacement_analysis_metric}

\paragraph{ProtoNet Head (Euclidean Centroid)}
The ProtoNet head computes class centroids via Global Average Pooling (GAP) and classifies using Euclidean distance.

\begin{itemize}
    \item \textbf{vs. DN4:} The ProtoNet head benefits from these backbones. DN4 preserves rich local descriptors, which, after pooling, create a denser and more discriminative feature space than ProtoNet's native training.

    \item \textbf{vs. ATL-NET } ATL-NET employs episodic attention, resulting in feature embeddings that highlight discriminative regions, which worsens slightly the ProtoNet centroid classification.
    
    \item \textbf{vs. MCL:} Performance drops significantly. MCL optimizes a ranking-based objective that alters the Euclidean structure of the feature space. ProtoNet assumes isotropic Gaussian clusters; this assumption is violated when features exhibit non-Euclidean geometry induced by MCL.
\end{itemize}

\paragraph{ADM Head (Adaptive Distance Metric)}
The ADM head learns a task-specific distance metric, such as a Mahalanobis-like matrix, for similarity measurement.

\begin{itemize}
    \item \textbf{vs. All Foreign Backbones:} ADM is highly sensitive to backbone changes. Its learned metric relies on specific feature covariance structures. Backbones not trained with ADM (e.g., ProtoNet, DN4) do not provide the expected correlations, resulting in performance degradation. The effect is most pronounced with MCL, indicating a geometric incompatibility between ADM's learned metric and MCL feature space.
\end{itemize}

\paragraph{DN4 Head (Image-to-Class / Local Descriptors)}
DN4 operates on unpooled feature maps using a local neighbor matching procedure rather than global feature vectors.

\begin{itemize}
    \item \textbf{vs. ProtoNet \& ATL-NET:} Accuracy decreases when paired with these backbones. ProtoNet and ATL-NET optimize global pooled representations, which do not necessarily preserve the spatial locality and fine-grained information required by DN4’s image-to-class matching. Consequently, the DN4 head receives embeddings that are less compatible with its matching mechanism, resulting in lower performance relative to its native backbone.
\end{itemize}

\paragraph{MCL Head}
MCL is trained to optimize a ranking-based or curriculum-based objective in the feature space.

\begin{itemize}
    \item \textbf{vs. All Foreign Backbones:} MCL exhibits performance degradation with any non-native backbone. This indicates that MCL’s learned feature manifold has a specialized geometry, which is incompatible with the Euclidean or local-matching assumptions used by standard metric-based heads such as ProtoNet or DN4.
\end{itemize}

\subsubsection{\emph{IID}--\emph{OOD} Background Perturbation Analysis}
\label{sec:appendix_iid_ood_background_perturbation_analysis_metric}
This section examines the performance gap between \emph{IID} (matched backgrounds) and \emph{OOD} (mismatched/spurious backgrounds).

\begin{itemize}
    \item \textbf{Robustness of DN4:} DN4 maintains high \emph{OOD} accuracy and one of the smallest performance drops. This supports the \textit{Local Feature Hypothesis}. By comparing sets of local descriptors rather than a single global vector, DN4 can effectively focus on foreground patches that match the query while ignoring background patches that do not, providing intrinsic robustness to background shifts.
    \item \textbf{ProtoNet Sensitivity:} ProtoNet experiences a substantial drop in OOD performance. Its reliance on Global Average Pooling (GAP) aggregates both foreground and background into a single vector. Background changes in the \emph{OOD} setting shift this vector away from the class centroid, leading to misclassification. ProtoNet lacks a mechanism to spatially filter out background features.
    \item \textbf{ATL-NET and Attention:} ATL-NET also shows a large \emph{OOD} performance drop despite using an attention mechanism. In \emph{IID} training, attention can leverage background correlations, which act as a shortcut. In \emph{OOD} conditions, these correlations are misleading, reducing accuracy because the attention mechanism cannot fully suppress background contributions.
\end{itemize}

\subsubsection{Linking Representation Geometry to Shortcut Exploitation}
\label{sec:appendix_linking_representation_geometry_metric}
\begin{itemize}
    \item \textbf{Global vs. Local Geometry:} Heads relying on \textbf{Global Pooling} (ProtoNet, ATL-NET) are sensitive to spurious correlations because signal and noise are combined into a single representation. Heads using \textbf{Local Descriptors} (DN4) preserve spatial separation of signal and noise, allowing the model to selectively match foreground features while ignoring irrelevant background.
    \item \textbf{The ``Attention'' Trap:} Large drops from \emph{IID} to \emph{OOD} indicate that attention mechanisms may overfit to background correlations present during training, rather than learning invariance to them.
\end{itemize}

\subsubsection{\emph{OOD} Heatmap Comparison: Shifts in Robustness}
\label{sec:ood_heatmap_comparison_metric}
Comparing the \emph{OOD} heatmap to the \emph{OOD} heatmap highlights shifts in backbone-head transferability under distributional changes.

\begin{itemize}
    \item \textbf{The ``ATL-NET Flip'' (ProtoNet Head):} In \emph{IID}, the ProtoNet head benefits from ATL-NET features, which encode both object and context. Under OOD conditions, the contextual correlation becomes misleading. ProtoNet, relying on pooled global representations, cannot disentangle object from background, resulting in reduced transfer performance.
    \item \textbf{DN4 Consistency:} DN4 consistently provides positive transfer to ProtoNet across \emph{IID} and \emph{OOD}. DN4 features are trained for local matching, creating a representation in which object and background are disentangled. Even when aggregated by a global pooling head, these features preserve robustness to background shifts.
    \item \textbf{MCL Isolation Persists:} MCL continues to exhibit negative transfer. The geometric incompatibility of its feature manifold with standard metric heads remains unchanged by the distribution shift, indicating a structural rather than data-dependent failure.
\end{itemize}

\subsection{Cross-Family Analysis of Few-Shot Learning Algorithms (\emph{IID} Setting)}
\label{sec:appendix_Cross-Family_Analysis_of_Few-Shot_Learning_Algorithms_IID}

This section analyzes the interactions between the three major families: \textbf{Finetuning-based} (Baseline, Baseline++, Meta-Baseline), \textbf{Meta-learning-based} (MAML, R2D2, METAL, ANIL, LEO, BOIL), and \textbf{Metric-based} (ProtoNet, ADM, DN4, ATL-NET, MCL, ADM\_KL). The analysis focuses on the geometric compatibility between the backbone's feature space and the head's inference mechanism in the \emph{IID} setting.

\begin{itemize}
    \item \textbf{Finetuning heads on Metric backbones:}
    Finetuning heads (typically linear classifiers like Softmax or Cosine classifiers) rely on the backbone producing features that are linearly separable. Metric backbones (like ProtoNet, MCL, ADM) are trained to optimize local clustering structure (minimizing intra-class distance, maximizing inter-class distance) often using Euclidean or KL-divergence metrics. They do not necessarily force the global linear separability required by a standard finetuning head. Consequently, the linear head cannot find a valid decision boundary in the "clustered" but potentially non-linear embedding space provided by metric backbones. Slight improvements where shown on the cosine classifier when using the strong feature representation of the DN4 and ATL-NET backbones showcased improvement on all finetuning heads.

    \item \textbf{Finetuning heads Meta-learning:}
      Applying Finetuning backbones (Baseline, Baseline++, Meta-Baseline) to Meta-learning heads reveals a sharp performance dichotomy driven by the compatibility between feature geometry and adaptation mechanisms. Heads that decouple feature extraction from task-specific adaptation, particularly ANIL, achieve their highest accuracy gains with standard backbones  with Baseline++), suggesting that robust, globally separable features provide a superior, "honest" foundation for flexible parameter generators compared to meta-learned features potentially corrupted by episode-specific shortcuts. Conversely, this positive transfer collapses when geometric assumptions clash: the cosine-constrained Baseline++ backbone is catastrophic for regression and optimization-based heads, causing massive drops for R2D2 and MAML  due to the incompatibility between angular embedding spaces and Euclidean or covariance-based update rules, while body-updating methods like BOIL consistently degrade because converged pre-trained backbones lack the malleability required for rapid inner-loop adaptation.
  
    \item \textbf{Meta-Learning Backbones on Metric Heads:}
      Replacing the native backbones of Metric heads (ProtoNet, MCL, ADM) with Meta-Learning backbones (MAML, BOIL, METAL) results in universal and severe degradation, highlighting a fundamental conflict between "adaptation-ready" and "inference-ready" geometries. Metric heads rely on static, tightly clustered embeddings to calculate distances (e.g., Euclidean or Cosine) without further training; however, meta-learning backbones are trained to produce "malleable" initializations that only become discriminative after gradient updates (inner-loop adaptation). Consequently, when a static Metric head is applied to a "raw" meta-learning backbone like BOIL or METAL, it encounters feature spaces that are not yet linearly or spherically separated, leading to catastrophic failures proving that meta-learning optimization objectives do not inherently align with the distance-minimization manifolds required by metric classifiers.

    \item \textbf{Meta-Learning Backbones on Finetuning Heads:}
      The application of Finetuning heads to Meta-Learning backbones is largely unsuccessful, with a notable exception for the cosine-based Baseline++ head, revealing that only specific meta-learning strategies produce globally separable features. Most meta-learning backbones (especially BOIL, MAML and METAL) cause significant drops for standard linear heads.
      However, Baseline++ achieves positive transfer with ANIL and a very slight performance drop in R2D2.
      This success occurs because ANIL explicitly learns a fixed feature extractor (compatible with fixed heads), and R2D2 optimizes a ridge-regression objective that aligns geometrically with the angular/cosine margins of Baseline++. This indicates that for a meta-learned backbone to be transferable to a standard classifier without further adaptation, it must be trained with constraints (like ANIL's frozen body or R2D2's regression) that enforce a stable, normalized embedding space compatible with cosine similarity.

      \item \textbf{Metric backbones on Meta-learning heads:}
      The replacement of meta-learning backbones with Metric backbones (ProtoNet, ADM, MCL) results in almost universal failure, driven by the incompatibility between the "static clustering" objective of metric learning and the "dynamic adaptation" requirement of meta-learning. 

      \item \textbf{Metric backbones on finetuning heads:}
        Applying Finetuning heads to Metric backbones yields a mixed but generally stable outcome, contrasting sharply with the failures seen in meta-learning heads, because the linear separability enforced by metric objectives is often sufficient for static linear classifiers. Standard Baseline heads show a slight moderate loss or slight gains and the cosine-based Baseline++ achieves significant boosts suggesting that these specific backbones produce well-normalized, globally separable manifolds that align perfectly with fixed cosine classifiers. ATL-NET produces positive results on all finetuning methods.
      
\end{itemize}
    
\subsection{Cross-Family Analysis of Few-Shot Learning Algorithms (\emph{OOD} Setting)}
\label{sec:appendix_Cross-Family_Analysis_of_Few-Shot_Learning_Algorithms_OOD}

This section analyzes the interactions between the three major families: \textbf{Finetuning-based} (Baseline, Baseline++, Meta-Baseline), \textbf{Meta-learning-based} (MAML, R2D2, METAL, ANIL, LEO, BOIL), and \textbf{Metric-based} (ProtoNet, ADM, DN4, ATL-NET, MCL, ADM\_KL). The analysis focuses on the geometric compatibility between the backbone's feature space and the head's inference mechanism in the \emph{OOD} setting.

\begin{itemize}
    \item \textbf{Metric backbones on finetuning heads:}
    In the \emph{OOD} setting, replacing the native heads of Metric backbones with standard Finetuning heads proves to be surprisingly robust. Unlike other cross-family transfers that suffer from severe degradation, standard linear classifiers (Baseline) and cosine classifiers (Baseline++) often maintain or even improve performance when applied to backbones like ProtoNet, DN4, and ATL-NET. This suggests that the embedding geometry learned by most metric algorithms is not "twisted" or overly specialized, but rather creates clean, globally separable clusters that even a simple linear boundary can respect, effectively filtering out perturbed background noise.
    
    \item \textbf{Metric backbones on Meta-Learning heads:}
    The replacement of meta-learning backbones with metric backbones results in universal degradation across the board. Meta-learning algorithms typically require the feature space to be "malleable" so that a few gradient steps can align the support and query sets. In the \emph{OOD} scenario, this rigidity is fatal; the backbone has effectively "memorized" the specific background-to-class relationships of the training data. Because the meta-learning head cannot reshape this rigid geometry sufficiently during the inner loop, it is forced to operate on features where the background acts as a misleading distraction, leading to consistent performance drops.

    \item \textbf{Finetuning backbones on Metric heads:}
    A significant shift occurs here when moving from \emph{IID} to \emph{OOD}. the varying backgrounds introduce large distance penalties—effectively "pushing" the query samples away from their correct class prototypes. Unlike linear heads which can learn to ignore certain dimensions, distance-based heads aggregate error from all dimensions, including the perturbed background noise.

    \item \textbf{Finetuning backbones on Meta-Learning heads:}
    This interaction remains the most robust, particularly for methods that decouple feature extraction from adaptation. When a standard pre-trained backbone is used, it tends to learn robust, global representations of the foreground objects (as it must distinguish all classes simultaneously)

    \item \textbf{Meta-learning backbones on Finetuning heads:}
    The application of Finetuning heads to Meta-Learning backbones reveals a sharp divide based on how the backbone was trained. Backbones from algorithms that update the entire network during the inner loop (like BOIL or METAL) perform poorly with standard heads because their pre-adaptation features are not yet linearly separable—they rely on the adaptation step to align the geometry. However, backbones from algorithms that explicitly freeze feature extraction (like ANIL) or use regression constraints (like R2D2) transfer remarkably well to Cosine-based heads (Baseline++). This indicates that meta-learning strategies which enforce a stable, normalized embedding space can produce features that are robust to background perturbations.

    \item \textbf{Meta-learning backbones on Metric heads:}
    applying Metric heads to Meta-Learning backbones results in consistent performance degradation across the board. This failure stems from a conflict in optimization objectives.
    In the \emph{OOD} context, where background noise changes, the un-adapted meta-learning backbone likely has not yet separated the foreground from the background, causing the metric head to calculate distances based on spurious noise rather than semantic class identity, leading to widespread failure.
    
\end{itemize}

\subsection{Cross-Family Analysis of Few-Shot Learning Algorithms in Deeper Architectures}
\label{sec:appendix_Cross-Family_Analysis_of_Few-Shot_Learning_Deeper_Algorithms}

The transition from Conv64 to ResNet12 reveals a fundamental shift towards higher feature robustness, characterized by a marked reduction in the severe geometric incompatibilities (deep red blocks in the figure) seen in Conv64,
and the emergence of FRN as a "universal" head. However, this deeper architecture exposes a specific vulnerability in ANIL, which flips from a generally compatible head in Conv64 to suffering catastrophic failures on ResNet12 backbones, indicating that deeper, more complex feature hierarchies require the full body adaptation that ANIL forbids. Crucially, ResNet12 effectively better neutralizes the spurious correlation gap that plagued Conv64; while Conv64 showed degradation under \emph{OOD} background perturbation, ResNet12 displays remarkable stability and improvement by switching backbones.
proving that the deeper residual representations successfully disentangle foreground semantics from background noise, making the model geometrically less invariant to the \emph{OOD} shifts that broke the shallower Conv64 architectures.


\begin{figure}[!htbp]
    \centering
        \includegraphics[width=\linewidth]{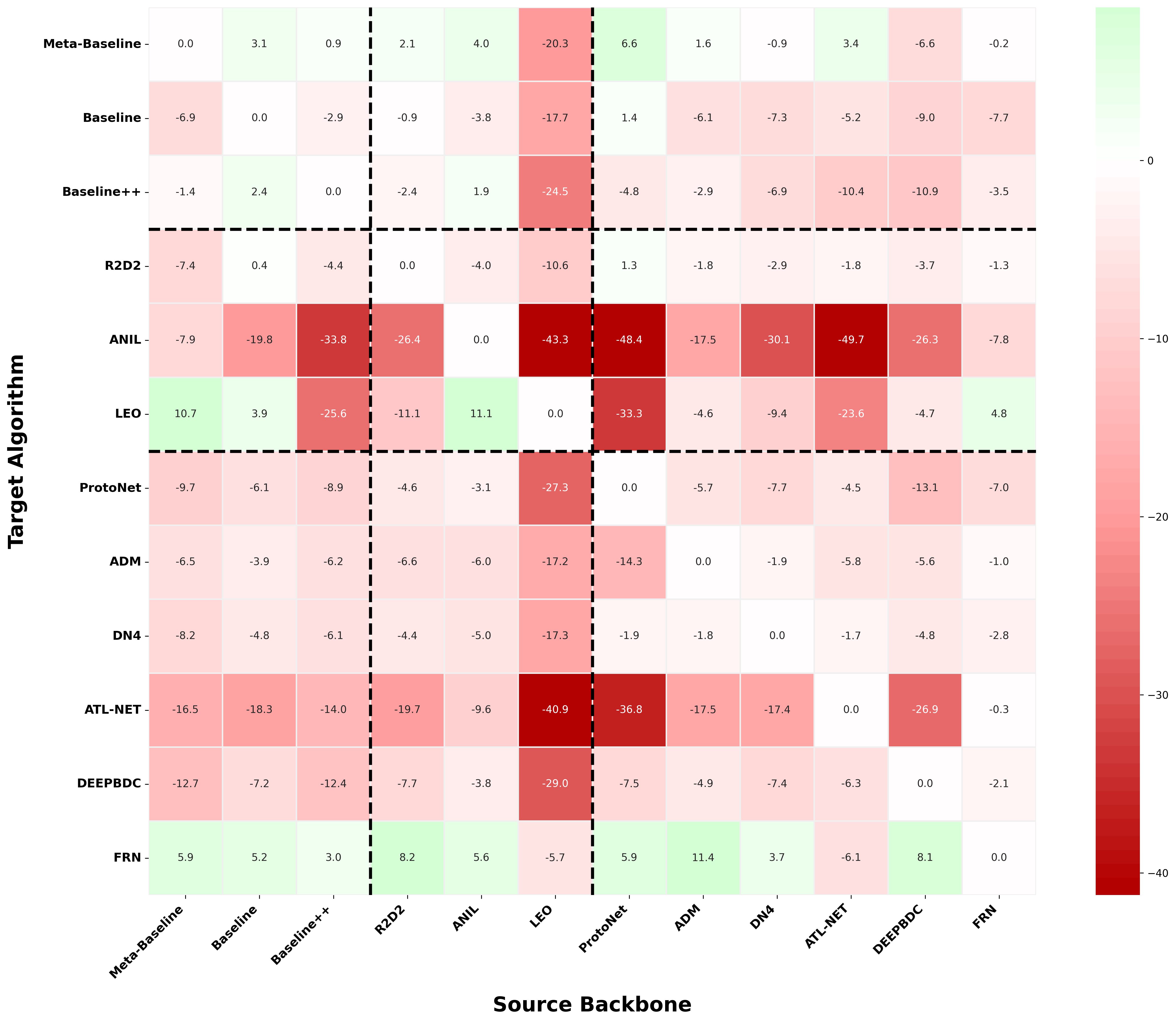}
        \caption{\emph{IID} configuration, multiple Resnet12 backbones on multiple algorithms}
        \label{fig:IID_backbone_algorithm_visualization_resnet12}
    \end{figure}
    \begin{figure}[!htbp]
        \centering

        \includegraphics[width=\linewidth]{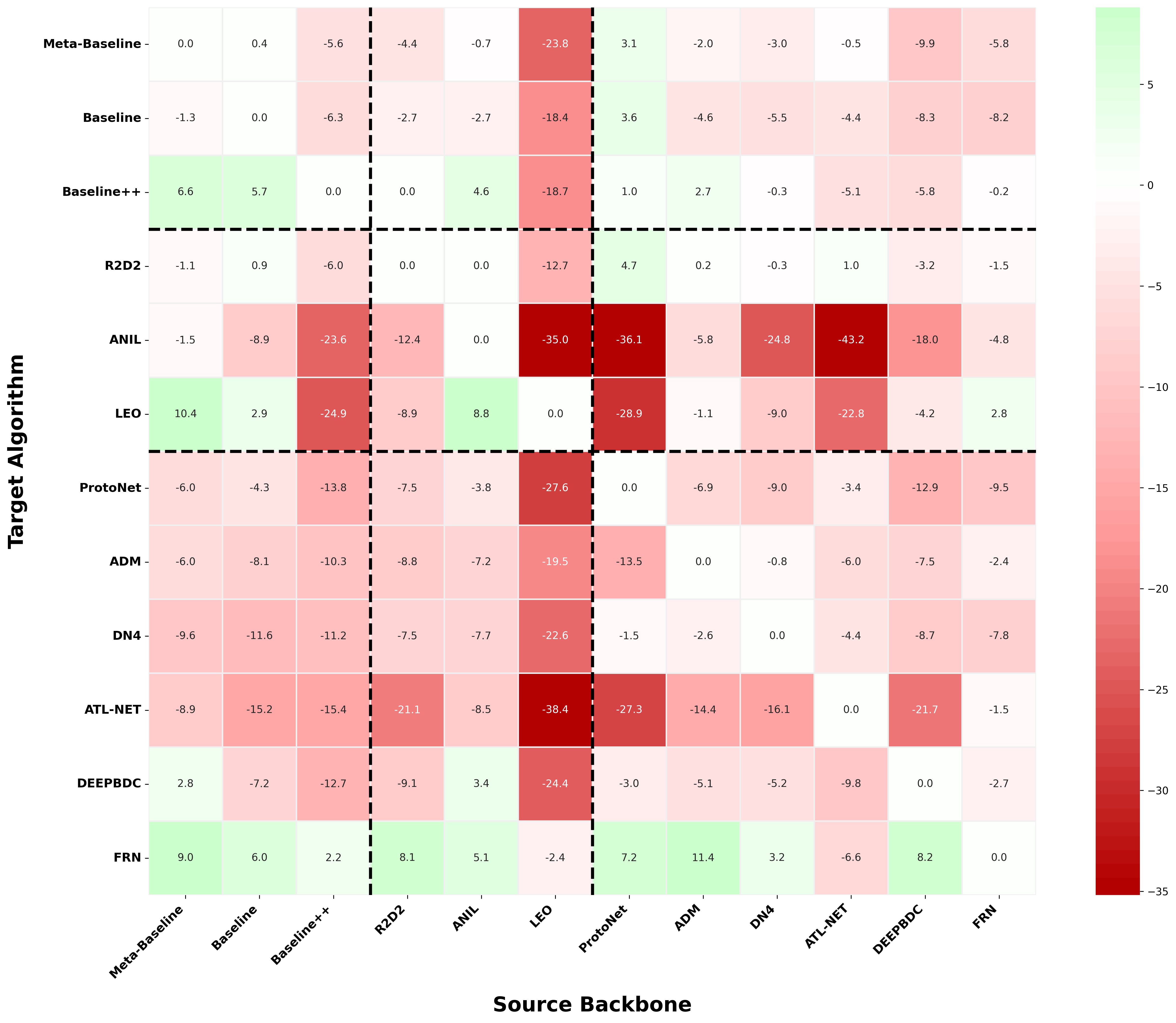}
        \caption{\emph{OOD} configuration, multiple Resnet12 backbones on multiple algorithms}
        \label{fig:OOD_backbone_algorithm_visualization_resnet12}   
\end{figure}

\begin{takeawaybox}{}
\begin{itemize}
    \item Cosine-based classifiers are the most robust head family in Conv64, because they suppress the magnitude-encoded background shortcut that magnitude-sensitive heads (Euclidean ProtoNet, dot-product Baseline) exploit.
    \item Swapping a cosine head onto a magnitude-sensitive backbone (Meta-Baseline, Baseline) yields the largest single robustness gain observed in the head–backbone replacement study; the reverse swap yields the largest drop.
    \item This robustness holds across backbone depths (Conv64, ResNet12, ResNet18): scaling the backbone alone does not close the gap, confirming that magnitude-invariant inference, i.e., not deeper representations, is the decisive factor.
\end{itemize}
\end{takeawaybox}

% \subsection{Concerning Resnet18 embedding model}
% \label{sec:appendix_embedding_model_algorithm_confusion_matrix_resnet18}
\clearpage

\section{Geometric Disentanglement of Background Correlations}
\label{sec:appendix_geometry_analysis}

To investigate how feature extractors encode spurious background correlations, we perform a geometric decomposition of the feature space. We hypothesize that CNN backbones naturally disentangle semantic foreground information from additive background noise by encoding the former in the \textit{angular} direction and the latter in the \textit{feature magnitude}.

\subsection{Methodology: Radial-Angular Decomposition}
\label{sec:appendix_radial_angular_decoposition}

Let $f_\theta: \mathcal{X} \rightarrow \mathbb{R}^d$ denote the backbone feature extractor. For a given input $x$, we decompose the resulting feature embedding $\mathbf{z} = f_\theta(x)$ into two components:
\begin{enumerate}
    \item \textbf{Magnitude (Radial Component):} $r = \|\mathbf{z}\|_2$, representing the activation intensity or "signal energy."
    \item \textbf{Direction (Angular Component):} $\hat{\mathbf{z}} = \frac{\mathbf{z}}{\|\mathbf{z}\|_2}$, representing the semantic identity on the hypersphere $\mathbb{S}^{d-1}$.
\end{enumerate}

To quantify the semantic alignment, we compute the \textit{Cosine Similarity} between a sample's direction $\hat{\mathbf{z}}$ and its corresponding clean (without backgrounds) class prototype $\mathbf{p}_c$. The prototype is defined as the mean direction of the clean, foreground-only samples for class $c$:
\begin{equation}
    \mathbf{p}_c = \text{Normalize}\left( \frac{1}{|\mathcal{S}_{clean}|} \sum_{x_i \in \mathcal{S}_{clean}} f_\theta(x_i) \right)
\end{equation}

\subsection{Empirical Observation: The "Magnitude Contraction" Phenomenon}
\label{sec:appendix_magnitude_contraction}
\begin{figure*}[t]
    \centering
    \includegraphics[width=\textwidth]{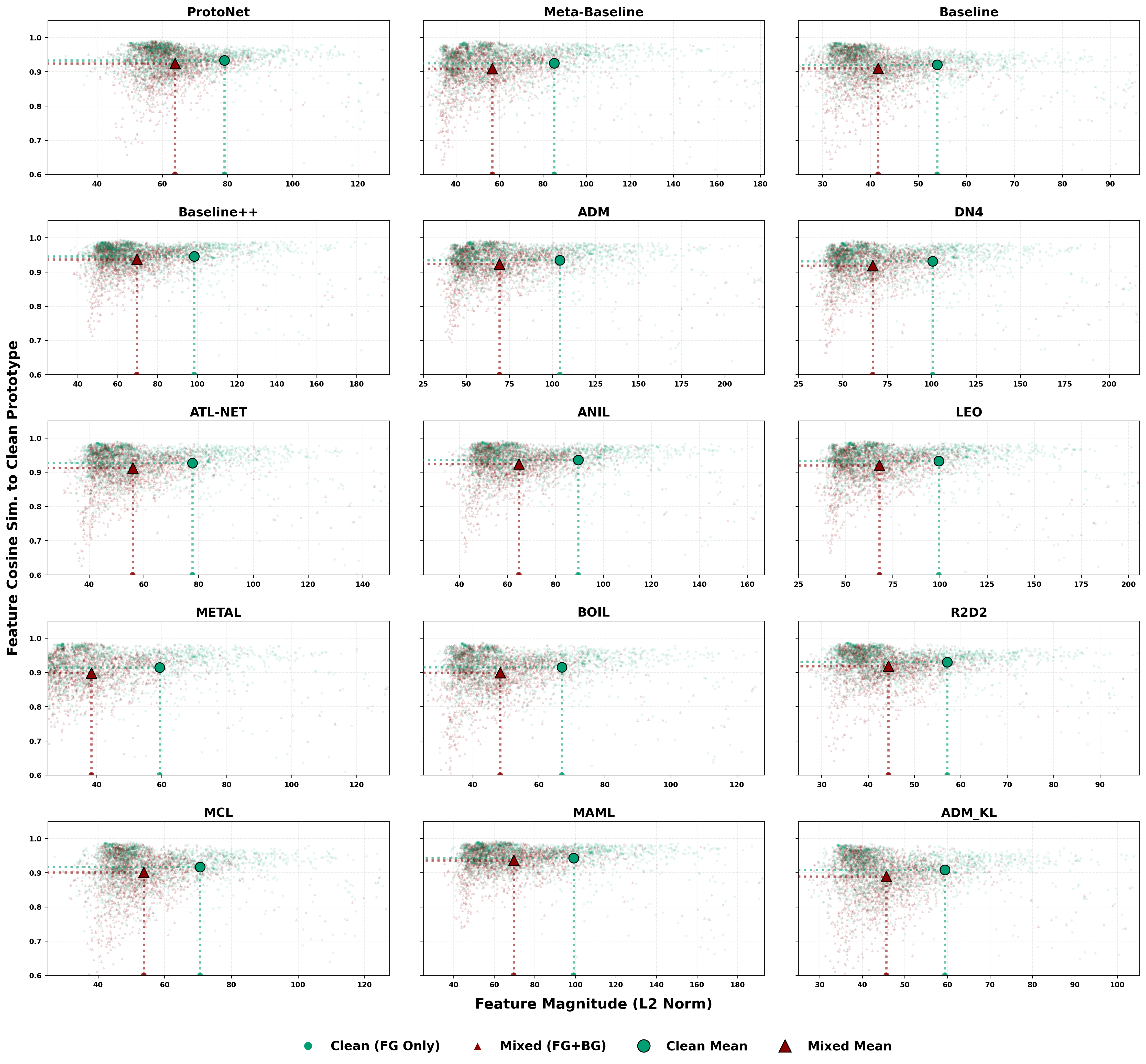}
    \caption{{Geometric Decomposition of Feature Embeddings across 15 Backbones.} 
    On the y-axis, Green circles represent clean (foreground-only) samples cosine similarity to the clean embeddings prototype and on the x-axis, the feature's magnitude ; Red triangles represent mixed (foreground + background) samples cosine similarity to the clean embeddings prototype on the y-axis and the feature magnitude on the x-axis. 
    The large circle and triangle markers denote the mean of those projections per category. 
    Across all backbones, we observe a universal phenomenon: background noise causes a {Magnitude Contraction} (shift left) while maintaining {Angular Stability} (stable y-axis). 
    This confirms that background correlations are encoded in the feature magnitude, rendering Euclidean-based methods sensitive to noise while preserving validity for Cosine-based methods.}
    \label{fig:geometry_grid}
\end{figure*}

Figure~\ref{fig:geometry_grid} visualizes the feature magnitudes (x-axis) as a function to the cosine alignment (y-axis) of the features with the clean prototype. Across all evaluated backbones—spanning metric learning (e.g., ProtoNet), optimization-based meta-learning (e.g., MAML), and standard fine-tuning (e.g., Baseline)—we observe a consistent geometric shift characterized by two distinct properties:

\begin{itemize}
    \item \textbf{Angular Stability (y-axis invariance):} The angular alignment of mixed samples remains statistically comparable to that of clean samples. Specifically, $\cos(\hat{\mathbf{z}}_{mixed}, \mathbf{p}_c) \approx \cos(\hat{\mathbf{z}}_{clean}, \mathbf{p}_c)$. This indicates that the semantic identity of the foreground object is preserved; the model does not "hallucinate" the background class, nor does the background vector rotationally displace the embedding into an orthogonal subspace.
    
    \item \textbf{Magnitude Contraction (x-axis shift):} We observe a systematic leftward shift in the feature distribution for mixed samples. As shown in the centroids of Figure \ref{fig:geometry_grid}, the mean magnitude of clean samples is consistently larger than that of mixed samples:
    \begin{equation}
        \mathbb{E}[\|\mathbf{z}_{clean}\|] > \mathbb{E}[\|\mathbf{z}_{mixed}\|]
    \end{equation}
\end{itemize}

\subsection{Implication for Few-Shot Metric Learning}
\label{sec:appendix_implication_for_fsc_metric}

This geometric disentanglement elucidates the performance disparity observed between Euclidean-based and Cosine-based few-shot algorithms:

\begin{enumerate}
    \item \textbf{Failure of Euclidean Metrics (e.g., ProtoNet):} Euclidean distance is sensitive to magnitude differences. For a query $q$ and prototype $p$:
    \begin{equation}
        \|q - p\|^2 = \underbrace{\|q\|^2 + \|p\|^2}_{\text{Magnitude Term}} - \underbrace{2\|q\|\|p\|\cos\theta}_{\text{Interaction Term}}
    \end{equation}
    The "Magnitude Contraction" observed in mixed samples ($\|q\| \downarrow$) reduces the interaction term and alters the magnitude term, creating a large Euclidean distance even if the angle $\theta$ is perfect. The model interprets the drop in signal energy as dissimilarity, leading to misclassification.

    \item \textbf{Robustness of Cosine Metrics (e.g., Baseline++, Meta-Baseline):} Cosine-based heads explicitly normalize feature vectors during inference:
    \begin{equation}
        \text{Score}(q, p) = \frac{q^T p}{\|q\|\|p\|} = \cos\theta
    \end{equation}
    By projecting all embeddings onto the unit hypersphere, these algorithms mathematically nullify the magnitude axis. Since the background information is sequestered primarily in the magnitude (as shown by our analysis), Cosine classifiers are inherently less variant to this specific type of distribution shift.
\end{enumerate}

We conclude that spurious background correlations in audio classification are not learned as semantic features (which would alter direction) but are encoded as signal dampeners (which alter magnitude). Consequently, robustness to background noise in few-shot learning is strictly dependent on the choice of a metric that ignores feature magnitude.

% ----------------------------------------------------------------------
% FIGURE PLACEHOLDER
% ----------------------------------------------------------------------

\subsection{Rigorous Quantitative Validation}

For each of the 15 few-shot algorithms where the backbone model is Conv64F, we ran Mann-Whitney U tests and computed $95\%$ confidence intervals. 

\paragraph{Magnitude Contraction (Table \ref{tab:magnitude}).} is confirmed with statistical significance across all 15 FS-Algs. The table reports the mean feature magnitude ($L_2$ norm) for clean and mixed samples with $95\%$ confidence intervals, along with the Mann-Whitney U test p-value, testing whether the two magnitude distributions are significantly different:

\begin{table}[!htb]
\centering
\caption{Few-shot algorithm performance on clean vs.\ mixed magnitude conditions with statistical significance.}
\begin{tabular}{l|cc|c}
\hline
\textbf{FS Algorithm} & \textbf{Clean Mag $\pm$ CI} & \textbf{Mixed Mag $\pm$ CI} & \textbf{p-value} \\
\hline
ProtoNet~\cite{snell2017prototypical} (2017)       & $83.26 \pm 2.57$ & $58.69 \pm 1.09$ & $9.59 \times 10^{-47}$ \\
Meta-Baseline~\cite{meta-baseline} (2021)  & $89.61 \pm 2.77$ & $63.34 \pm 1.13$ & $1.99 \times 10^{-48}$ \\
Baseline~\cite{baseline} (2019)        & $81.06 \pm 2.23$ & $60.46 \pm 0.92$ & $3.20 \times 10^{-39}$ \\
Baseline++~\cite{baseline}  (2019)     & $63.02 \pm 1.58$ & $48.40 \pm 0.68$ & $1.27 \times 10^{-37}$ \\
ADM~\cite{adm} (2020)            & $65.19 \pm 1.78$ & $48.25 \pm 0.76$ & $1.02 \times 10^{-42}$ \\
DN4~\cite{dn4} (2019)            & $79.62 \pm 2.89$ & $50.99 \pm 1.28$ & $3.28 \times 10^{-53}$ \\
ATL-NET~\cite{atlnet} (2020)        & $83.67 \pm 2.72$ & $57.45 \pm 1.14$ & $4.59 \times 10^{-48}$ \\
ANIL~\cite{anil} (2020)           & $81.16 \pm 2.67$ & $55.01 \pm 1.17$ & $1.08 \times 10^{-49}$ \\
LEO~\cite{rusu2018meta} (2019)            & $89.94 \pm 3.10$ & $59.68 \pm 1.30$ & $1.40 \times 10^{-52}$ \\
METAL~\cite{baik2021metal} (2021)          & $90.58 \pm 2.01$ & $73.07 \pm 0.80$ & $6.24 \times 10^{-36}$ \\
BOIL~\cite{oh2020boil} (2020)            & $75.74 \pm 2.24$ & $55.21 \pm 0.90$ & $2.27 \times 10^{-39}$ \\
R2D2~\cite{r2d2} (2019)           & $91.65 \pm 2.95$ & $63.30 \pm 1.26$ & $3.75 \times 10^{-47}$ \\
MCL~\cite{mcl} (2022)            & $78.69 \pm 2.72$ & $52.17 \pm 1.16$ & $7.09 \times 10^{-50}$ \\
MAML~\cite{maml} (2017)           & $80.34 \pm 2.44$ & $57.50 \pm 1.02$ & $2.81 \times 10^{-43}$ \\
ADM KL~\cite{adm} (2020)        & $89.48 \pm 2.69$ & $63.23 \pm 1.11$ & $2.28 \times 10^{-49}$ \\
\hline
\end{tabular}
\label{tab:magnitude}
\end{table}

Across all few-shot algorithms, clean samples consistently show larger magnitudes than mixed samples, with the difference being highly significant (all p-values $< 10^{-35}$).

\paragraph{Angular Stability (Table \ref{tab:cosine}).} is also confirmed. The table reports the mean cosine similarity to the clean class prototype for clean and mixed samples, the absolute difference between them (Diff), and the Mann-Whitney U test p-value. While differences are statistically detectable, the Diff column shows that the absolute effect size is negligibly small ($< 0.025$ across all backbones), confirming that background noise does not meaningfully alter the semantic direction of feature embeddings:

\begin{table}[!htb]
\centering
\caption{Cosine similarity of few-shot algorithm embeddings under clean vs.\ mixed conditions with difference and statistical significance.}
\begin{tabular}{l|cc|c|c}
\hline
\textbf{Backbone} & \textbf{Clean CosSim} & \textbf{Mixed CosSim} & \textbf{Diff} & \textbf{p-value} \\
\hline
ProtoNet~\cite{snell2017prototypical} (2017)       & $0.938$ & $0.928$ & $0.009$ & $2.53 \times 10^{-22}$ \\
Meta-Baseline~\cite{meta-baseline} (2021)  & $0.920$ & $0.900$ & $0.020$ & $8.52 \times 10^{-29}$ \\
Baseline~\cite{baseline} (2019)       & $0.927$ & $0.914$ & $0.013$ & $6.87 \times 10^{-20}$ \\
Baseline++~\cite{baseline}  (2019)     & $0.934$ & $0.924$ & $0.010$ & $2.91 \times 10^{-18}$ \\
ADM~\cite{adm} (2020)            & $0.937$ & $0.930$ & $0.007$ & $1.15 \times 10^{-13}$ \\
DN4~\cite{dn4} (2019)            & $0.905$ & $0.880$ & $0.025$ & $2.08 \times 10^{-36}$ \\
ATL-Net~\cite{atlnet} (2020)       & $0.913$ & $0.895$ & $0.017$ & $7.74 \times 10^{-25}$ \\
ANIL~\cite{anil} (2020)           & $0.938$ & $0.927$ & $0.011$ & $1.46 \times 10^{-26}$ \\
LEO~\cite{rusu2018meta} (2019)            & $0.937$ & $0.925$ & $0.012$ & $1.63 \times 10^{-25}$ \\
METAL~\cite{baik2021metal} (2021)          & $0.916$ & $0.903$ & $0.013$ & $1.97 \times 10^{-11}$ \\
BOIL~\cite{oh2020boil} (2020)           & $0.927$ & $0.912$ & $0.015$ & $3.14 \times 10^{-21}$ \\
R2D2~\cite{r2d2} (2019)           & $0.926$ & $0.912$ & $0.014$ & $5.80 \times 10^{-26}$ \\
MCL~\cite{mcl} (2022)            & $0.910$ & $0.885$ & $0.024$ & $3.89 \times 10^{-34}$ \\
MAML~\cite{maml} (2017)           & $0.932$ & $0.919$ & $0.013$ & $8.43 \times 10^{-28}$ \\
ADM KL~\cite{adm} (2020)        & $0.923$ & $0.910$ & $0.013$ & $4.50 \times 10^{-19}$ \\
\hline
\end{tabular}

\label{tab:cosine}
\end{table}

Taken together, background noise induces a statistically significant magnitude contraction (p $< 10^{-35}$ across all backbones) while feature direction remains largely intact (cosine similarity drop $< 0.025$ in all cases).

\FloatBarrier
\section{Where in the Network Are Spurious Correlations Encoded?}
\label{sec:where_is_spurious_correlation_encoded}

To investigate where in the network spurious correlations are encoded, we performed a stage-wise analysis by hooking into the intermediate stages of both Conv64F and ResNet12, evaluating IID-OOD performance using 4 inference heads, specifically, Proto, Baseline, Baseline++, and DN4.

An important note on interpreting the 1-shot results: as shown in Figure 15 of the paper, higher shot counts amplify shortcut reliance, meaning the IID-OOD gap is naturally larger in the 5-shot setting. The 1-shot gaps at intermediate stages are therefore expected to be smaller, and should be interpreted in this context rather than as evidence against the effect.

\begin{table}[ht]
\centering
\caption{IID and OOD accuracy with generalization gap across convolution stages and few-shot heads for 5-shot and 1-shot settings.}
\resizebox{\textwidth}{!}{%
\begin{tabular}{c|l|ccc|ccc}

\hline
\multirow{2}{*}{\textbf{Conv. Stage}} & \multirow{2}{*}{\textbf{Head}} 
  & \multicolumn{3}{c|}{\textbf{5-shot}} 
  & \multicolumn{3}{c}{\textbf{1-shot}} \\
 & & IID & OOD & $\Delta$ & IID & OOD & $\Delta$ \\
\hline
\multirow{4}{*}{1}
  & Proto~\cite{snell2017prototypical} (2017)      & $47.73\%$ & $41.24\%$ & $6.49\%$ & $49.19\%$ & $42.52\%$ & $6.67\%$ \\
  & Baseline~\cite{baseline} (2019)   & $51.23\%$ & $46.37\%$ & $4.86\%$ & $52.13\%$ & $46.78\%$ & $5.35\%$ \\
  & Baseline++~\cite{baseline}  (2019) & $52.95\%$ & $45.30\%$ & $7.65\%$ & $51.48\%$ & $43.87\%$ & $7.61\%$ \\
  & DN4~\cite{dn4} (2019)        & $53.00\%$ & $46.67\%$ & $6.33\%$ & $51.22\%$ & $45.43\%$ & $5.79\%$ \\
\hline
\multirow{4}{*}{2}
  & Proto~\cite{snell2017prototypical} (2017)      & $52.59\%$ & $46.05\%$ & $6.54\%$  & $55.05\%$ & $48.97\%$ & $6.08\%$ \\
  & Baseline~\cite{baseline} (2019)   & $60.57\%$ & $50.55\%$ & $10.01\%$ & $58.54\%$ & $50.85\%$ & $7.68\%$ \\
  & Baseline++~\cite{baseline}  (2019) & $58.88\%$ & $51.83\%$ & $7.05\%$  & $57.25\%$ & $50.38\%$ & $6.87\%$ \\
  & DN4~\cite{dn4} (2019)        & $59.23\%$ & $51.45\%$ & $7.78\%$  & $57.30\%$ & $50.49\%$ & $6.81\%$ \\
\hline
\multirow{4}{*}{3}
  & Proto~\cite{snell2017prototypical} (2017)      & $57.74\%$ & $52.27\%$ & $5.47\%$  & $56.20\%$ & $50.03\%$ & $6.17\%$ \\
  & Baseline~\cite{baseline} (2019)   & $59.63\%$ & $49.21\%$ & $10.43\%$ & $60.10\%$ & $51.63\%$ & $8.47\%$ \\
  & Baseline++~\cite{baseline}  (2019) & $58.01\%$ & $51.20\%$ & $6.81\%$  & $58.26\%$ & $51.22\%$ & $7.04\%$ \\
  & DN4~\cite{dn4} (2019)        & $59.51\%$ & $51.18\%$ & $8.33\%$  & $57.96\%$ & $50.92\%$ & $7.04\%$ \\
\hline
\end{tabular}%
}
\label{tab:conv_stage_results}
\end{table}

For Conv64F, the IID-OOD gap is consistently strong across both 1-shot and 5-shot settings at every layer, ranging from $5$--$10\%$. The gap is already clearly present at Layer 1, confirming that spurious background correlations are encoded from the very first convolutional stage.

\begin{table}[ht]
\centering
\caption{IID and OOD accuracy with generalization gap across ResNet12 residual blocks and few-shot heads for 5-shot and 1-shot settings.}
\resizebox{\textwidth}{!}{%
\begin{tabular}{c|l|ccc|ccc}
\hline
\multirow{2}{*}{\textbf{Residual Block}} & \multirow{2}{*}{\textbf{Head}} 
  & \multicolumn{3}{c|}{\textbf{5-shot}} 
  & \multicolumn{3}{c}{\textbf{1-shot}} \\
 & & IID & OOD & $\Delta$ & IID & OOD & $\Delta$ \\
\hline
\multirow{4}{*}{1}
  & Proto~\cite{snell2017prototypical} (2017)      & $47.23\%$ & $40.62\%$ & $6.61\%$ & $39.57\%$ & $37.96\%$ & $1.61\%$ \\
  & Baseline~\cite{baseline} (2019)   & $53.48\%$ & $48.09\%$ & $5.39\%$ & $39.19\%$ & $39.11\%$ & $0.07\%$ \\
  & Baseline++~\cite{baseline}  (2019) & $51.93\%$ & $43.75\%$ & $8.17\%$ & $42.68\%$ & $39.89\%$ & $2.79\%$ \\
  & DN4~\cite{dn4} (2019)        & $51.77\%$ & $45.93\%$ & $5.84\%$ & $41.07\%$ & $39.55\%$ & $1.53\%$ \\
\hline
\multirow{4}{*}{2}
  & Proto~\cite{snell2017prototypical} (2017)      & $53.55\%$ & $47.01\%$ & $6.53\%$ & $43.53\%$ & $42.85\%$ & $0.68\%$ \\
  & Baseline~\cite{baseline} (2019)   & $59.08\%$ & $52.75\%$ & $6.33\%$ & $43.73\%$ & $42.39\%$ & $1.35\%$ \\
  & Baseline++~\cite{baseline}  (2019) & $59.17\%$ & $52.97\%$ & $6.20\%$ & $46.94\%$ & $45.49\%$ & $1.45\%$ \\
  & DN4~\cite{dn4} (2019)        & $58.31\%$ & $51.31\%$ & $7.00\%$ & $44.79\%$ & $43.39\%$ & $1.39\%$ \\
\hline
\multirow{4}{*}{3}
  & Proto~\cite{snell2017prototypical} (2017)      & $58.99\%$ & $53.21\%$ & $5.78\%$ & $47.77\%$ & $45.72\%$ & $2.05\%$ \\
  & Baseline~\cite{baseline} (2019)   & $62.47\%$ & $56.26\%$ & $6.21\%$ & $46.02\%$ & $45.47\%$ & $0.55\%$ \\
  & Baseline++~\cite{baseline}  (2019) & $63.05\%$ & $55.35\%$ & $7.69\%$ & $49.06\%$ & $47.57\%$ & $1.49\%$ \\
  & DN4~\cite{dn4} (2019)        & $62.28\%$ & $54.30\%$ & $7.98\%$ & $49.12\%$ & $45.94\%$ & $3.18\%$ \\
\hline
\end{tabular}%
}
\label{tab:resnet12_results}
\end{table}

For ResNet12, the 5-shot gaps are consistently present from residual block 1 ($5$--$8\%$) and grow in deeper layers. The 1-shot gaps at intermediate layers are smaller ($0.07\%$--$3.18\%$), consistent with the shot-count amplification effect noted above. Notably, the intermediate layer gaps are smaller than the final embedding gap ($\approx 10\%$), which is expected: intermediate representations are less task-discriminative overall, and the spurious correlation effect amplifies progressively as representations become more semantically structured through the network.

Taken together, across both architectures, both shot settings, and all inference heads, the IID-OOD gap is present from the earliest feature extraction stage. This confirms that spurious background correlations are encoded throughout the feature hierarchy rather than emerging only at the final embedding layer.

\FloatBarrier
\section{On the Distribution of SpurAudio}
\label{sec:appendix_distribution_spuraudio}
In this section we conducted a comprehensive distributional analysis against the FSD50K benchmark~\cite{fonseca2021fsd50k}. As FSD50K is widely regarded as a gold standard for ``in-the-wild'' multi-label audio event classification, aligning with its semantic topology is critical for establishing the utility of our dataset. We utilized the Contrastive Language-Audio Pretraining (CLAP) model~\cite{elizalde2023clap} to extract rich, semantically aware embeddings from both datasets, enabling a direct comparison in a shared latent space.

We employed two complementary metrics to evaluate the semantic and structural alignment of our data: Maximum Mean Discrepancy (MMD) and centroid cosine similarity.

\subsection{Methodology and Metrics}
\label{sec:appendix_methodology_and_metrics}
\begin{itemize}
    \item \textbf{Maximum Mean Discrepancy (MMD):} This metric measures the distance between the feature distributions of our synthesized data and the real-world FSD50K data using a kernel-based approach (RBF kernel). Lower values indicate closer distributional matching. MMD is particularly effective at detecting non-linear discrepancies between the manifolds of the two domains.
    
    \item \textbf{Centroid Cosine Similarity:} This metric quantifies the semantic alignment of class prototypes. We compute the cosine similarity between the centroid of a class in \textit{SpurAudio} and the centroid of the corresponding class in FSD50K. High values (approaching 1.0) indicate that the ``average'' representation of a class (e.g., ``Cat'') in our dataset is semantically identical to its real-world counterpart.
    
\end{itemize}

\subsection{Analysis of Results}
\label{sec:appendix_analysis_distribution}
\begin{figure}[h]
    \centering
    \includegraphics[width=\linewidth]{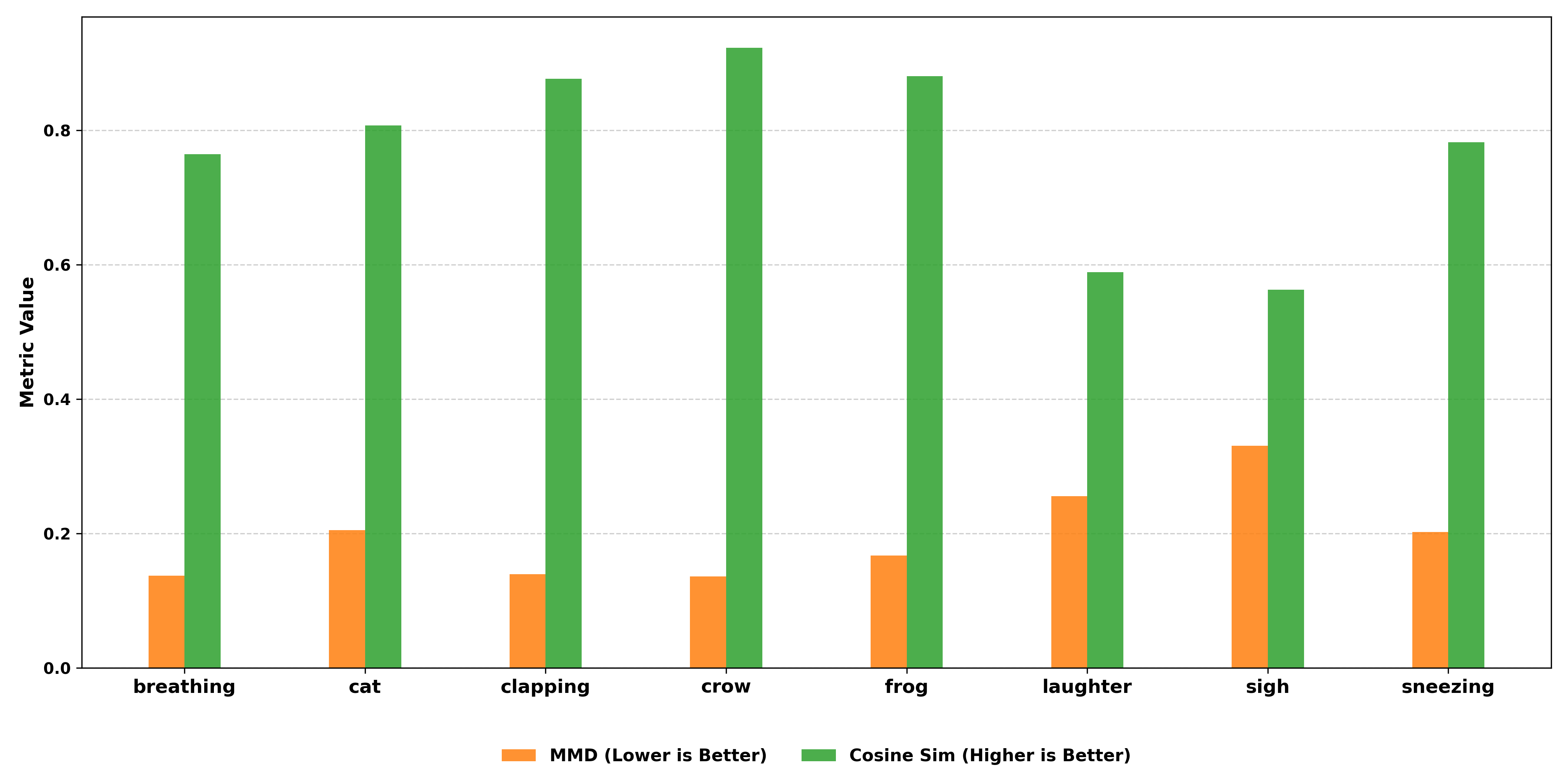}
    \caption{{Distributional Analysis Metrics per Class.} Comparison of \textit{SpurAudio} against FSD50K using CLAP embeddings. Green bars denote Centroid Cosine Similarity (higher is better), indicating strong semantic alignment for acoustic events. Orange bars represent MMD (lower is better).}
    \label{fig:distribution_analysis}
\end{figure}

Figure \ref{fig:distribution_analysis} presents the per-class metrics alongside the global structural alignment score.

\subsubsection{Semantic Alignment}
\label{sec:appendix_semantic_alignment}
The Centroid Cosine Similarity results reveal a distinct ``Event vs. Paralinguistic'' split. Distinct acoustic events such as \textbf{Crow} (Cosine: 0.92) and \textbf{Frog} (Cosine: 0.88) exhibit near-perfect semantic alignment with FSD50K. This confirms that our synthesis pipeline correctly captures the core spectral and temporal characteristics of these classes, positioning them identically in the CLAP latent space. Conversely, highly variable human paralinguistic classes like \textit{Laughter} and \textit{Sigh} show lower cosine similarity ($\approx 0.56$). This is an expected consequence of the ``in-the-wild'' variance; FSD50K contains a vast diversity of laughter types (e.g., giggles, guffaws) mixed with background noise, whereas our source data represents a more canonical subset.

\subsubsection{Distributional and Structural Integrity}
\label{sec:appendix_distritional_structural_integrity}
The MMD scores (Avg: 0.19) quantify the distributional gap between our synthesized foregrounds and the FSD50K recordings. While these non-zero values indicate a detectable domain shift-attributable to the cleaner, high-SNR nature of \textit{SpurAudio} compared to the noisy FSD50K environments; the structural metrics confirm this shift is uniform and non-destructive.

\section{More Results On Large-Audio Models And Transformers}
\label{sec:appendix_more_results_audio_models}
In this section we showcase the remaining 1-shot results on the large foundation models CLAP~\cite{laionclap2023}, AST~\cite{gong2021ast}, AudioMAE-AS20K~\cite{huang2022masked},Qwen2-Audio-7B~\cite{Qwen2-Audio} for 1-shot (Table ~\ref{tab:large_1shot}) as well as other variants HTSAT-CLAP~\cite{laionclap2023}, self-supervised (BEATS~\cite{chen2022beats}) and PaSST~\cite{koutini2021efficient} at Table ~\ref{tab:results_new_encoders_1_5_shot} on 1-shot and 5-shot. We also showcase the performance on a pre-trained transformer based encoder Wav2Vec 2.0~\cite{schneider2019wav2vec,baevski2020wav2vec} in Table~\ref{tab:results_wav2vec2_1_5_shot}.

\begin{table}[!ht]
\centering
\caption{IID and OOD accuracy (\%) across few-shot methods (rows) on pretrained large audio models for $1$-shot classification.}
\resizebox{\textwidth}{!}{%
\setlength{\tabcolsep}{2pt}
\begin{tabular}{l|ccc|ccc|ccc|ccc}
\hline
\multirow{2}{*}{\textbf{Method}} 
  & \multicolumn{3}{c|}{\textbf{CLAP}} 
  & \multicolumn{3}{c|}{\textbf{AudioMAE-AS20K}} 
  & \multicolumn{3}{c|}{\textbf{AST}} 
  & \multicolumn{3}{c}{\textbf{Qwen2-audio-7b}} \\
 & IID & OOD & $\Delta$ & IID & OOD & $\Delta$ & IID & OOD & $\Delta$ & IID & OOD & $\Delta$ \\
\hline
\textbf{Proto}~\cite{snell2017prototypical} & $88.99 \pm 0.48$ & $85.16 \pm 0.62$ & 3.83 & $79.25 \pm 0.49$ & $68.40 \pm 0.61$ & 10.85 & $88.08 \pm 0.46$ & $83.73 \pm 0.59$ & 4.35 & $81.68 \pm 0.57$ & $75.34 \pm 0.65$ & 6.34 \\
\textbf{Baseline}~\cite{baseline} & $88.93 \pm 0.49$ & $85.42 \pm 0.60$ & 3.51 & $79.43 \pm 0.51$ & $70.89 \pm 0.62$ & 8.54 & $87.85 \pm 0.48$ & $84.56 \pm 0.58$ & 3.29 & $83.06 \pm 0.59$ & $76.94 \pm 0.66$ & 6.12 \\
\textbf{Baseline++}~\cite{baseline} & $89.15 \pm 0.48$ & $85.06 \pm 0.60$ & 4.09 & $79.25 \pm 0.50$ & $69.21 \pm 0.61$ & 10.04 & $87.95 \pm 0.49$ & $84.35 \pm 0.59$ & 3.60 & $82.77 \pm 0.58$ & $77.16 \pm 0.65$ & 5.61 \\
\textbf{DN4}~\cite{dn4} & $89.69 \pm 0.47$ & $85.64 \pm 0.59$ & 4.05 & $79.41 \pm 0.52$ & $66.46 \pm 0.63$ & 12.95 & $88.01 \pm 0.47$ & $84.51 \pm 0.58$ & 3.50 & $82.06 \pm 0.60$ & $75.96 \pm 0.67$ & 6.10 \\
\textbf{Proto-LP}~\cite{protolp} & $\mathbf{95.41 \pm 0.50}$ & $\mathbf{94.28 \pm 0.62}$ & $\mathbf{1.13}$ & $\mathbf{95.79 \pm 0.45}$ & $\mathbf{93.70 \pm 0.60}$ & 2.09 & $\mathbf{95.93 \pm 0.45}$ & $\mathbf{93.53 \pm 0.69}$ & 2.40 & $\mathbf{96.96 \pm 0.39}$ & $\mathbf{96.73 \pm 0.39}$ & $\mathbf{0.23}$ \\
\textbf{BPA}~\cite{shalam2024balanced} & $90.05 \pm 0.46$ & $84.55 \pm 0.62$ & 5.50 & $86.40 \pm 0.48$ & $80.70 \pm 0.60$ & 5.70 & $88.85 \pm 0.45$ & $83.10 \pm 0.59$ & 5.75 & $82.75 \pm 0.55$ & $75.85 \pm 0.65$ & 6.90 \\
\textbf{Hela-VFA}~\cite{lee2024hela} & $87.85 \pm 0.52$ & $79.85 \pm 0.65$ & 8.00 & $84.20 \pm 0.53$ & $74.85 \pm 0.63$ & 9.35 & $85.29 \pm 0.53$ & $80.62 \pm 0.59$ & 4.67 & $80.95 \pm 0.60$ & $70.85 \pm 0.66$ & 10.10 \\
\textbf{Laplacian}~\cite{laplacianshot} & $90.43 \pm 0.49$ & $87.45 \pm 0.68$ & 2.98 & $86.91 \pm 0.53$ & $80.60 \pm 0.68$ & 6.31 & $89.30 \pm 0.49$ & $85.22 \pm 0.66$ & 4.08 & $83.10 \pm 0.59$ & $77.06 \pm 0.68$ & 6.04 \\
\textbf{BD-CSPN}~\cite{bdcspn} & $93.38 \pm 0.43$ & $91.70 \pm 0.61$ & 1.68 & $92.55 \pm 0.44$ & $90.65 \pm 0.56$ & $\mathbf{1.90}$ & $93.24 \pm 0.43$ & $90.99 \pm 0.60$ & $\mathbf{2.25}$ & $90.79 \pm 0.45$ & $86.25 \pm 0.67$ & 4.54 \\
\textbf{PADDLE}~\cite{paddle} & $91.35 \pm 0.46$ & $88.65 \pm 0.56$ & 2.70 & $89.65 \pm 0.46$ & $86.97 \pm 0.54$ & 2.68 & $90.47 \pm 0.46$ & $87.85 \pm 0.55$ & 2.62 & $87.11 \pm 0.47$ & $82.55 \pm 0.59$ & 4.56 \\
\textbf{ECPE}~\cite{ecpe} & $93.45 \pm 0.44$ & $91.40 \pm 0.56$ & 2.05 & $92.11 \pm 0.43$ & $89.53 \pm 0.56$ & 2.58 & $92.07 \pm 0.45$ & $89.05 \pm 0.61$ & 3.02 & $89.79 \pm 0.44$ & $85.46 \pm 0.59$ & 4.33 \\
\hline
\end{tabular}%
}
\label{tab:large_1shot}
\end{table}

\begin{table}[!htbp]
\centering
\caption{1-shot, 5-shot IID and OOD accuracy (\%) on three additional pretrained
encoders: PaSST (AudioSet-pretrained), HTSAT-CLAP (HTSAT encoder fused with CLAP), and BEATS.}
\resizebox{\textwidth}{!}{%
\begin{tabular}{l|ccc|ccc|ccc|ccc|ccc|ccc}
\hline
\multirow{3}{*}{\textbf{Method}}
  & \multicolumn{6}{c|}{\textbf{PaSST}}
  & \multicolumn{6}{c|}{\textbf{HTSAT-CLAP}}
  & \multicolumn{6}{c}{\textbf{BEATS}} \\
 & \multicolumn{3}{c|}{\textbf{1-shot}} & \multicolumn{3}{c|}{\textbf{5-shot}}
 & \multicolumn{3}{c|}{\textbf{1-shot}} & \multicolumn{3}{c|}{\textbf{5-shot}}
 & \multicolumn{3}{c|}{\textbf{1-shot}} & \multicolumn{3}{c}{\textbf{5-shot}} \\
 & IID & OOD & $\Delta$ & IID & OOD & $\Delta$
 & IID & OOD & $\Delta$ & IID & OOD & $\Delta$
 & IID & OOD & $\Delta$ & IID & OOD & $\Delta$ \\
\hline
\textbf{Proto}~\cite{snell2017prototypical} (2017)
& $48.05 \pm 0.68$ & $46.34 \pm 0.75$ & $\mathbf{1.71}$
& $61.07 \pm 0.62$ & $55.76 \pm 0.73$ & 5.31
& $83.35 \pm 0.57$ & $77.79 \pm 0.64$ & 5.56
& $93.36 \pm 0.33$ & $84.67 \pm 0.58$ & 8.69
& $72.53 \pm 0.67$ & $66.00 \pm 0.68$ & 6.53
& $89.28 \pm 0.39$ & $76.95 \pm 0.62$ & 12.33 \\
\textbf{Baseline}~\cite{baseline} (2019)
& $49.50 \pm 0.70$ & $46.93 \pm 0.75$ & 2.57
& $58.15 \pm 0.61$ & $54.57 \pm 0.68$ & $\mathbf{3.58}$
& $83.92 \pm 0.54$ & $78.06 \pm 0.64$ & 5.86
& $93.09 \pm 0.35$ & $84.70 \pm 0.57$ & 8.39
& $74.47 \pm 0.68$ & $68.27 \pm 0.70$ & 6.20
& $88.93 \pm 0.40$ & $77.39 \pm 0.61$ & 11.54 \\
\textbf{Baseline++}~\cite{baseline} (2019)
& $50.34 \pm 0.72$ & $47.74 \pm 0.77$ & 2.60
& $\mathbf{63.74 \pm 0.59}$ & $\mathbf{58.02 \pm 0.68}$ & 5.72
& $83.93 \pm 0.58$ & $78.57 \pm 0.67$ & 5.36
& $93.58 \pm 0.32$ & $84.27 \pm 0.56$ & 9.31
& $74.46 \pm 0.67$ & $68.31 \pm 0.70$ & 6.15
& $90.19 \pm 0.37$ & $77.52 \pm 0.61$ & 12.67 \\
\textbf{DN4}~\cite{dn4} (2019)
& $48.44 \pm 0.66$ & $45.72 \pm 0.76$ & 2.72
& $60.96 \pm 0.61$ & $54.62 \pm 0.72$ & 6.34
& $84.58 \pm 0.54$ & $77.39 \pm 0.65$ & 7.19
& $92.81 \pm 0.33$ & $84.05 \pm 0.60$ & 8.76
& $73.59 \pm 0.69$ & $66.17 \pm 0.74$ & 7.42
& $88.65 \pm 0.40$ & $67.17 \pm 0.65$ & 21.48 \\
\textbf{Proto-LP}~\cite{protolp} (2023)
& $49.62 \pm 0.92$ & $47.47 \pm 0.97$ & 2.15
& $61.37 \pm 0.67$ & $55.65 \pm 0.82$ & 5.72
& $\mathbf{93.19 \pm 0.64}$ & $\mathbf{91.31 \pm 0.63}$ & $\mathbf{1.88}$
& $\mathbf{95.66 \pm 0.34}$ & $\mathbf{92.21 \pm 0.52}$ & $\mathbf{3.45}$
& $\mathbf{87.12 \pm 0.83}$ & $\mathbf{83.12 \pm 0.96}$ & $\mathbf{4.00}$
& $\mathbf{92.44 \pm 0.45}$ & $\mathbf{85.68 \pm 0.81}$ & $\mathbf{6.76}$ \\
\textbf{BPA}~\cite{shalam2024balanced} (2024)
& $47.50 \pm 0.72$ & $44.50 \pm 0.78$ & 3.00
& $59.83 \pm 0.59$ & $46.07 \pm 0.52$ & 13.76
& $84.50 \pm 0.55$ & $76.00 \pm 0.66$ & 8.50
& $94.00 \pm 0.32$ & $83.60 \pm 0.59$ & 10.40
& $73.95 \pm 0.65$ & $65.90 \pm 0.68$ & 8.05
& $90.71 \pm 0.35$ & $75.84 \pm 0.56$ & 14.87 \\
\textbf{Hela-VFA}~\cite{lee2024hela} (2024)
& $48.50 \pm 0.70$ & $43.80 \pm 0.78$ & 4.70
& $61.99 \pm 0.56$ & $46.65 \pm 0.53$ & 15.34
& $82.00 \pm 0.60$ & $72.50 \pm 0.68$ & 9.50
& $92.90 \pm 0.35$ & $78.90 \pm 0.61$ & 14.00
& $70.85 \pm 0.70$ & $60.45 \pm 0.73$ & 10.40
& $87.36 \pm 0.40$ & $69.16 \pm 0.53$ & 18.20 \\
\textbf{LaplacianShot}~\cite{laplacianshot} (2020)
& $48.30 \pm 0.70$ & $46.07 \pm 0.79$ & 2.23
& $60.30 \pm 0.61$ & $55.06 \pm 0.75$ & 5.24
& $86.10 \pm 0.57$ & $80.41 \pm 0.69$ & 5.69
& $93.65 \pm 0.34$ & $85.65 \pm 0.57$ & 8.00
& $74.45 \pm 0.72$ & $67.43 \pm 0.73$ & 7.02
& $89.48 \pm 0.39$ & $77.58 \pm 0.62$ & 11.90 \\
\textbf{BD-CSPN}~\cite{bdcspn} (2020)
& $48.05 \pm 0.74$ & $46.25 \pm 0.82$ & 1.80
& $57.58 \pm 0.64$ & $52.13 \pm 0.82$ & 5.45
& $89.96 \pm 0.54$ & $85.81 \pm 0.66$ & 4.15
& $94.10 \pm 0.33$ & $88.05 \pm 0.59$ & 6.05
& $81.73 \pm 0.67$ & $76.90 \pm 0.70$ & 4.83
& $90.23 \pm 0.38$ & $80.87 \pm 0.61$ & 9.36 \\
\textbf{PADDLE}~\cite{paddle} (2022)
& $51.15 \pm 0.70$ & $48.18 \pm 0.77$ & 2.97
& $59.92 \pm 0.59$ & $55.20 \pm 0.73$ & 4.72
& $86.40 \pm 0.54$ & $81.52 \pm 0.61$ & 4.88
& $94.05 \pm 0.32$ & $86.94 \pm 0.54$ & 7.11
& $78.29 \pm 0.61$ & $73.08 \pm 0.68$ & 5.21
& $90.56 \pm 0.36$ & $79.11 \pm 0.60$ & 11.45 \\
\textbf{ECPE}~\cite{ecpe} (2026)
& $\mathbf{52.17 \pm 0.80}$ & $\mathbf{49.62 \pm 0.82}$ & 2.55
& $62.84 \pm 0.62$ & $56.52 \pm 0.76$ & 6.32
& $89.53 \pm 0.49$ & $85.74 \pm 0.62$ & 3.79
& $94.38 \pm 0.33$ & $88.66 \pm 0.54$ & 5.72
& $80.90 \pm 0.63$ & $75.69 \pm 0.71$ & 5.21
& $91.60 \pm 0.35$ & $81.85 \pm 0.65$ & 9.75 \\
\hline
\end{tabular}%
}
\label{tab:results_new_encoders_1_5_shot}
\end{table}

\begin{table}[!htbp]
\centering
\caption{1-shot and 5-shot IID and OOD accuracy (\%) on Wav2Vec 2.0}
\resizebox{\textwidth}{!}{%
\begin{tabular}{l|ccc|ccc}
\hline
\multirow{2}{*}{\textbf{Method}}
  & \multicolumn{3}{c|}{\textbf{1-shot}}
  & \multicolumn{3}{c}{\textbf{5-shot}} \\
 & IID & OOD & $\Delta$ & IID & OOD & $\Delta$ \\
\hline
\textbf{Proto}~\cite{snell2017prototypical} (2017)
& $37.55 \pm 0.55$ & $33.09 \pm 0.48$ & 4.46
& $50.79 \pm 0.52$ & $39.15 \pm 0.45$ & 11.64 \\
\textbf{Baseline}~\cite{baseline} (2019)
& $38.64 \pm 0.55$ & $34.00 \pm 0.48$ & 4.64
& $49.19 \pm 0.51$ & $38.79 \pm 0.45$ & 10.40 \\
\textbf{Baseline++}~\cite{baseline} (2019)
& $38.25 \pm 0.53$ & $33.72 \pm 0.49$ & 4.53
& $53.44 \pm 0.51$ & $40.54 \pm 0.45$ & 12.90 \\
\textbf{DN4}~\cite{dn4} (2019)
& $37.67 \pm 0.54$ & $33.33 \pm 0.47$ & 4.34
& $51.08 \pm 0.50$ & $39.81 \pm 0.44$ & 11.27 \\
\textbf{Proto-LP}~\cite{protolp} (2023)
& $37.05 \pm 0.67$ & $32.60 \pm 0.64$ & 4.45
& $47.74 \pm 0.60$ & $36.60 \pm 0.57$ & 11.14 \\
\textbf{BPA}~\cite{shalam2024balanced} (2024)
& $38.50 \pm 0.50$ & $33.95 \pm 0.45$ & 4.55
& $53.73 \pm 0.43$ & $\mathbf{43.39 \pm 0.38}$ & 10.34 \\
\textbf{Hela-VFA}~\cite{lee2024hela} (2024)
& $37.40 \pm 0.55$ & $33.05 \pm 0.48$ & 4.35
& $51.81 \pm 0.44$ & $41.91 \pm 0.36$ & $\mathbf{9.90}$ \\
\textbf{LaplacianShot}~\cite{laplacianshot} (2020)
& $37.33 \pm 0.57$ & $33.47 \pm 0.48$ & $\mathbf{3.86}$
& $49.86 \pm 0.52$ & $38.23 \pm 0.47$ & 11.63 \\
\textbf{BD-CSPN}~\cite{bdcspn} (2020)
& $37.47 \pm 0.58$ & $33.36 \pm 0.50$ & 4.11
& $48.07 \pm 0.54$ & $37.36 \pm 0.46$ & 10.71 \\
\textbf{PADDLE}~\cite{paddle} (2022)
& $39.07 \pm 0.56$ & $\mathbf{35.05 \pm 0.49}$ & 4.02
& $50.85 \pm 0.51$ & $39.99 \pm 0.48$ & 10.86 \\
\textbf{ECPE}~\cite{ecpe} (2026)
& $\mathbf{39.84 \pm 0.60}$ & $33.97 \pm 0.52$ & 5.87
& $\mathbf{54.49 \pm 0.52}$ & $41.35 \pm 0.47$ & 13.14 \\
\hline
\end{tabular}%
}
\label{tab:results_wav2vec2_1_5_shot}
\end{table}

% \begin{table}[!ht]
% \centering
% \caption{IID and OOD accuracy (\%) with generalization gap across few-shot methods and models when number of shots is set to $1$.}
% \resizebox{\textwidth}{!}{%
% \begin{tabular}{l|ccc|ccc|ccc|ccc}
% \hline
% \multirow{2}{*}{\textbf{Model}} 
%   & \multicolumn{3}{c|}{\textbf{Proto}} 
%   & \multicolumn{3}{c|}{\textbf{Baseline}} 
%   & \multicolumn{3}{c|}{\textbf{Baseline++}} 
%   & \multicolumn{3}{c}{\textbf{DN4}} \\
%  & IID & OOD & $\Delta$ & IID & OOD & $\Delta$ & IID & OOD & $\Delta$ & IID & OOD & $\Delta$ \\
% \hline
% CLAP                     & 88.99 & 85.16 & 3.83  & 88.93 & 85.42 & 3.51  & 89.15 & 85.06 & 4.09  & 89.69 & 85.64 & 4.05  \\
% AudioMAE-AS20K           & 79.25 & 68.40 & 10.85 & 79.43 & 70.89 & 8.54  & 79.25 & 69.21 & 10.04 & 79.41 & 66.46 & 12.95 \\
% AST                      & 88.08 & 83.73 & 4.35  & 87.85 & 84.56 & 3.29  & 87.95 & 84.35 & 3.60  & 88.01 & 84.51 & 3.50  \\
% Qwen2-audio-7b-instruct  & 81.68 & 75.34 & 6.34  & 83.06 & 76.94 & 6.12  & 82.77 & 77.16 & 5.61  & 82.06 & 75.96 & 6.10  \\
% BEATS                    & 72.53 & 66.00 & 6.53  & 74.47 & 68.27 & 6.20  & 74.46 & 68.31 & 6.15  & 73.59 & 66.17 & 7.42  \\
% \hline
% \end{tabular}%
% }
% \end{table}

\section{Classification Accuracy as a Function of Number Of Shots}
\label{sec:appendix_more_shots_effect}

\begin{figure*}[!htbp]
    \centering

    % Row 1: Meta-Baseline
    \begin{subfigure}{\textwidth}
        \centering
        \includegraphics[width=0.65\textwidth]{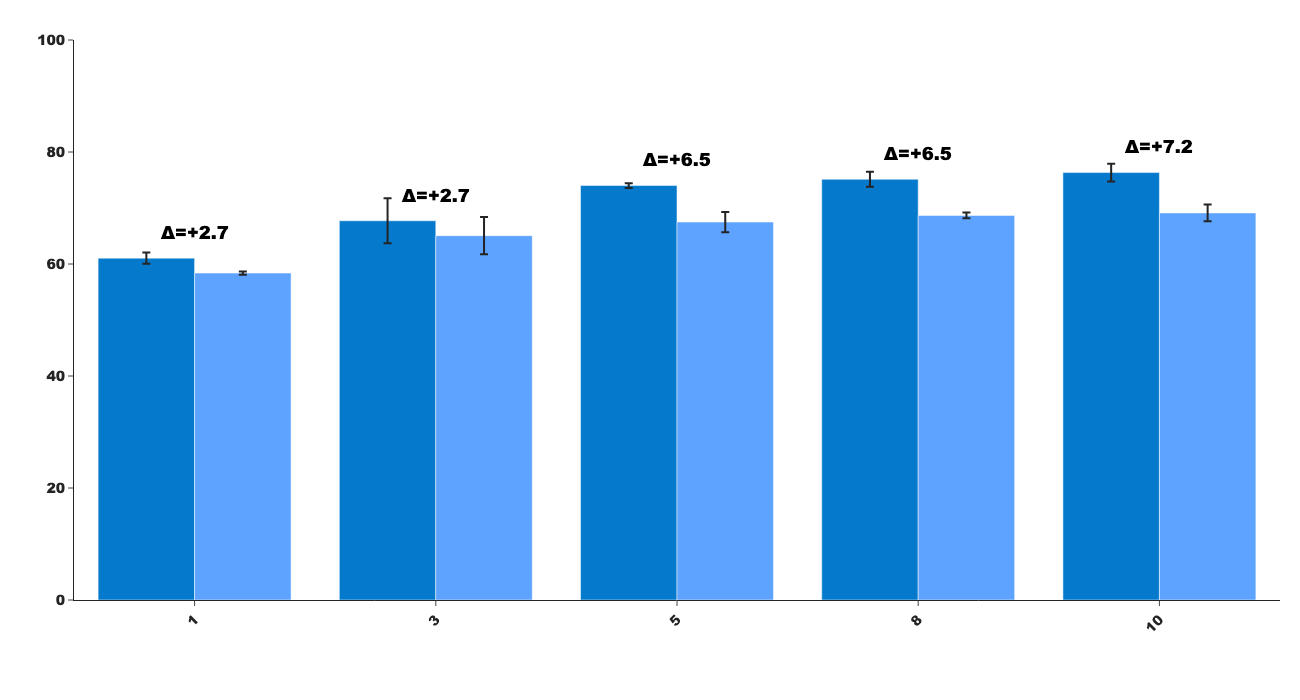}
        \caption{Meta-Baseline 1-shot 3-shot 5-shot 8-shot 10-shot \emph{IID} vs. \emph{OOD}}
        \label{fig:meta_baseline_shots_figure_iid_vs_ood}
    \end{subfigure}

    \vspace{0.6em}

    % Row 2: R2D2
    \begin{subfigure}{\textwidth}
        \centering
        \includegraphics[width=0.65\textwidth]{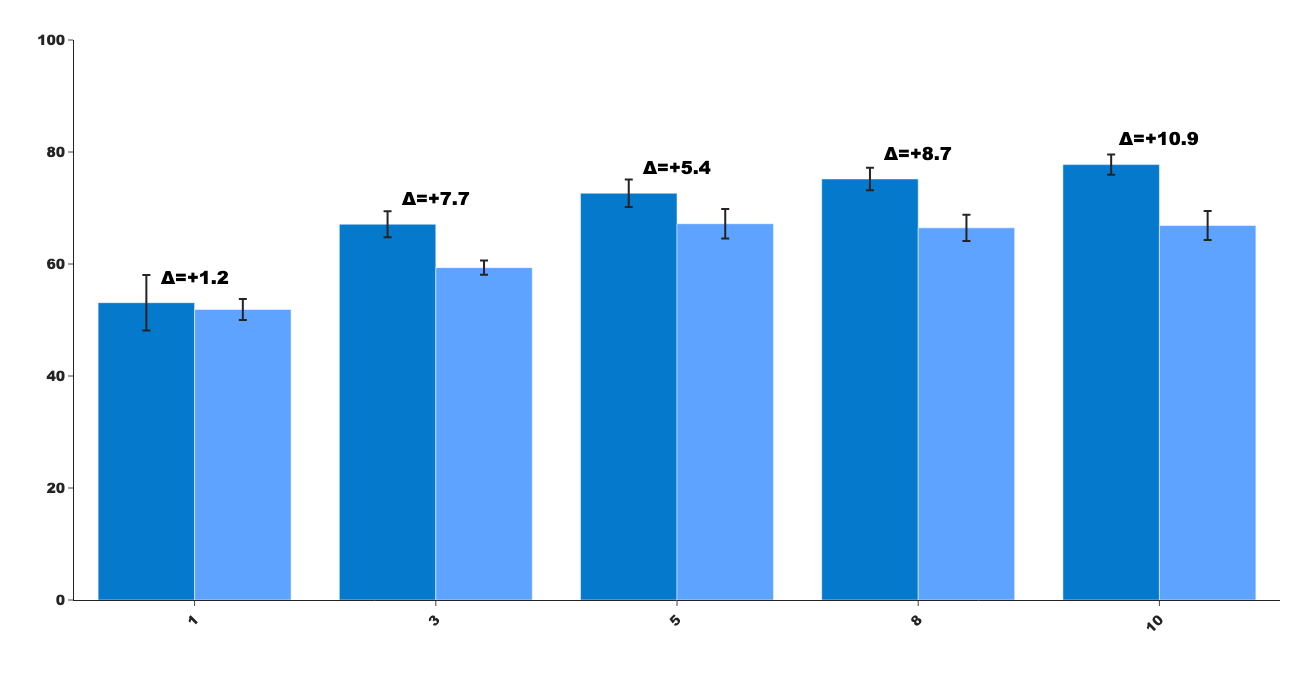}
        \caption{R2D2 1-shot 3-shot 5-shot 8-shot 10-shot \emph{IID} vs. \emph{OOD}}
        \label{fig:r2d2_shots_figure_iid_vs_ood}
    \end{subfigure}

    \vspace{0.6em}

    % Row 3: DeepBDC
    \begin{subfigure}{\textwidth}
        \centering
        \includegraphics[width=0.65\textwidth]{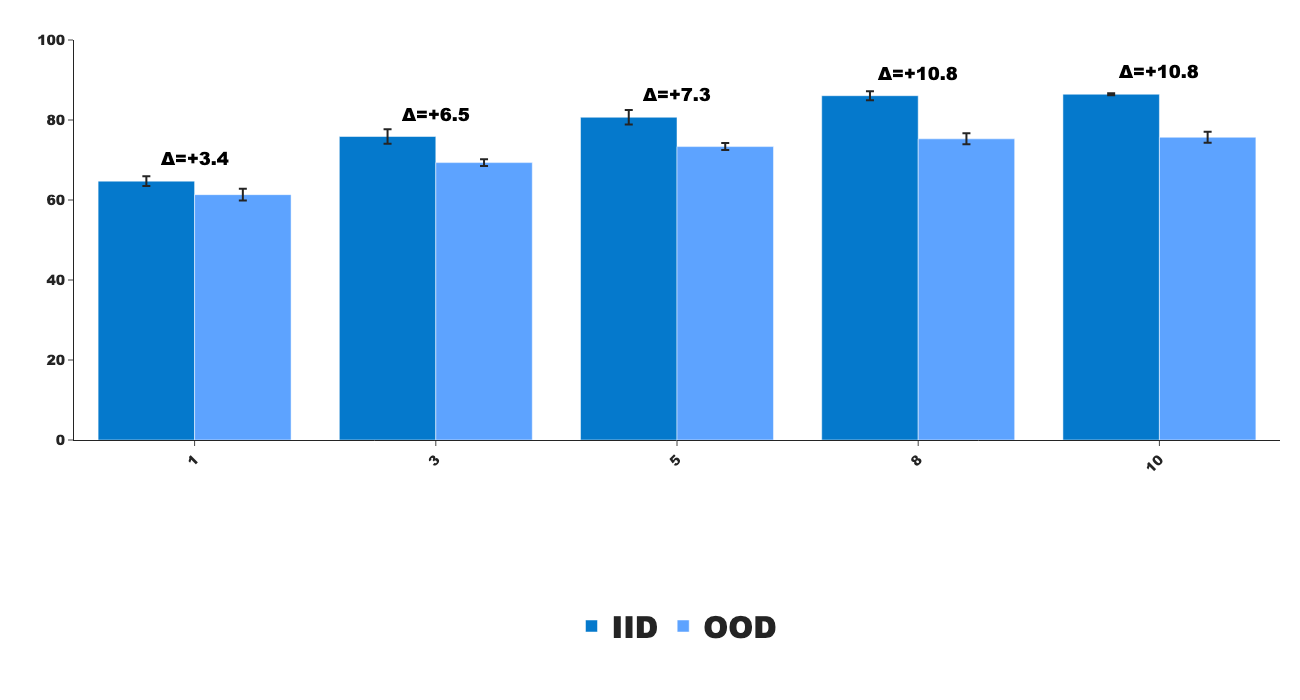}
        \caption{DeepBDC 1-shot 3-shot 5-shot 8-shot 10-shot \emph{IID} vs. \emph{OOD}}
        \label{fig:deepbdc_shots_figure_iid_vs_ood}
    \end{subfigure}

    \caption{Impact of support set size ($K$) on \emph{IID} versus \emph{OOD} performance. The widening gap illustrates a ``Simple Bias,'' where the model increasingly relies on spurious background correlations as $K$ grows. Notably, this generalization gap eventually plateaus, indicating that the model's convergence on the spurious feature saturates at higher shot counts.}
    \label{fig:more_shots_effect}
\end{figure*}

As the number of shots increases, we observe a consistent widening of the performance gap between the \emph{IID} and \emph{OOD} settings. While additional shots improve absolute accuracy in both regimes, the gains are substantially larger under \emph{IID} conditions, indicating that the models increasingly exploit dataset-specific correlations. Notably, this gap does not grow indefinitely: beyond a certain number of shots, performance in both settings saturates and the \emph{IID}--\emph{OOD} gap converges, suggesting that the learned representations reach a stable regime where additional supervision yields diminishing returns. We refer the reader to Figure~\ref{fig:more_shots_effect} for a visual illustration of this phenomenon.

\FloatBarrier
\section{Additional Contrastive Learning Experiments}
\label{sec:contrastive_experiments}

Table~\ref{tab:contrastive_benchmark} shows that adding contrastive objectives to prototypical learning does not consistently improve robustness under OOD background shifts.
Across both 1-shot and 5-shot settings, contrastive variants achieve strong IID performance but retain substantial IID--OOD gaps, with only limited and inconsistent gains from augmentation or attention.
We evaluated SimCLR ~\cite{simclr} and contrastive proto ~\cite{proto-contrastive}. Specifically, for the contrastive framework, we employed SpecAugment techniques such as Time Masking and Frequency Masking to randomly mask blocks of time and frequency on the log-mel spectrograms to generate the positive pairs for the contrastive loss.

\begin{table}[ht]
\centering
\caption{Contrastive few-shot accuracy (\%) under IID and OOD settings.}
\resizebox{\textwidth}{!}{%
\begin{tabular}{l|c|cc|cc}
\hline
\multirow{2}{*}{\textbf{Method}} & \multirow{2}{*}{\textbf{With Augmentations}} 
  & \multicolumn{2}{c|}{\textbf{1-shot}} 
  & \multicolumn{2}{c}{\textbf{5-shot}} \\
 & & IID & OOD & IID & OOD \\
\hline
Proto + Contr + Conv64~\cite{proto-contrastive} (2025)                & No  & $59.25$ & $55.94$ & $75.27$ & $64.48$ \\
Proto + Contr + Attn + Conv64 ~\cite{proto-contrastive} (2025)               & No  & $54.54$ & $53.61$ & $74.67$ & $70.50$ \\
\hline
Proto + SimCLR + ResNet12    & Yes & $44.08$ & $39.85$ & $54.15$ & $49.00$ \\
Proto + SimCLR + ResNet18    & Yes & $30.38$ & $29.06$ & $35.53$ & $30.42$ \\
\hline
\end{tabular}%
}
\label{tab:contrastive_benchmark}
\end{table}

Moreover, SimCLR-pretrained backbones exhibit degraded few-shot accuracy as model capacity increases, highlighting a mismatch between contrastive pretraining and distance-based few-shot inference in the presence of background shifts.

\FloatBarrier
\section{Limitations}
\label{sec:limitations}
SpurAudio is designed as a controlled benchmark for studying spurious foreground--background correlations in few-shot audio classification. Its mixtures are synthetic, combining foregrounds with semantically paired backgrounds. This design enables precise control over the spurious association between the foreground and backgrounds signals. Our distributional analysis (Appendix~\ref{sec:appendix_distribution_spuraudio}) shows close alignment with real audio in CLAP embedding space. At the same time, it abstracts away some factors present in fully in-situ recordings, such as room effects, recording variability, and spatial cues. Finally, SpurAudio inherits the licenses, biases, and consent regimes of its five constituent datasets, so any biases in those corpora propagate into our benchmark and any downstream analysis conducted with it.

% \FloatBarrier
% \section{Additional Few Shot Algorithms}

% We have extended our evaluation to include two more recent few-shot learning methods: HELA-VFA~\cite{lee2024hela}, BPA\cite{shalam2024balanced}. The results are presented below:

% \begin{table}[ht]
% \centering
% \caption{IID and OOD accuracy with generalization gap for HELA-VFA and BPA methods across backbones and shot settings.}
% \begin{tabular}{l|l|c|ccc}
% \hline
% \textbf{Method} & \textbf{Backbone} & \textbf{Shot} & \textbf{IID} & \textbf{OOD} & \textbf{GAP} \\
% \hline
% \multirow{2}{*}{HELA-VFA} & \multirow{2}{*}{ResNet12 + attn} 
%   & 1 & $48.35$ & $46.49$ & $1.86$ \\
%   &  & 5 & $64.26$ & $55.57$ & $8.70$ \\
% \hline
% \multirow{4}{*}{BPA} & \multirow{2}{*}{Conv64}
%   & 1 & $53.54$ & $50.88$ & $2.66$ \\
%   &  & 5 & $72.30$ & $62.60$ & $9.70$ \\
% \cline{2-6}
% & \multirow{2}{*}{ResNet12}
%   & 1 & $59.88$ & $52.82$ & $7.06$ \\
%   &  & 5 & $78.70$ & $73.30$ & $5.40$ \\
% \hline
% \end{tabular}
% \label{tab:extra_methods_comparison}
% \end{table}

% We note that HELA-VFA uses an attention layer to compute support prototypes on top of a Resnet12 encoder. On the other hand, BPA is a transductive method that clusters queries with the supports by using the Sinkhorn algorithm.

% We finalize this experiment with the same recurring observation that the IID-OOD gap persists consistently across both methods and all configurations, where in this case the gaps reach up to $9.7\%$ in the 5-shot setting.

% \newpage
% \input{checklist.tex}

\end{document}